\documentclass[journal]{IEEEtran}

\usepackage{amsfonts,amssymb}
\usepackage{multirow}
\usepackage{bbm}
\usepackage{url}
\usepackage{amsmath}
\usepackage{algorithm}
\usepackage{algpseudocode}
\usepackage{amsfonts}
\usepackage{graphics} 
\usepackage{graphicx}
\usepackage{array}
\usepackage{amsmath}
\usepackage[hidelinks,colorlinks]{hyperref}
\usepackage{mathtools}
\usepackage{cite}
\usepackage{textcomp}
\usepackage{cleveref}
\usepackage{times}
\usepackage{epsfig}
\usepackage{gensymb}
\usepackage{multirow}
\usepackage{caption}
\usepackage{pifont}
\usepackage{subfigure}
\usepackage{url}
\usepackage{tikz}
\usepackage{cleveref}
\usepackage{soul}
\usepackage{mathtools,xparse}
\usepackage[super]{nth}

\usepackage{float}  

\DeclareMathOperator*{\argminB}{argmin}   

\usepackage{makecell}   
\usepackage{booktabs}   

\ifCLASSINFOpdf

\else

\fi

\begin{document}

\title{An Object SLAM Framework for Association, Mapping, and High-Level Tasks}

\author{Yanmin~Wu,
        Yunzhou~Zhang,
        Delong~Zhu,
        Zhiqiang~Deng,
        Wenkai~Sun,
        Xin~Chen,
        and Jian~Zhang 
\thanks{
This work was supported by National Natural Science Foundation of China (No. 61973066), Major Science and Technology Projects of Liaoning Province (No.2021JH1/10400049), Fundation of Key Laboratory of Aerospace System Simulation (No.6142002200301), Fundation of Key Laboratory of Equipment Reliability (No.WD2C20205500306) and Fundamental Research Funds for the Central Universities (N2004022).}
\thanks{Yanmin Wu and Xin Chen are with Faculty of Robot Science and Engineering, Northeastern University, Shenyang 110819, China (Email: {\tt\small wuyanminmax@gmail.com}).}
\thanks{Yunzhou Zhang, Zhiqiang Deng, and Wenkai Sun are with College of Information Science and Engineering, Northeastern University, Shenyang 110819, China. (\textbf{Corresponding author}: Yunzhou Zhang, Email: {\tt\small zhangyunzhou@mail.neu.edu.cn}).}
\thanks{Delong Zhu is with the Department of Electronic Engineering, The Chinese University of Hong Kong, Shatin, N.T., Hong Kong SAR, China.}
\thanks{Jian Zhang is with the School of Electronic and Computer
Engineering, Peking University Shenzhen Graduate School, Shenzhen 518055,
China.}}

\markboth{IEEE TRANSACTIONS ON ROBOTICS}%
{Shell \MakeLowercase{\textit{et al.}}: Bare Demo of IEEEtran.cls for IEEE Journals}

\maketitle

\begin{abstract}
Object SLAM is considered increasingly significant for robot high-level perception and decision-making. Existing studies fall short in terms of data association, object representation, and semantic mapping and frequently rely on additional assumptions, limiting their performance. In this paper, we present a comprehensive object SLAM framework that focuses on object-based perception and object-oriented robot tasks. First, we propose an ensemble data association approach for associating objects in complicated conditions by incorporating parametric and nonparametric statistic testing. In addition, we suggest an outlier-robust centroid and scale estimation algorithm for modeling objects based on the iForest and line alignment. Then a lightweight and object-oriented map is represented by estimated general object models. Taking into consideration the semantic invariance of objects, we convert the object map to a topological map to provide semantic descriptors to enable multi-map matching. Finally, we suggest an object-driven active exploration strategy to achieve autonomous mapping in the grasping scenario. A range of public datasets and real-world results in mapping, augmented reality, scene matching, relocalization, and robotic manipulation have been used to evaluate the proposed object SLAM framework for its efficient performance.
\end{abstract}


%
\IEEEpeerreviewmaketitle

\section{Introduction}
%
%
%
%

\IEEEPARstart{T}{HE} fundamental issues in terms of the accuracy and efficiency of visual SLAM have been vastly improved over the past two decades, which enables a wide application of visual SLAM in robots, autonomous driving, and augmented reality. The next generation of SLAM will require support for more intelligent tasks with a better capacity that we call ``geometric and semantic Spatial AI perception''\cite{davison2018futuremapping}. This will greatly extend the scope of traditional geometric localization and mapping.

In terms of geometric perception (\textit{e.g.}, point-based appearance modeling and handcrafted feature-based localization), more visual landmarks, such as the line\cite{wang2021line}, edge\cite{zhou2018canny}, and plane\cite{yunus2021manhattanslam}, are exploited to overcome environmental and motion challenges. Omnidirectional geometric perception is achieved by multi-sensor fusion of visual, thermal, inertia, LiDAR, GNSS, and UWB\cite{zhao2020tp,cao2021gvins,nguyen2021viral}. In onboard applications, these versatile and robust algorithms are extensively employed. However, due to the absence of semantic cues, geometric clues alone are insufficient for intelligent robot interaction and active decision-making, such as semantic mapping, object goal navigation, and object searching. This article focuses on another aspect of the next-generation SLAM: semantic perception, aiming at representing and understanding environmental information at a semantic level, which extends beyond the basic geometric appearance and position perception.

In semantic SLAM, the semantic cues provided by deep learning technology play an essential role in various sub-components, \textit{e.g.}, localization, mapping, loop closure, and optimization. In this work, we focus on \textbf{semantic-aided mapping and exploring multiple high-level applications based on the semantic map.} Popular semantic mapping pipelines~\cite{mccormac2017semanticfusion,yang2017semantic,rosinol2020kimera} parallelize the geometric SLAM workflow and learning-based semantic segmentation, and then annotate 3D point clouds (or volume, mesh) with 2D image segmentation labels. Finally, the multi-frame segmentation results are fused with probabilistic approaches to build a global semantic map. Although, these point clouds-based semantic maps are visually appealing, they are not detailed and lack sufficient instance-specific information to assist the robot in performing fine-grained tasks. Therefore, the first insight of this article is that \textbf{a helpful semantic map for robot operation should be instance- and object-oriented}.

Object SLAM is an object-oriented branch of semantic SLAM that focus on constructing the map with objects as central entities and typically takes instance-level segmentation or object detection as the semantic network. Most studies on sparse SLAM~\cite{frost2018recovering,iqbal2018localization} associate point clouds with object landmarks and take the centroids of point clouds as the positions of objects. Other studies~\cite{mu2016slam,sunderhauf2017meaningful,grinvald2019volumetric,sharma2020compositional} on dense SLAM improve the mapping results by denser point clouds and more precise segmentation/detection, enabling object-level reconstruction and dense semantic representation of objects. Nonetheless, these studies focus on the accuracy of object position, while the orientation and size of objects are not investigated, which are indeed indispensable for robotic tasks like manipulation and navigation. The second point of view presented in this article is that \textbf{the object's position, orientation, and size in the map should all be parameterized}.

Object parameterization or representation is one of the primary missions of object SLAM. To address this problem, typical studies \cite{salas2013slam++,choudhary2016multi,labbe2020cosypose} usually include object models as a prior and the point clouds or shapes of target objects are known. The pose estimation of objects is then achieved by model retrieval and matching. The prior model is also integrated into the map and engaged in object-level bundle adjustment. Studies\cite{bowman2017probabilistic,parkhiya2018constructing,joshi2018integrating,sucar2020nodeslam} are examples that focus on categorized object models, which only take a partial knowledge of the object, such as the structure and shape, as the prior and use requires one model to represent one category. Although the object parameters are well encoded in the prior instance model or category model, obtaining the prior knowledge is difficult and expensive. In addition, the generalization capability of these models is limited. The third observation made in this work is that \textbf{objects should be represented by general models with a high degree of generality and a low cost prior}, such as the cube, cylinder, and quadric.

To summarize, this work aims to present an object SLAM framework that generates an object-oriented map with general models, which can parameterize the position, orientation, and size of objects in the map. In addition, we further explore high-level applications based on the object-oriented map. Some previous studies~\cite{yang2019cubeslam,nicholson2018quadricslam} pursued a similar objective but encountered the following challenges. \textbf{1)} The data association algorithms are insufficiently robust and accurate for dealing with complex settings involving various classes and numbers of objects. \textbf{2)} Object parametrization is sloppy, typically depending on strict assumptions or achieving only incomplete modeling, both of which are difficult to achieve in practice. \textbf{3)} Most studies focus on creating the object or semantic map, but the application in downstream tasks is not explored, nor is the map's utility demonstrated. Instead, we discuss \textbf{not only the fundamental techniques of object mapping but also high-level and object map-oriented applications}.

In this paper, we propose an object SLAM framework to achieve the desired objective while overcoming the aforementioned challenges. 
Firstly, we integrate the parametric and nonparametric statistic tests and the traditional IoU-based method to conduct model ensembling for data association. Compared with conventional methods, our approach sufficiently exploits the nature of different statistics, \textit{e.g.}, Gaussian, non-Gaussian, 2D, and 3D measurements, hence exhibiting significant advantages in association robustness. 
Then, for object parametrization, we offer an algorithm for centroid, size, and orientation estimation and an object pose initialization approach based on the iForest (isolation forest) and line alignment. The proposed methods are robust to outliers and exhibit high accuracy, which significantly facilitates the joint pose optimization process. 
Finally, an object-oriented map is constructed using the general models taking cubes and quadrics as representations.
Based on the map, we develop an augmented reality system to enable virtual-real fusion and interaction, transplant a framework for the robot arm to realize common objects' modeling and grasping, and propose a novel object descriptor for sub-scene matching and relocalization.

This article extends our previous works ~\cite{wu2020eao,wu2020object}. Extensions include semantic descriptor-based scene matching/relo-\ calization (Sec.\ref{S_TOPO} and Sec.\ref{sec:exp_scene_matching}) and expanded experiments and analysis (Sec.\ref{S_DIS}). The contributions are summarized:
\begin{itemize}
	\item We propose an ensemble data association strategy that can effectively aggregate different measurements of the objects to improve association accuracy.
	\item We propose an object pose estimation framework based on the iForest and line alignment, which is robust to outliers and can accurately estimate the pose and size of objects.
	\item We build a lightweight and object-oriented map with general models, upon which we develop an augmented reality application aware of occlusion and collisions.
	\item We extend the object map to a topological map and design a semantic descriptor based on the parameterized object information to enable multiple scene matching and object-based relocalization.
	\item We integrate object SLAM with robotic grasping tasks to propose an object-driven active exploration strategy that accounts for object observation completeness and pose estimation uncertainty, achieving accurate object mapping and complex robotic grasping.
	\item We propose a comprehensive object SLAM framework that explores the key challenges and powerfully demonstrates its utility in various scenarios and tasks.
\end{itemize} 

 

\section{Related Work}

\subsection{Data Association}

Data association establishes the 2D-3D relationship between objects in image frames and the global map and the 2D-2D correspondence of objects between sequential frames. The most popular strategy considers it an object-tracking issue \cite{frost2018recovering,chen2020accurate,zhang2020simple}. Li \textit{et al.} \cite{li2019semantic} project 3D objects to the image plane and then perform association using the projected 2D bounding boxes via the Hungarian object tracking algorithm. Some approaches \cite{mccormac2018fusion++,sharma2020compositional,wang2021dsp,xu2022learning} use Intersection over Union (IoU) algorithm to track objects between frames, while tracking-based approaches are prone to create erroneous priors in complicated contexts resulting in wrong association results.

Some studies increase the utilization of shared information. Liu~\textit{et al.} \cite{liu2019global} create a descriptor representing the topological relationships between objects, and instances with the greatest number of the shared descriptors are considered identical. Instead, Yang \textit{et al.} \cite{yang2019cubeslam} suggest using the number of matched map points on detected objects as an association criterion. Grinvald \textit{et al.} \cite{grinvald2019volumetric} preset a measurement of semantic label similarity, while Ok \textit{et al.} \cite{ok2019robust} propose to leverage the hue saturation histogram correlation. Sünderhauf~\textit{et al.} \cite{sunderhauf2017meaningful} compare the distance between distinct instances more directly. Typically, the designed criteria are inadequately general, exhaustive, or robust, leading to incorrect associations.

In terms of learning-based studies, Xiang \textit{et al.} \cite{xiang2017darnn} suggest utilizing recurrent neural networks to achieve semantic label data association between consecutive images. However, they only focus on pixel-level associations. Similarly, Li \textit{et al.} \cite{li2021odam} use an attention-based GNN to maintain the detected 2D and 3D attributes. Merrill \textit{et al.} \cite{merrill2022symmetry} propose a keypoint-based object-level SLAM system that projects the 3D key points to the image as the prior of the objects in the next frame. However, this method is not verified on the SLAM dataset and cannot be generalized to previously unseen objects. Using a deep graph convolutional network, Xing \textit{et al.} \cite{xing2022descriptellation} extract object features and perform feature matching. Nevertheless, this method is only suitable for well-constructed maps and is challenging for incremental maps of real-time SLAM.

Another viable option is the probabilistic-based solution. Bowman \textit{et al.} \cite{bowman2017probabilistic} use a probabilistic method to model the data association process and leverage the EM algorithm to identify correspondences between observed landmarks. Subsequent studies~\cite{strecke2019fusion,yang2019probabilistic} extend the concept to associate dynamic objects or perform dense semantic reconstructions. However, their efficiency is limited by the high cost of the EM optimizers. Weng \textit{et al.} \cite{mu2016slam} present a nonparametric Dirichlet process for semantic data association, which can address the challenges that arise when the statistics do not follow a Gaussian distribution. Later, Zhang \textit{et al.} \cite{zhang2019hierarchical} and Ran \textit{et al.} \cite{ran2021not} introduce two variations of the hierarchical Dirichlet method for lowering association uncertainty. Iqbal \textit{et al.} \cite{iqbal2018localization} also demonstrate the efficiency of nonparametric data association. However, this strategy cannot properly address statistics with Gaussian distributions and is thus incapable of adequately leveraging diverse data in SLAM. We combine the parametric and nonparametric methods to execute model ensembling, which exhibits superior association performance in complex scenarios with numerous object categories.

\vspace{-5pt}
\subsection{Object Representation}

Object representation in object SLAM can be divided into shape reconstruction-based and model-based methods. For the former category, Sucar \textit{et al.} \cite{sucar2020nodeslam} infer object volume from images using a Variational Auto Encoder and then jointly optimize object shape and pose. Wang \textit{et al.} \cite{wang2021dsp} adopt DeepSDF \cite{park2019deepsdf} as shape embedding, minimizing the surface consistency and depth rendering loss by observed point clouds. Similarly, Xu \textit{et al.} \cite{xu2022learning} train a shape completion network based on the pre-trained DeepSDF to achieve complete shape reconstruction of partially seen objects. However, these methods are data-driven and significantly dependent on large-scale shape priors.

Model-based object representations are classified broadly into three types: prior instance-level models~\cite{salas2013slam++,choudhary2016multi,labbe2020cosypose,han2021reconstructing}, category-specific models, and general models. Prior instance-level models rely on a well-established or trained database, such as detailed point clouds or CAD models. Since such models must be known in advance, their application scenarios are limited. In addition, studies~\cite{parkhiya2018constructing,joshi2018integrating,sucar2020nodeslam} on category-specific models focus on identifying category-level characteristics. Parkhiya \textit{et al.} \cite{parkhiya2018constructing} and Joshi \textit{et al.} \cite{joshi2018integrating} represent different categories through the combination of line segments, but the category-specific feature is insufficiently general and is impossible to describe an excessive number of classes.

The general object models are represented by simple geometric elements, \textit{e.g.}, cube, quadric, and cylinder, which are the most efficient models. There are two typical modeling method. The first type infers the 3D pose from the 2D detection result. Yang \textit{et al.} \cite{yang2019cubeslam} leverage the vanishing point to sample 3D cube proposal from a single view, and then optimize the object pose using geometric measurements. Nicholson \textit{et al.} \cite{nicholson2018quadricslam} combine multi-view observation to parametrize object landmarks as constrained dual quadrics. Subsequent studies~\cite{ok2019robust,cao2022object} refine quadric representation by incorporating shape and semantic priors and plane constraints. However, this inference from the 2D object has a poor precision with significant errors. Li \textit{et al.} \cite{li2021odam} apply superquadric to tune between 3D boxes and quadrics adaptively. However, they rely on additional 3D object detection. 
Another type of methods resolve the 3D object pose by 3D point cloud measurements. Some studies~\cite{mu2016slam,frost2018recovering,iqbal2018localization} portray object position using point cloud centers, which is an imprecise way of expressing object properties. Runz \textit{et al.} \cite{runz2018maskfusion} get a dense object reconstruction result using more accurate instance and geometry-based segmentation. While the object's position and size are viable, the orientation is ignored. Some other studies~\cite{yang2019cubeslam,lin2021topology,lu2022real} involving direction estimation use the geometric characteristics of images or point clouds for orientation sampling and analysis. However, they face the problem of insufficient robustness. In contrast, studies~\cite{li2020view,ming2021object} use learning-based methods from orientation regression from the image, but there are issues in terms of accuracy and generalization. 
In this work, based on the general object model, we propose an outlier-robust object pose estimation algorithm using the iForest and line alignment method for better parametrization of object size and orientation.

\vspace{-3mm}
\subsection{Semantic Scene Matching}

Scene matching is critical for robot relocalization, loop closure, and multi-agent collaboration. Conventional studies \cite{mur2017orb,schmuck2019ccm} rely on keyframes and geometric features, which are vulnerable to failure when faced with changes in viewpoint, illumination, and appearance. Conversely, semantic-based scene matching is more efficient because of the time and space invariance of the semantic information (\textit{e.g.}, label and size).

Gawel \textit{et al.} \cite{gawel2018x} focus on global scene matching for multi-view robots and they propose a random-walk-based semantic descriptor to enable global localization by semantic graph match. Guo \textit{et al.} \cite{guo2021semantic} investigate large-scale scene matching problem and suggest a semantic histogram-based fast graph matching algorithm, resulting in more accurate and faster localization and map merging. However, these methods only account for the global matching of large scenes, disregarding the local information. Additionally, the semantic information is not at the object level. Liu \textit{et al.} \cite{liu2019global} are interested in the issue of localization when environmental appearance changes. They suggest characterizing the scene with a dense semantic topology map and performing 6-DOF object localization by matching object descriptors. Similarly, Li \textit{et al.} \cite{li2019semantic} focus on the relocalization of perspective changes. They use object landmarks to establish the correspondence between different views and conduct relocalization through graph matching based on the Hungarian algorithm. However, given the limited number of objects and the fact that they are not well-parameterized, their method is doubtful in a complicated setting with several repeating objects. 
To address the loop closure problem in multi-object scenes, Qin \textit{et al.} \cite{qin2021semantic} propose to generate semantic sub-graphs using objects' semantic labels  and then leverage Kuhn–Munkres to align sub-graphs for estimating the transformation. However, the semantic clues are only used to determines the resemblance of scenes,while the translation between them is still calculated by geometric measures instead of semantic measures. In this work, we focus on scene matching and scene translation with multiple objects. Similar to previous studies, we create a topological map and design an object descriptor. In the map, the objects are fully parameterized, and the matching strategy based on object descriptors are also improved.

\vspace{-10pt}
\subsection{Active Perception and Object Map-based Grasping}  

Active perception is the process of actively adjusting sensor states by analyzing existing data to gather more valuable information for executing specific tasks, which is a critical characteristic of robot autonomy. 
Zhang \textit{et al.} \cite{zhang2019beyond} leverage Fisher information to predict the optimal sensor position to reduce localization uncertainty. 
Zeng \textit{et al.} \cite{zeng2020semantic} exploit prior knowledge between objects to establish a semantic link graph for active object search. 
More specifically, active mapping is a specific type of active perception task concerned with autonomous map construction. 
Charrow \textit{et al.} \cite{charrow2015information} utilize the quadratic mutual information to guide 3D dense mapping. 
Wang \textit{et al.} \cite{wang2019srm} also leverage the mutual information to perform Next-Best View (NBV) selection on a sparse road map, which subsequently acts as a semantic landmark to aid the mapping process. 
Kriegel \textit{et al.} \cite{kriegel2015efficient} propose a surface reconstruction method for single unknown objects. In addition to the information gain, they also integrate the measurement of reconstruction quality into the objective function, achieving high accuracy and completeness.
The key to active mapping is defining the measurement and strategy to guide the agent moving autonomously. 
We propose an information entropy-based uncertainty quantification and an object-driven active exploration strategy. Another significant difference from other methods is that the output of our suggested method is an object map compatible with complex robot manipulation tasks.

The object map encoded with object pose is available for robot object manipulation tasks, such as object placement and arrangement.
Wada \textit{et al.} \cite{wada2020morefusion} propose reconstructing objects by incremental object-level voxel mapping. Voxel points initialize the object pose, and the ICP algorithm is then used to align the initialized object with the CAD model to optimize the pose further, which is heavily dependent on the CAD model's registration accuracy. 
In NodeSLAM \cite{sucar2020nodeslam}, the object is regarded as a landmark and is involved in joint optimization to help generate an accurate object map. The primary deficiency of this method is that the model requires a tedious category-level training process for each object. 
Labb{\'e} \textit{et al.} \cite{labbe2020cosypose} present a single-view 6-DoF object pose estimation method and utilize the object-level bundle adjustment in the SLAM framework to optimize the object map. However, this method only focuses on known objects. 
Almeida \textit{et al.} \cite{almeida2019detection} leverage the SLAM framework to densely map unknown objects for accurate grasping point detection, but the object pose is not estimated. 
In this work, we use the proposed SLAM framework to generate the object map actively, which enables the global perception to aid the robot in performing more intelligent tasks autonomously. Additionally, unlike previous studies, we focus on the pose estimation of unknown objects.

\section{System Overview}
\label{S_SO}

\begin{figure*}[thbp]
	\centering
    \setlength{\abovecaptionskip}{3pt}
	\captionsetup{belowskip=-10pt}
	\includegraphics[width=0.95\textwidth]{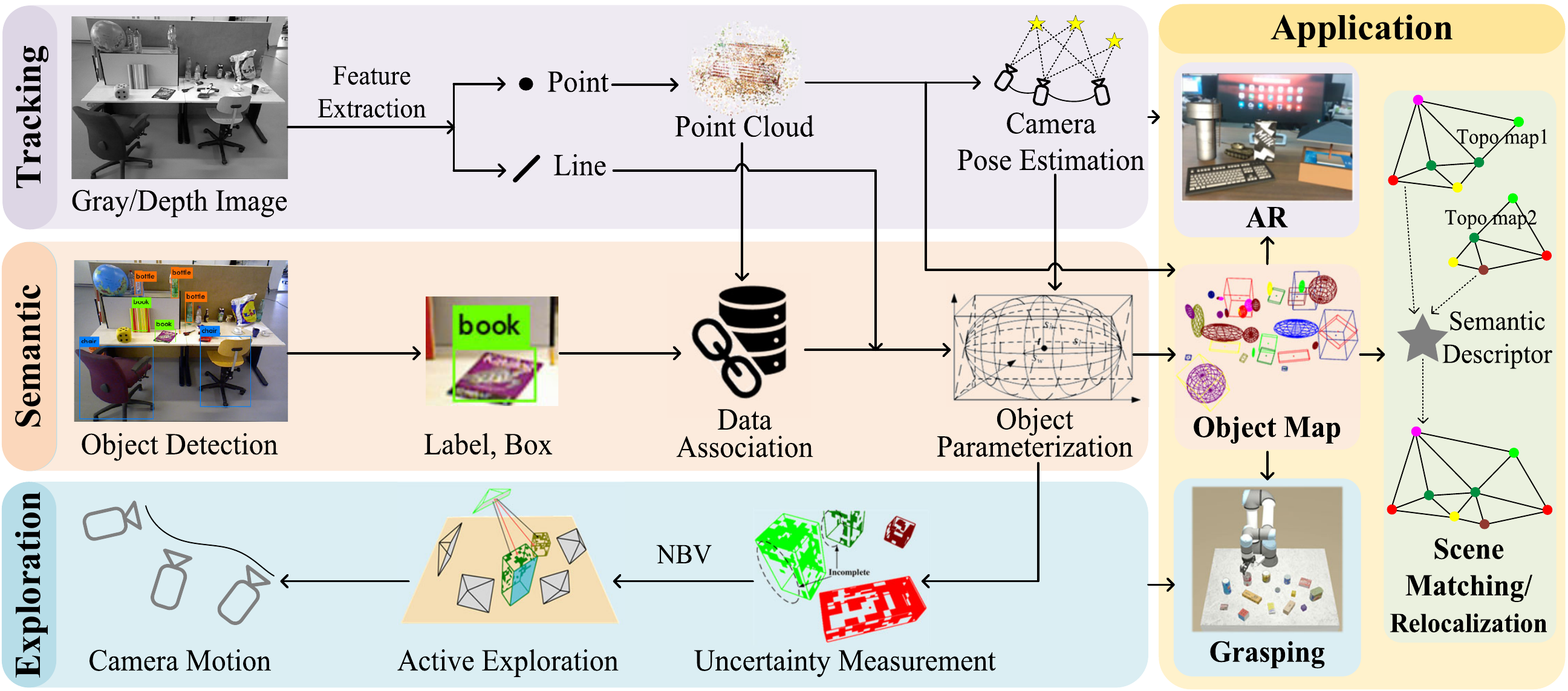}
	\caption{The proposed object SLAM framework.}
	\label{framework}
\end{figure*}

The proposed object SLAM framework is demonstrated in Fig. \ref{framework} including four parts. The \textbf{tracking module} builds upon the ORB-SLAM2\cite{mur2017orb}, which generates incremental sparse point clouds and estimates camera pose by extracting and matching multi-view features. Our main contributions lie in the remaining three parts. The \textbf{semantic module} employs YOLO\cite{redmon2018yolov3} as the object detector to provide semantic labels and bounding boxes which are then combined with point cloud measurements to associate the 2D detected objects with 3D global objects. After that, the iForest and the line alignment algorithms are applied to refine the point clouds and 2D lines generated by the tracking module. Based on the association and refinement results, the objects are parameterized using the cube and quadric models.

The \textbf{object map} comprises of multiple parameterized objects and achieves a lightweight representation of the environment, which is a vital component of the application module. For the \textbf{augmented reality} application, virtual models' 3D registration is based on the real-world object pose rather than the conventional point-based approach. Additionally, we convert the object map to a topological map, a graph representation of the objects and their relative poses. Based on this map, a semantic descriptor is designed to enable multi-scene matching and relocalization tasks.

The accurate object pose is encoded in the object map, which provides the fundamental clues (\textit{e.g.}, grasping points) for \textbf{robotic grasping} applications. Notably, the object map is created actively, as depicted in the \textbf{exploration module}. Here, we propose an uncertainty measurement model to predict the next best view for exploration. The manipulator then actively moves to scan the table with the best view until building up a complete and accurate object map.

In short, the proposed object SLAM leverages geometric and semantic measurements to simultaneously realize camera localization and object map building, resulting in a comprehensive system that addresses various challenges in this field and facilitates many intelligent and fascinating applications. The remainder of our paper is organized as follows: \ref{S_DA} and \ref{S_OP} present the principal theories of data association and object parameterization. The semantic descriptor and scene-matching method are defined in Section \ref{S_TOPO}. The active exploration strategy is introduced in Section \ref{S_ACTIVE}. Section \ref{S_EX} demonstrates the performance of our system through comprehensive experiments. Section~\ref{S_DIS} provides the discussion and analysis, and Section \ref{S_CON} provides the conclusion.

\section{Object-Level Data Association}
\label{S_DA}

\begin{figure}[t]
	\centering
    \setlength{\abovecaptionskip}{3pt}
	\captionsetup{belowskip=-10pt}
	\includegraphics[width=0.49\textwidth]{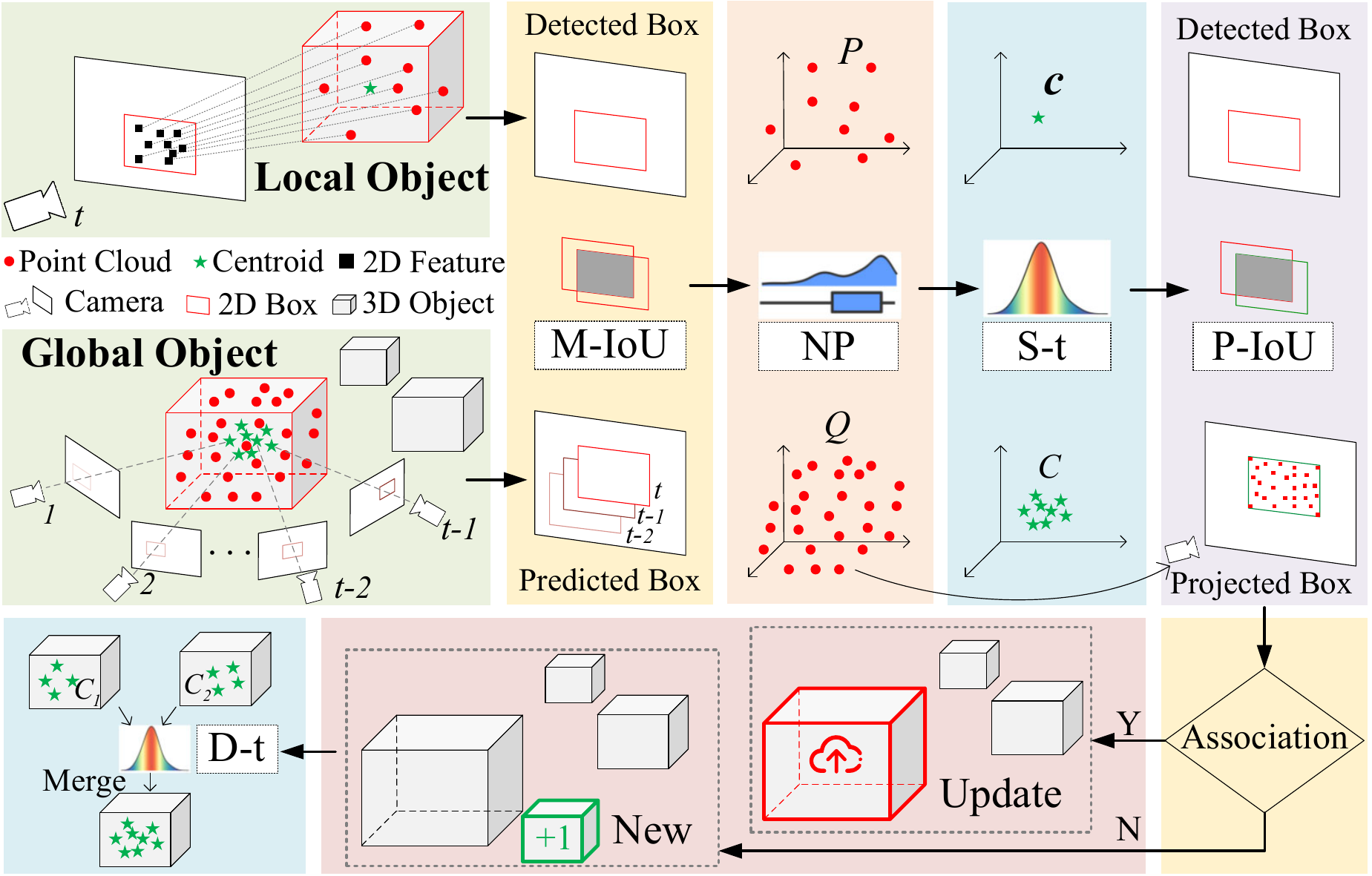}
	\caption{The pipeline of object-level data association.}
	\label{data_association}
\end{figure}

Fig. \ref{data_association} presents the pipeline of the proposed data association strategy. The \textbf{local object} is a 3D instance observed in the current single view ($t$), where the point clouds correspond to ORB features that lie in the 2D bounding box, and the centroid is the mean of the points. The \textbf{global object} is an entity observed by multiple frames (before $t$) and already exists on the map, where point clouds and centroids also come from incremental measurement of multi-views. Data association aims to determine which global object in the map is associated with the local object in the current view.
As shown in the pipeline, the camera motion IoU (M-IoU), nonparametric (NP) test, single sample-$t$ (S-$t$) test, and project IoU (P-IoU) will be used to determine whether or not the association is successful. In the experiment, the successful case should satisfy the fourth item and any of the first three items. If so, the existing global object will be updated; otherwise, a new global object will be created. Finally, the double sample-$t$ (D-$t$) test is utilized to check whether duplicates exist.

Throughout this section, the following notations are used:
\begin{itemize}
	\item $P \in \mathbb{R}^{3 \times |P|}, Q \in \mathbb{R}^{3 \times |Q|}$ - the point clouds of the local object and the global object.
	\item $\mathcal{R}$ - the rank (position) of a data point in a sorted list.
	\item $\mathbf{c} \in \mathbb{R}^{3 \times 1}$ - the currently observed local object centroid.
	\item $C=[\mathbf{c}_1, \mathbf{c}_2, \dots,  \mathbf{c}_{|C|}] \in \mathbb{R}^{3 \times |C|}$ - a series of centroids of a global object observed by historical views. $C_1,C_2$ are similar.
	\item $f(\cdot)$ - the probability function used for statistic test.
	\item $m(\cdot),\sigma(\cdot) \in \mathbb{R}^{3 \times 1}$ - the mean and variance functions.
\end{itemize}

\vspace{-5pt}
\subsection{Intersection over Union (IoU) Model}
If a global object is observed in the previous two frames ($t-1$ and $t-2$), we then predict the bounding box in the current frame ($t$) based on the hypothesis of uniform motion, and calculate the IoU between the predicted box of a global object and the detected box of a local object, which we defined as Motion-IoU (See Fig. \ref{data_association} M-IoU part). If the IoU value is large enough, there may be a potential association between the two objects. 
After NP and S-t (see Sections \ref{SNP} and \ref{SDAT}), the Project-IoU will validate this association by projecting 3D point clouds of the global object to 2D points on the current frame and fitting a box to these points. After that, we calculate the IoU between the projected and detected boxes (See Fig.~Fig. \ref{data_association} P-IoU part).

\subsection{Nonparametric Test Model}
\label{SNP}
The m-IoU model provides a straightforward and efficient way of dealing with the scenario of consecutive frames. However, it will malfunction when 1) the object is missed by the detector, 2) the object is occluded, or 3) the object disappears from the camera view.

The Nonparametric test model does not require continuous observations of the object, and can be directly applied to process two sets of point clouds, $P$ and $Q$ (see Fig. \ref{data_association} NP part), based on the hypothesis that point clouds follow a non-Gaussian distribution (which will be demonstrated in Section \ref{SEXDDS}). Theoretically, if $P$ and $Q$ represent the same object, they should follow the same distribution, \textit{i.e.}, $f_P = f_Q$. We use the \textit{Wilcoxon Rank-Sum test} \cite{wilcoxon1992individual} to verify whether the null hypothesis holds.

We first mix the two point clouds $X = [P|Q] = [\mathbf{x}_1, \mathbf{x}_2, \dots, \mathbf{x}_{|X|}]\in \mathbb{R}^{3\times(|P|+|Q|)}$, and then sort $X$ in three dimensions respectively. Define $W_{P} \in \mathbb{R}^{3\times1}$ as follows, 
\begin{equation}
\small
 W_{P}=\left \{ \sum_{k=1}^{|X|} \mathcal{R}(1\{\mathbf{x}_k\in P \})-\frac{|P|(|P|+1)}{2} \right \},
\vspace*{-0.4\baselineskip}
\end{equation}
and $W_{Q}$ is with the same formula. The Mann-Whitney statistics is $W \small{=} \min({W_{P}},{W_{Q}})$, which is proved to follow a Gaussian distribution asymptotically \cite{sidney1957nonparametric,lehmann1975nonparametrics}. Herein, we essentially construct a Gaussian statistics using the non-Gaussian point clouds. The mean and variance of $W$ are calculated:
\vspace*{-0.4\baselineskip}
\begin{align}
 m(W) &= (|P||Q|) / 2,\\
 \sigma(W) &= \frac{{|P||Q|\Delta^{+}}}{{12}} - \frac{{|P||Q|(\sum_{i} {\tau _i^3}  - \sum_{i} {{\tau _i}} )}}{{12(|P|+|Q|)\Delta^{-}}},
\vspace*{-0.4\baselineskip}
\label{eq:3}
\end{align}
where $\Delta^{+} = |P|+|Q|+1, \Delta^{-} = |P|+|Q|-1$, and $\tau \in P \cap Q$. 
$\tau$ represents the number of shared points between two objects; because its value is small, the complicated and low-contributing second term in Eq. (\ref{eq:3}) is ignored in our implementation.
 
To make the null hypothesis stand, $W$ should meet the following constraints:
\vspace*{-0.4\baselineskip}
\begin{equation}
 \label{nonp}
 f(W) \geq f\left(r_r \right)=f\left(r_l \right)=\alpha/2,
\vspace*{-0.4\baselineskip}
\end{equation}
where $\alpha$ is the significance level, $1-\alpha$ is the confidence level, and $[r_l, r_r] \approx [m-s\sqrt{\sigma}, m+s\sqrt\sigma\,]$ defines the confidence region. The scalar $s > 0$ is defined on a normalized Gaussian distribution $\mathcal{N}(s|0,1) \small{=} \alpha$. In summary, if the Mann-Whitney statistics $W$ of two point clouds $P$ and $Q$ satisfies Eq. \eqref{nonp}, we temporarily assume they come from the same object.

\subsection{Single-sample and Double-sample T-test Model}
\label{SDAT}

The single-sample $t$-test is used to process object centroids observed in different views (see Fig. \ref{data_association} S-$t$ part), which typically follow a Gaussian distribution (see Section \ref{SEXDDS}). 

Suppose the null hypothesis is that $C$ and $\mathbf{c}$ are from the same object, and define $t$ statistics as follows,
\vspace*{-0.4\baselineskip}
\begin{equation}
t = \frac{m(C) - \boldsymbol{c}}{\sigma(C)/\sqrt{\left |C  \right |}}\sim t(\left |C  \right | - 1).
\vspace*{-0.4\baselineskip}
\end{equation}
For the null hypothesis to hold, $t$ should satisfy:
\vspace*{-0.4\baselineskip}
\begin{equation}
f(t)\geq f(t_{\alpha /2,v}) = \alpha /2,
\label{condition of t-test}
\vspace*{-0.4\baselineskip}
\end{equation}
where $t_{\alpha /2,v}$ is the upper $\alpha /2$ quantile of the t-distribution of $v$ degrees of freedom, and $v=\left |C  \right | - 1$. If $t$ statistics satisfy \eqref{condition of t-test}, we temporarily assume $\mathbf{c}$ and $C$ come from the same object.

Some existing objects may be misidentified as new due to the above-described strict data association strategy, poor observation views, or erroneous object detection, resulting in duplicates.
Consequently, a double-sample $t$-test is leveraged to determine whether to merge the two objects by analyzing their historical centroids (see Fig. \ref{data_association} D-$t$ part).

Construct $t$-statistics for $C_1$ and $C_2$ as follows,
\vspace*{-0.4\baselineskip}
\begin{equation}
t=\frac{m(C_1) - m(C_2)}{\sigma_d} \sim t(\left|C_1\right| + \left |C_2 \right | - 2),\\
\vspace*{-0.4\baselineskip}
\end{equation}
\vspace*{-0.4\baselineskip}
\begin{equation}
\small
\sigma_d = \sqrt {\frac{{\left( {{\left|C_1\right|} - 1} \right)\sigma_1^2 + \left( {{\left|C_2\right|} - 1} \right)\sigma_2^2}}{{{\left|C_1\right|} + {\left|C_2\right|} - 2}}\left( {\frac{1}{{{\left|C_1\right|}}} + \frac{1}{{{\left|C_2\right|}}}} \right)},
\vspace*{-0.4\baselineskip}
\end{equation}
where $\sigma_d$ is the pooled standard deviation of the two objects. Similarly, if $t$ satisfies \eqref{condition of t-test}, $v=\left|C_1\right| + \left |C_2 \right | - 2$, it means that $C_1$ and $C_2$ belong to the same object, then we merge them.

\section{Object parameterization}
\label{S_OP}

Data association provides the global object with multi-view measurements that ensure more observations for parameterization to model objects effectively. Throughout this section, the following notations are used:
\begin{itemize}
	\item $\boldsymbol{t}=[t_x, t_y, t_z]^T$ - the translation (location) of object frame in world frame.
		\item $\boldsymbol{\theta}=[\theta_r, \theta_y, \theta_p]^T$ - the rotation of object frame w.r.t. world frame. $R(\boldsymbol{\theta})$ is matrix representation.
		\item $T=\left \{ R(\boldsymbol{\theta}), \boldsymbol{t} \right \}$ - the transformation of object frame w.r.t. world frame.
		\item $\boldsymbol{s}=[s_l, s_w, s_h]^T$ - half of the side length of a 3D bounding box, \textit{i.e.}, the scale of an object.
		\item $P_o, P_w \in \mathbb{R}^{3 \times 8}$ - the coordinates of eight vertices of a cube in object and world frame, respectively. 
		\item $Q_o, Q_w \in \mathbb{R}^{4 \times 4}$ - the quadric parameterized by its semiaxis in object and world frame, respectively, where $Q_{o}=\operatorname{diag}\left\{s_{l}^{2}, s_{w}^{2}, s_{h}^{2},-1\right\}$. 
		\item $\alpha(\cdot)$ - calculate the angle of line segments in the image. 
		\item $K,T_c$ - the intrinsic and extrinsic parameters of camera. 
		\item $\boldsymbol{p} \in \mathbb{R}^{3 \times 1}$ - the  coordinates of a point in world frame. 
\end{itemize} 

\begin{figure}[t]
		\centering
    \setlength{\abovecaptionskip}{1pt}
	\captionsetup{belowskip=-10pt}
	\includegraphics[scale=0.42]{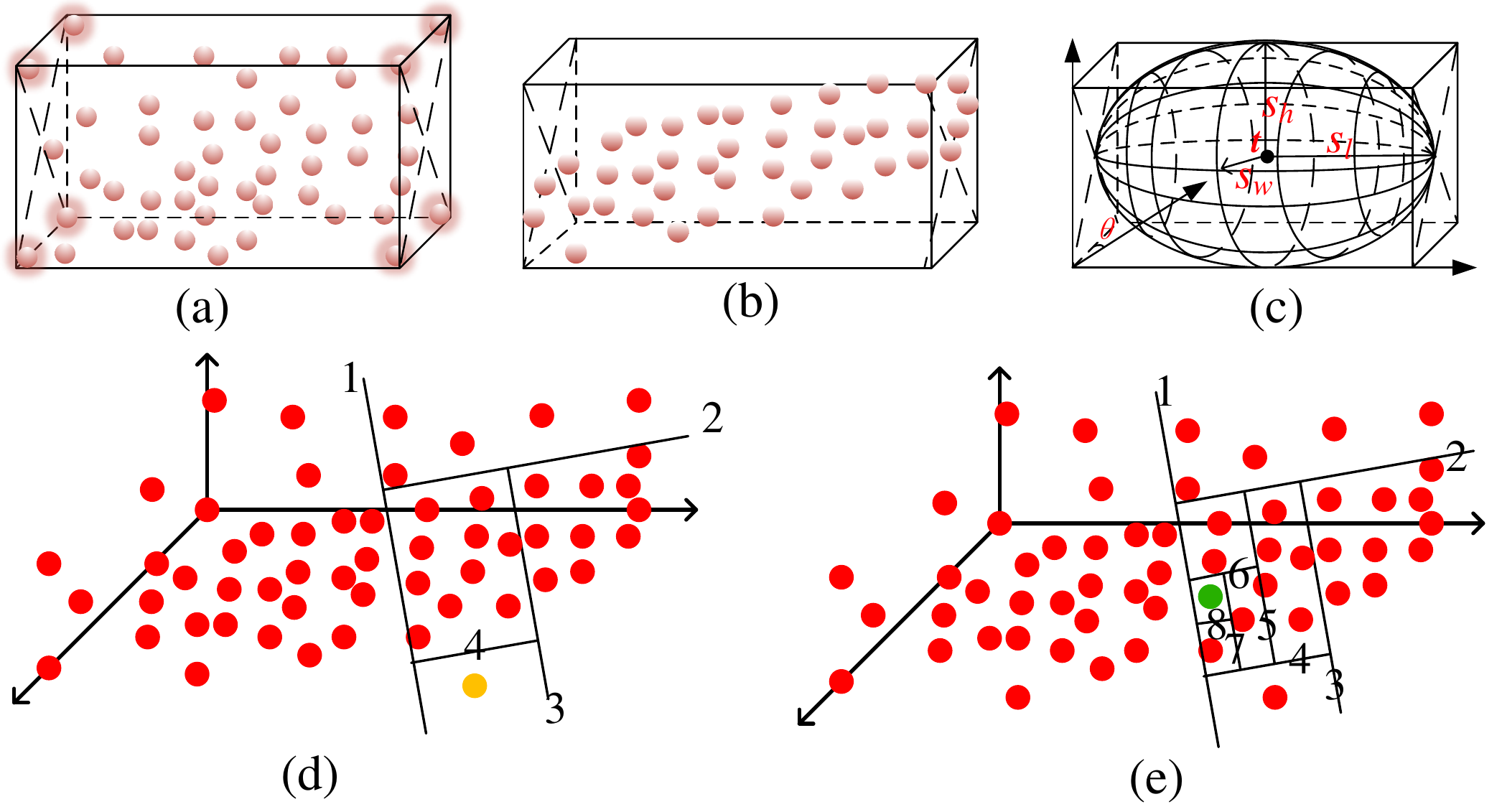}
	\caption{(a-c) Demonstration of object parameterization. (d-e) Demonstration of iForest.}
	\label{parameterization}	
\end{figure}

\vspace{-5pt}
\subsection{Object Representation}

In this work, we leverage the cubes and quadrics/cylinders to represent objects, rather than the complex instance-level or category-level model. For objects with regular shapes, such as the book, keyboard, and chair, we use cubes (encoded by their vertices $P_o$) to represent them. For non-regular objects without an explicit direction, such as the ball, bottle, and cup, the quadric/cylinder (encoded by its semiaxis $Q_o$) is used for representation, and its orientation parameter is ignored. Here, $P_o$ and $Q_o$ are expressed in the object frame and only depend on the scale $\boldsymbol{s}$. To register these elements to the global map, we also need to estimate their translation $\boldsymbol{t}$ and orientation $\boldsymbol{\theta}$ w.r.t. the global frame. Cubes and quadrics in the global frame are expressed as follows:
\begin{equation}
\label{cube para}
P_w= R(\boldsymbol{\theta})P_o + \boldsymbol{t},
\end{equation}
\begin{equation}
\label{quadric para}
Q_w= {T}Q_oT^T.
\end{equation}
These two models can be switched conveniently, as shown in fig.\ref{parameterization}(c). Assuming that objects are placed parallel with the ground, as in other works\cite{yang2019monocular, pire2020online} , \textit{i.e.}, $\theta_r\small{=}\theta_p\small{=}0$, we only need to estimate $[\theta_y, \boldsymbol{t}, \boldsymbol{s}]$ for a cube and $[\boldsymbol{t}, \boldsymbol{s}]$ for a quadric.

\vspace{-5pt}
\subsection{Estimation of translation($\boldsymbol{t}$) and scale($\boldsymbol{s}$)}

Assuming that the object point clouds $X$ are in the global frame, we follow conventions and denote its mean by $\boldsymbol{t}$, based on which the scale can be calculated by $\boldsymbol{s}=(\max(X)-\min(X))/2$, as shown in Fig.\ref{parameterization}(a). The main challenge here is that $X$ is typical with many outliers, which will introduce a substantial bias to $\boldsymbol{t}$ and $\boldsymbol{s}$. One of our major contributions in this paper is the development of an outlier-robust centroid and scale estimation algorithm based on the iForest \cite{liu2012isolation} to improve the estimation accuracy. The detailed procedure of our algorithm is presented in Alg. \ref{algorithm_isolation}.

The key idea of the algorithm is to recursively separate the data space into a series of isolated data points, and then take the easily isolated ones as outliers. The philosophy is that, normal points are typically located more closely and thus need more steps to isolate, while the outliers usually scatter sparsely and can be easily isolated with fewer steps. 
As indicated by the algorithm, we first create $t$ isolated trees (the iForest) using the point cloud of an object (lines 2 and 14-33), and then identify the outliers by counting the path length of each point $\mathbf{x}\in X$ (lines 3-9), in which the score function is defined as follows:
\begin{equation}
\label{score}
\begin{split}
s(\mathbf{x})&=2\exp{\frac{-E(h(\mathbf{x}))}{C}},
\end{split}
\end{equation}
\begin{equation}
\label{ptlen}
\begin{split}
C=2 &H(|X|-1)-\frac{2(|X|-1)}{|X|},
\end{split}
\vspace*{-0.4\baselineskip}
\end{equation}
where $C$ is a normalization parameter, $H$ is a harmonic number $H(i) = \ln{(i)} + 0.5772156649$, $h(\mathbf{x})$ is the height of point $\mathbf{x}$ in the isolated tree, and $E$ is the operation to calculate the average height. As demonstrated in Fig. \ref{parameterization}(d)-(e), the yellow point is isolated after four steps; hence its path length is 4, whereas the green point has a path length of 8. Therefore, the yellow point is more likely to be an outlier.
In our implementation, points with a score greater than 0.6 are removed and the remaining are used to calculate $\boldsymbol{t}$ and $\boldsymbol{s}$ (lines 10-12). Based on $\boldsymbol{s}$, we can initially construct the cubics and quadratics in the object frame, as shown in Fig. \ref{parameterization}(a)-(c). 

\renewcommand{\algorithmicrequire}{\textbf{Input:}} 
\renewcommand{\algorithmicensure}{\textbf{Output:}}
\begin{algorithm}[t]
	\caption{Centroid and Scale Estimation Based on iForest}
	\label{algorithm_isolation}
	\begin{algorithmic}[1] 
		\Require $X$ - The point cloud of an object, $t$ - The number of iTrees in iForest, $\psi$ - The subsampling size for an iTree.
		\Ensure $\mathcal{F}$ - The iForest, a set of iTrees, $\boldsymbol{t}$ - The origin of local frame, $\boldsymbol{s}$ - The initial scale of the object. 
		\vspace{1mm}		
		\Procedure{paraObject}{$X, t, \psi$}
		\State $\mathcal{F} \gets$ \Call{buildForest}{$X, t, \psi$}
		\For{point $\mathbf{x}$ in $X$}
		\State $E(h) \gets$ averageDepth($\mathbf{x},\mathcal{F}$) 
		\State $s \gets $ score($E(h), C$) \Comment Eq. \eqref{score} and \eqref{ptlen}		
		\If{$s > 0.6$} \Comment an empirical value
		\State remove($\mathbf{x}$) \Comment remove $\mathbf{x}$ from $X$
		\EndIf
		\EndFor
		\State $\boldsymbol{t} \gets$ meanValue($X$)
		\State $\boldsymbol{s} \gets$ (max($X$) - min($X$)) / 2
		\State \Return $\mathcal{F}, \boldsymbol{t}, \boldsymbol{s}$
		\EndProcedure
		\vspace{1mm}
		
		\Procedure{buildForest}{$X, t, \psi$}
		\State $\mathcal{F} \gets \phi$
		\State $l \gets $ ceiling($\log_2\psi$) \Comment maximum times of iterations
		\For{$i=$ 1 to $t$}
		\State $X^{ (i)}\gets$ randomSample($X, \psi$)
		\State $\mathcal{F} \gets \mathcal{F} \;\cup$ \Call{buildTree}{$X^{(i)}, 0, l$}
		\EndFor \\
		\Return $\mathcal{F}$
		\EndProcedure
		\vspace{1mm}
		
		\Procedure{buildTree}{$X, e, l$}
		\If{$e\geq l$ or $|X| \leq 1$}
		\State \Return exNode\{$|X|$\} \Comment record the size of $X$
		\EndIf
		\State $i \gets$ randomDim(1, 3) \Comment get one dimension
		\State $q \gets$ randomSpitPoint($X[i]$)
		\State $X_l, X_r \gets$ split($X[i], q$)
		\State $L\gets$\Call{buildTree}{$X_l, e+1, l$} \Comment get child pointer
		\State $R\gets$\Call{buildTree}{$X_r, e+1, l$}
		\State \Return inNode\{$L, R, i, q$\}
		\EndProcedure
	\end{algorithmic}
\end{algorithm}

\vspace{-5pt}
\subsection{Estimation of orientation($\theta_y$)}

\begin{figure}[t]
	\centering
    \setlength{\abovecaptionskip}{2pt}
	\captionsetup{belowskip=-6pt}
	\includegraphics[scale=0.45]{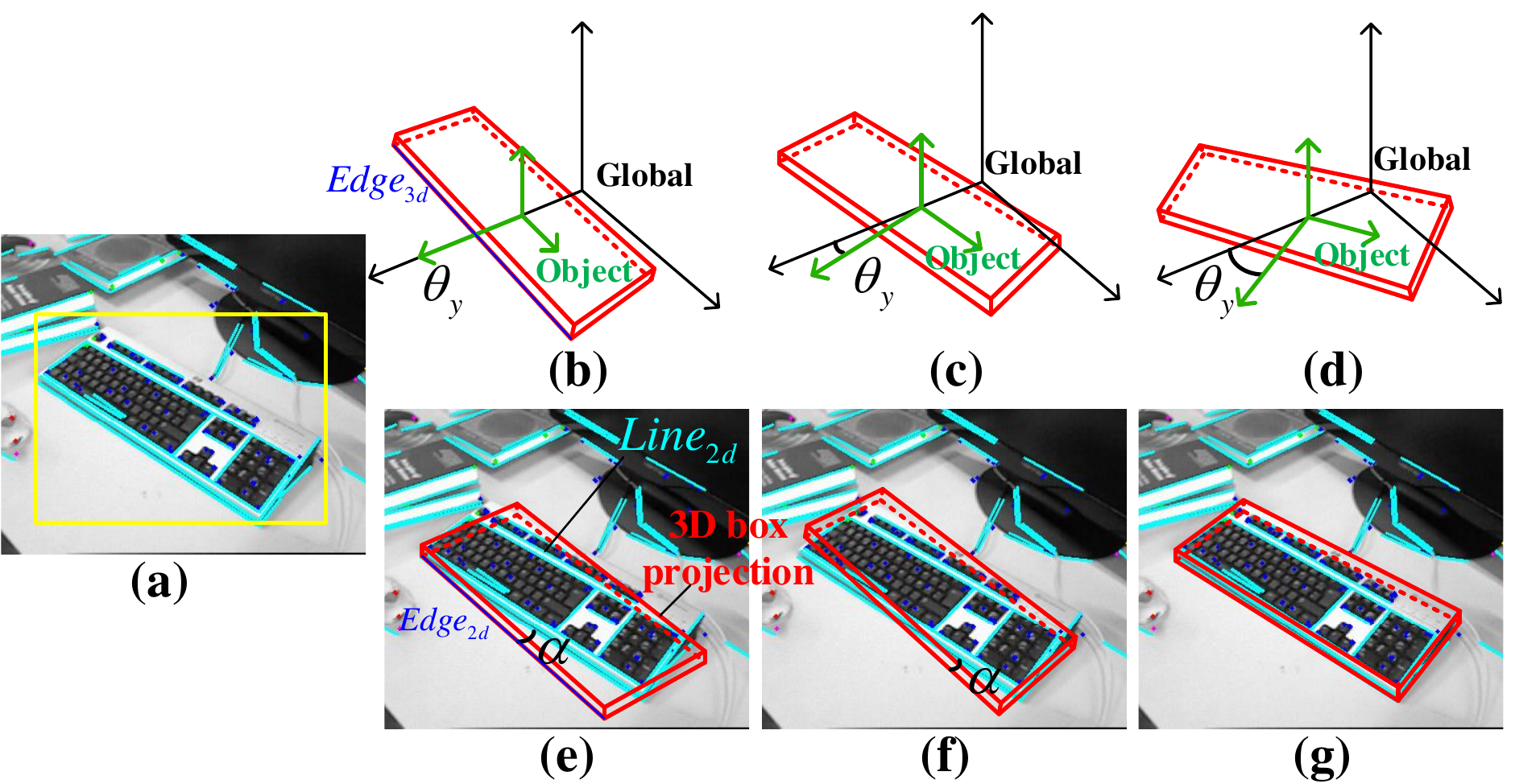}
	\caption{Line alignment to initialize object orientation. (a) Object and line detection in 2D image. (b-d) Angle sampling in 3D space; (e-g) Projection of angle sampling process in 2D images.}
	\label{line_alignment}
    \vspace{-10pt}
\end{figure}

The estimation of $\theta_y$ is divided into two steps, namely, to find a good initial value for $\theta_y$ first and then conduct numerical optimization based on the initial value. Since pose estimation is a non-linear process, a good initialization is very important to help improve the optimality of the estimation result. Conventional methods \cite{runz2018maskfusion,li2019semantic} usually neglect the initialization process, which typically yields inaccurate results. 

The detail of orientation initialization algorithm is presented in Alg. \ref{algorithm_orientation}. The inputs are obtained as follows: 1) LSD (Line Segment Detector \cite{von2012lsd}) segments are extracted from $t$ consecutive images, and those falling in the bounding boxes are assigned to the corresponding objects (see Fig. \ref{line_alignment}(a)); 2) The initial pose of an object is assumed to be consistent with the global frame, \textit{i.e.}, $\theta_0\small{=}0$ (see Fig. \ref{line_alignment}b). In the algorithm, we first uniformly sample thirty angles within $[-\pi/2, \pi/2]$ (line 2). For each sample, we then evaluate its score by calculating the accumulated angle errors between LSD segments $Z_{lsd}$ and the projected 2D edges of 3D edges $Z$ of the cube (lines 3-12). The error is defined as follows:  
\begin{equation}
\label{costf}
\begin{aligned}
&e(\boldsymbol{\theta})=||\alpha(\hat{Z}(\boldsymbol{\theta}))-\alpha(Z_{lsd})||^2, \\
&\hat{Z}(\boldsymbol{\theta})=K T_{c}\left(R(\boldsymbol{\theta})Z+\boldsymbol{t}\right).\\
\end{aligned}
\end{equation}
The demonstration of the calculation of $e(\boldsymbol{\theta})$ is visualized in Fig. \ref{line_alignment}(e)-(g).
The score function is defined as follows:
\begin{equation}
\label{cost}
\text { Score }=\frac{N_{\text {p}}}{N_{\text {a }}}(1+0.1(\xi- E(e))),
\end{equation}
where $N_{\text {a}}$ is the total number of line segments of the object in the current frame, $N_{\text {p}}$ is the number of line segments that satisfy $e < \xi$, $\xi$ is a manually defined error threshold (five degrees here), and $E(e)$ is the average error of these line segments with $e < \xi$.
After evaluating all the samples, we choose the one that achieves the highest score as the initial yaw angle for optimization (line 13).
\begin{algorithm}[t]
	\caption{Initialization for Object Pose Estimation}
	\label{algorithm_orientation}
	\begin{algorithmic}[1] 
		\Require $Z_1, Z_2, \dots, Z_t$ - Line segments detected by LSD in $t$ consecutive images, $\theta_0$ - The initial guess of yaw angel.	
		\Ensure  $\theta$ - The estimation result of yaw angel, $e$ - The estimation errors.
		\vspace{1mm}		
		\State $\mathcal{S}, \mathcal{E} \gets \phi$
		\State $ \mathbf{\Theta} \gets$ sampleAngles($\theta_0$, 30) \Comment see Fig. \ref{line_alignment} (b)-(d)
		\For{sample $\theta$ in $\mathbf{\Theta}$} 
		\State $s_{\theta},e_{\theta} \gets 0$
		\For{$Z$ in \{$Z_1, Z_2, \dots, Z_t$\}}
		\State $s, e \gets$ score($\theta, Z$) \Comment Eq. \eqref{costf} and \eqref{cost}
		\State $s_{\theta} \gets s_{\theta} + s $ 
		\State $e_{\theta} \gets e_{\theta} + e $ 
		\EndFor
		\State $\mathcal{S} \gets \mathcal{S} \cup \{s_{\theta}\}$
		\State $\mathcal{E} \gets \mathcal{E} \cup \{e_{\theta}\}$		
		\EndFor	
		\State $\theta^* \gets$ argmax($\mathcal{S}$)
		\State \Return $\theta^*$, $e_{\theta^*}$
	\end{algorithmic}
\end{algorithm}

\subsection{Object pose optimization}
After obtaining the initial $\boldsymbol{s}$ and $\theta_y$, we then jointly optimize object and camera poses:
\begin{equation}
\small
\{O, T_c\}^{*}=\argminB_{\{\theta_y, \boldsymbol{s}\}} \sum\left( {e}(\boldsymbol{\theta}) + {e}(\boldsymbol{s}) \right)+\argminB_{\{T_c\}} \sum e(\boldsymbol{p}),
\end{equation}  
where the first term is the object pose error defined in Eq.~\eqref{costf} and the scale error ${e}(\boldsymbol{s})$ is defined as the distance between the projected edges of a cube and their nearest parallel LSD segments. The second term $e(\boldsymbol{p})$ is the commonly-used reprojection error in the traditional SLAM framework.

\section{Object Descriptor on the Topological Map}
\label{S_TOPO}

After the step of object parameterization, we obtain the label, size, and pose information of a single object. To present the relationship between objects and that between objects and the scene, we create a topological map.  The map is then used to generate an object descriptor for scene matching. 

\vspace{-5pt}
\subsection{Semantic Topological Map}

The topological map is an abstract representation of the scene. In this work, to construct the semantic topological map, the 3D object centroid is used to represent the node $N$ that encodes the semantic label $l$ and the object parameters $t,\theta ,s$. Then, under the distance and number constraints, we generate the undirected edge $E$ between objects, which includes the distance $d$ and angle $\alpha$ of two objects:

\begin{equation}
\label{Node}
N=\left \langle l, t,\theta ,s\right \rangle,
E=\left \langle d, \alpha \right \rangle.
\end{equation}

\begin{figure}[ht]
	\centering
    \setlength{\abovecaptionskip}{2pt}
	\captionsetup{belowskip=-16pt}
	\includegraphics[scale=0.53]{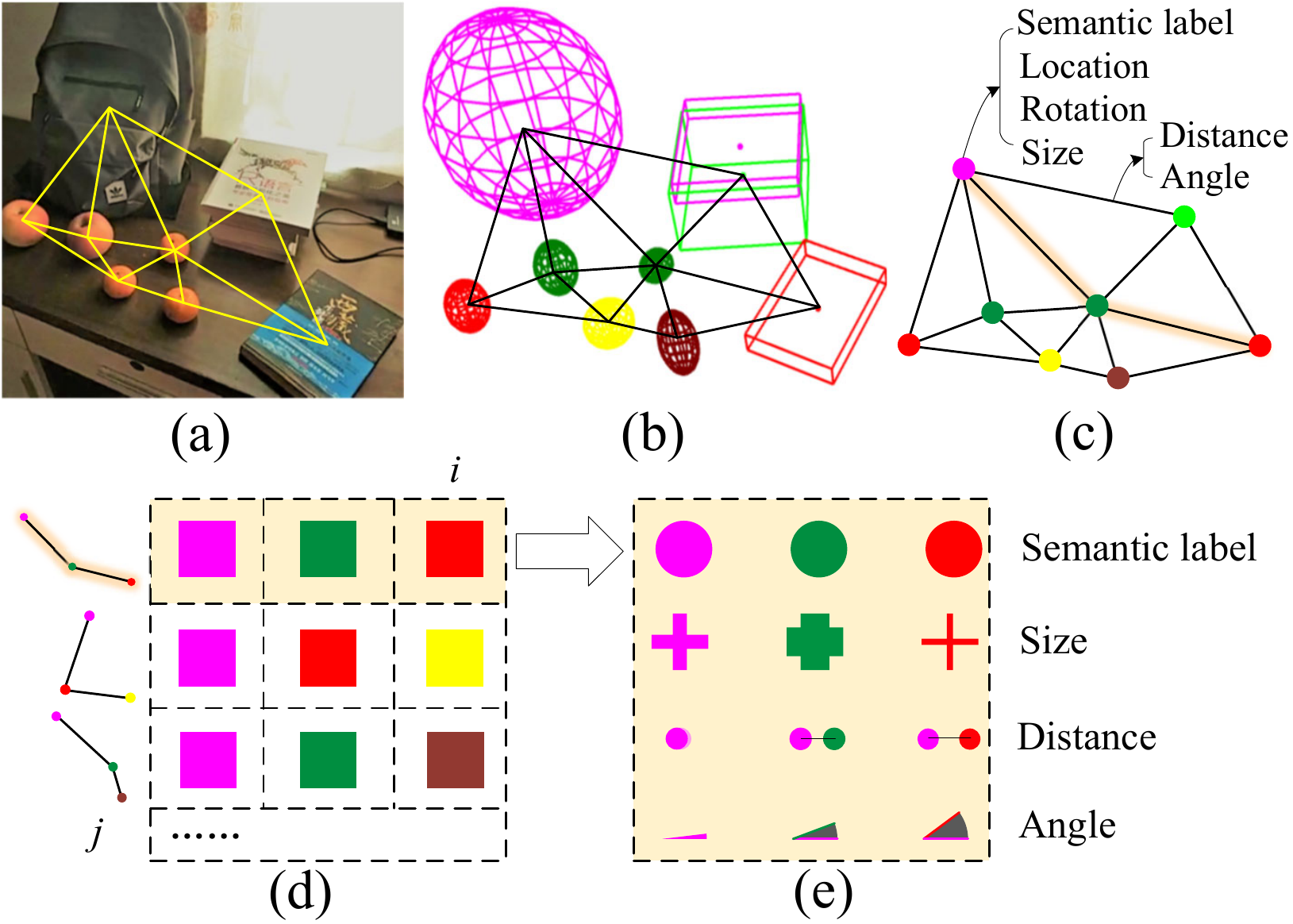}
	\caption{(a) Real-world scene. (b) Object-level map. (c) Semantic topological map. (d) Random walk descriptor. (e) 3D matrix visualization of a single descriptor.}
	\label{topo_map}
\end{figure}

Fig. \ref{topo_map}(a) presents a real-world scene with multiple objects. Fig. \ref{topo_map}(b) shows the object modeling result by the method of Section \ref{S_OP}, which is then used to create a semantic topological map (Fig. \ref{topo_map}(c)) that expresses the scene in an abstract way and shows the connection relationship between objects as symbolized in Eq. (\ref{Node}).

\subsection{Semantic Descriptor}

Since the object information, including semantic label, position, and scale, is not unique, the computation for undirected graph matching, an NP problem \cite{cook1971complexity}, is extremely high. To reduce the computational complexity and enhance the matching accuracy, we introduce a random-walk descriptor that weights multi-neighborhood measurements to describe an object, improving object uniqueness and the relationship with the scene. 

The random-walk descriptor is represented by a 2D matrix, as shown in Fig. \ref{topo_map}(d), with each row storing a walking route that starts at the described object, and randomly points to the next object. It is worth noting that each object only appears once in a route and the process ends when reaching a certain depth $i$ or time $j$ limit.

The previous work \cite{gawel2018x} only considers the semantic label $\mathbf{l} = (l_1,l_2,\cdots,l_i)$ as the descriptor.
Benefiting from the above accurate object parameterization, we add three additional measurements, object size $\mathbf{s} = (s_1,s_2,\cdots,s_i)$, distance $\mathbf{d} = (d_{11},d_{12},\cdots,d_{1i})$, and angle $\mathbf{\alpha } = (\alpha _{11},\alpha _{12},\cdots,\alpha _{1i})$, to improve the robustness of the descriptor. As shown in Fig. \ref{topo_map}(e), thus we transfer the random-walk descriptor to a 3D matrix form:
\begin{equation}
\label{Descriptor}
\upsilon = (r_1, r_2, \cdots, r_j)^{T}, 
r_j = (\mathbf{l},\mathbf{s},\mathbf{d},\mathbf{\alpha} )^{T}.
\end{equation}
In our implementation, the additional measurement does not increase the computation. Instead, it accelerates the matching process by eliminating irrelevant candidates with more clues, such as label and size.

Alg. \ref{algorithm_descriptor} describes the procedure for scene matching. Firstly, each object's semantic descriptor is generated in two independent sub-topological maps (lines 3-4, 10-17). Then find the best matching object-pair by scoring the similarity of each element $(\mathbf{l},\mathbf{s},\mathbf{d},\mathbf{\alpha})$ (lines 5-7, 18-21). Finally, the transformation between two scenes is solved by singular value decomposition (SVD) according to the multiple object pairs (line 8).

There are some points worth mentioning:
\textbf{1) Scale ambiguity}: Two maps are initialized with different depths resulting in distinct scales. While object size, like Li \textit{et al.} \cite{li2020view}, provides a scale by length, width, and height, it is insufficiently robust. Instead, we find the matched object pair between two maps, then calculate the scale factor by averaging the ratio of the distance $d$.
\textbf{2) Anomalous object}: The mismatch resulting from the error object or novel object may cause a considerable inaccuracy in the resolution of the translation; therefore, the RANSAC algorithm is used to eliminate the disturbance caused by anomalous objects.

\renewcommand{\algorithmicrequire}{\textbf{Input:}} 
\renewcommand{\algorithmicensure}{\textbf{Output:}}
\begin{algorithm}[t]
	\caption{Scene matching based on object descriptor}
	\label{algorithm_descriptor}
	\begin{algorithmic}[1] 
		\Require $T_1,T_2$ - Two sub-topo maps, $i,j$ - threshold of depth and number of random-walk.
		\Ensure $\mathbb{T}$ - Transformation between two maps. 
		\vspace{1mm}		
		\Procedure{poseSolve}{$T_1, T_2, i, j$}
		\State $\mathcal{V}_1, \mathcal{V}_2, \mathcal{M} \gets \phi$
		\State $\mathcal{V}_1 \gets$ \Call{objectDescriptor}{$T_1, i, j$}
		\State $\mathcal{V}_2 \gets$ \Call{objectDescriptor}{$T_2, i, j$}
		\For{object $v_1$ in $\mathcal{V}_1$}
		\State $\mathcal{M} \gets \mathcal{M} \;\cup$ \Call{match}{$v_1$, $\mathcal{V}_2$}
		\EndFor \\
		\Return $\mathbb{T} \gets$ SVD($\mathcal{M}$)
		\EndProcedure
        \vspace{1mm}
        
		\Procedure{objectDescriptor}{$T$,$i$,$j$}
		\For{object $o$ in $T$}
		\State $v.row \gets$ random-walk from $o$ to the $i_{th}$ object
		\State $v.col \gets$ repeat random-walk $j$ times
		\State $\mathcal{V} \gets \mathcal{V} \;\cup v$ \Comment Eq. \ref{Descriptor} and Fig. \ref{topo_map}(d,e)
		\EndFor \\
		\Return $\mathcal{V}$
		\EndProcedure
		\vspace{1mm}
		
		\Procedure{match}{$v_1, \mathcal{V}_2$}
		\State $v_2 \gets maxScore(v_1, \mathcal{V}_2)$ \\
		\Return ($v_1, v_2$)
		\EndProcedure
	\end{algorithmic}
\end{algorithm}

\section{Object-Driven Active exploration}
\label{S_ACTIVE}

\begin{figure}[t]
	\centering
    \setlength{\abovecaptionskip}{3pt}
	\captionsetup{belowskip=-10pt}
	\includegraphics[width=0.44\textwidth]{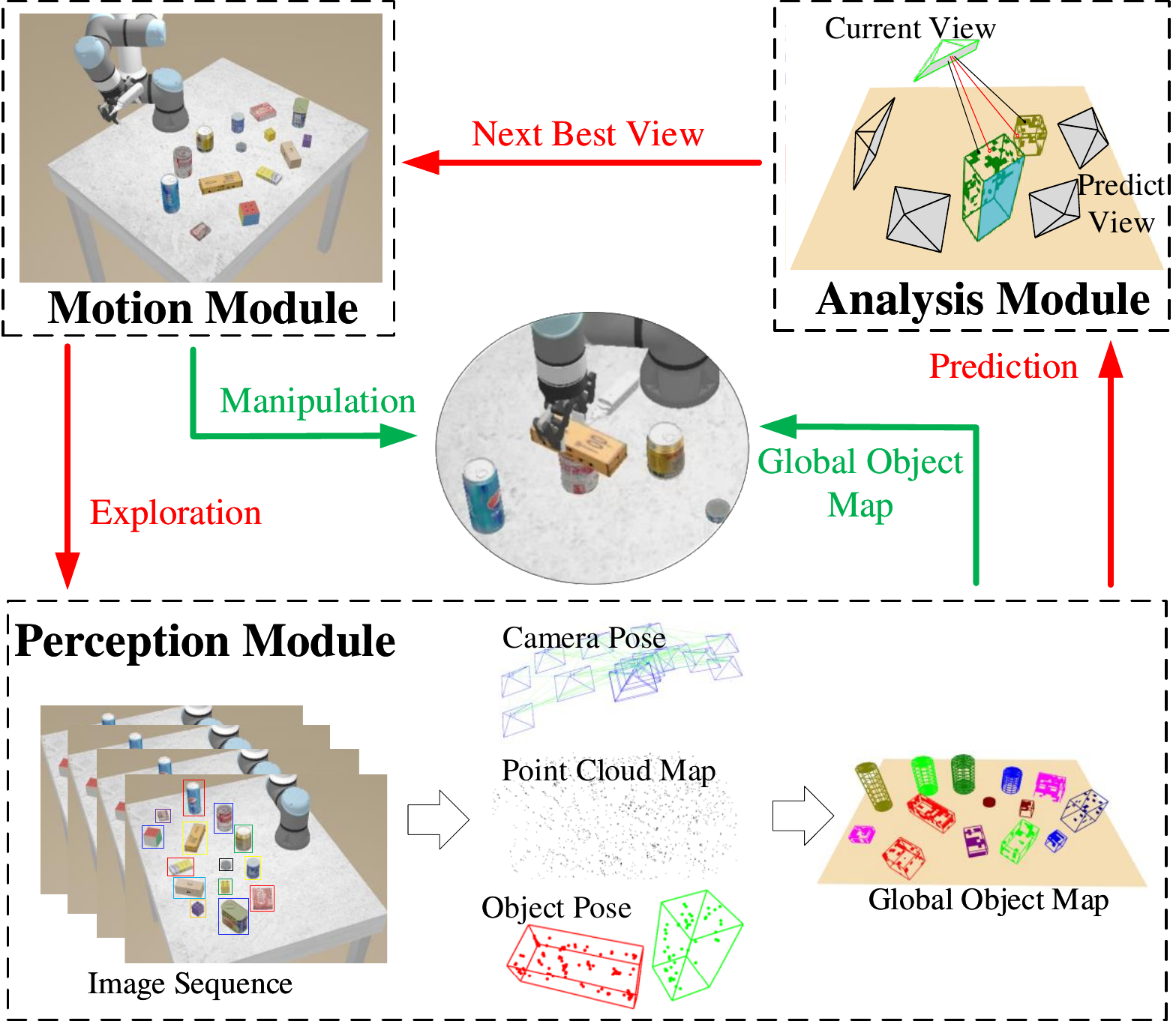}
	\caption{The active mapping framework.}
	\label{Active_Mapping_Framework}
\end{figure}

Object parameterization is good for quantifying the incompleteness of the object or map, and the incompleteness provides a driving force for active exploration. We consider the robotic grasping scene as an example. As shown in Fig. \ref{Active_Mapping_Framework}, the robot arm is fitted with a camera, the motion module controls the robot to execute observation commands. The perception module parametrized the object map by Section \ref{S_DA} and Section \ref{S_OP}. The analysis module measures object uncertainty and predicts different camera views' information gains. The view with the greatest information gain is selected as the Next Best View (NBV) and passed to the motion module to enable active exploration. We aim to incrementally build a global object map with the minimum effort and the maximum accuracy for robotic grasping.

\subsection{Observation Completeness Measurement}
\label{OCM}

\begin{figure}[t]
	\centering
    \setlength{\abovecaptionskip}{3pt}
	\captionsetup{belowskip=-10pt}
	\includegraphics[scale=0.33]{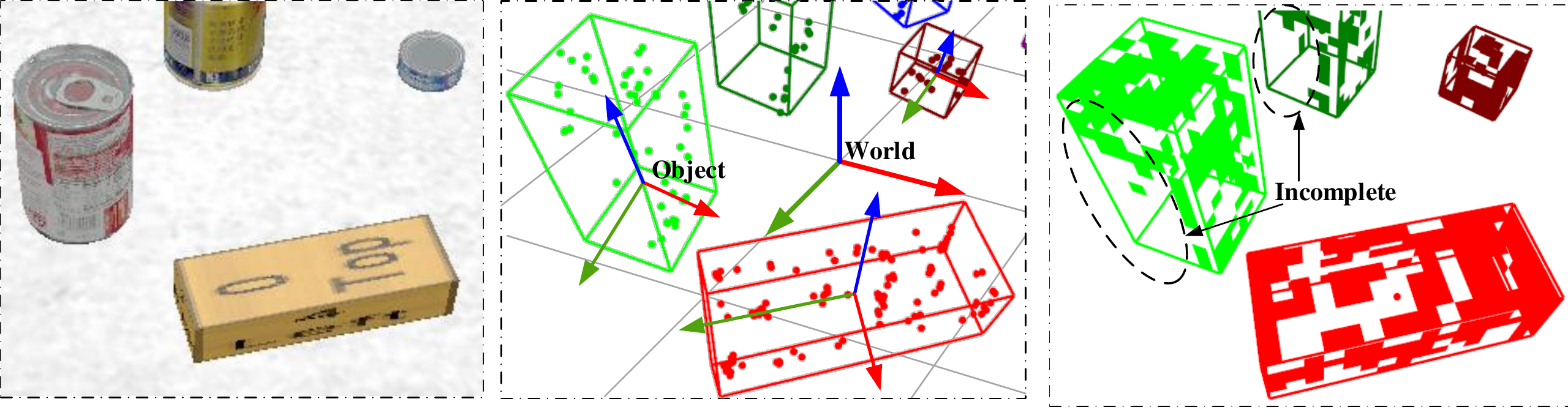}
	\caption{Illustration of observation completeness measurement. {Left: Raw image. Center: Objects with point cloud. Right: Objects with surface grids.}}
	\label{Object Incompleteness}
\end{figure}

We focus on active map building and regard the incompleteness of the map as a motivating factor for active exploration. Existing studies usually take the entire environment as the exploration target \cite{charrow2015information, kahn2015active} or focus on reconstructing a single object \cite{kriegel2015efficient, arruda2016active}, neither of which is ideal for building the object map required by robotic grasping. The reasons are as follows: 1) The insignificant environmental regions will interfere with the decisions made for exploration and misguide the robot into the non-object area; 2) it will significantly increase the computational cost and thus reduce the efficiency of the whole system. We propose an object-driven active exploration strategy for building the object map incrementally. The strategy is designed based on the observation completeness of the object, which is defined as follows.

As demonstrated in Fig. \ref{Object Incompleteness}, the point clouds of an object are translated from the world frame to the object frame and then projected onto the five surfaces of the estimated 3D cube. Here, the bottom face is not considered. 
Each of the five surfaces is discretized into a surface occupancy grid map ~\cite{elfes1989using} with cell size $m \ast m$ ($m = 1cm$ in our implementation).
Each grid cell can be in one of three states:
\begin{itemize}
	\item \textbf{unknown}: the grid is not observed by the camera;
	\item \textbf{occupied}: the grid is occupied by the point clouds;
	\item \textbf{free}: the grid can be seen by the camera but is not occupied by the point clouds.
\end{itemize}

We use information entropy \cite{shannon1948mathematical} to determine the completeness of observations based on the occupancy grid map, as information entropy has the property of symptomatizing uncertainty.
The entropy of each grid cell is defined by a binary entropy function:
\begin{equation}
\label{eq:grid}
H_{gr i d}(p)=-p \log (p)-(1-p) \log (1-p),
\end{equation}
where $p$ is the probability of a grid cell being occupied and its initial value before exploration is set to 0.5. The total entropy is therefore defined as
\begin{equation}
\label{eq:obj}
H_{o b j}=\sum_{o\in\mathbb{O}} H_{o}+\sum_{f \in \mathbb{F}} H_{f}+\sum_{u \in \mathbb{U}} H_{u},
\end{equation}
and the normalized total entropy is
\begin{equation}
\label{eq:ba_grid}
\bar{H}_{obj}=H_{o b j} /(|\mathbb{O}|+|\mathbb{F}|+|\mathbb{U}|),
\end{equation}
where $H_o, H_f, H_u$ are the entropy of occupied, free, and unknown grids, $\mathbb{O}, \mathbb{F}, \mathbb{U}$ are sets of the occupied, free, and unknown grid cells, respectively. $|\mathbb{X}|$ represents the size of $\mathbb{X}$. As objects continue to be explored, the number of unknown grid cells is gradually reduced, making all grids' normalized entropy $\bar{H}_{g ri d}$ a smaller value. The lower the $\bar{H}_{g ri d}$ is, the higher the observation completeness is. The exploration objective is to minimize $\bar{H}_{g ri d}$. 

\vspace{-10pt}
\subsection{Object-Driven Exploration}

\begin{figure}[t]
	\centering
    \setlength{\abovecaptionskip}{0pt}
	\subfigure[Different definitions of information gain in exploration.]{\includegraphics[scale=0.59]{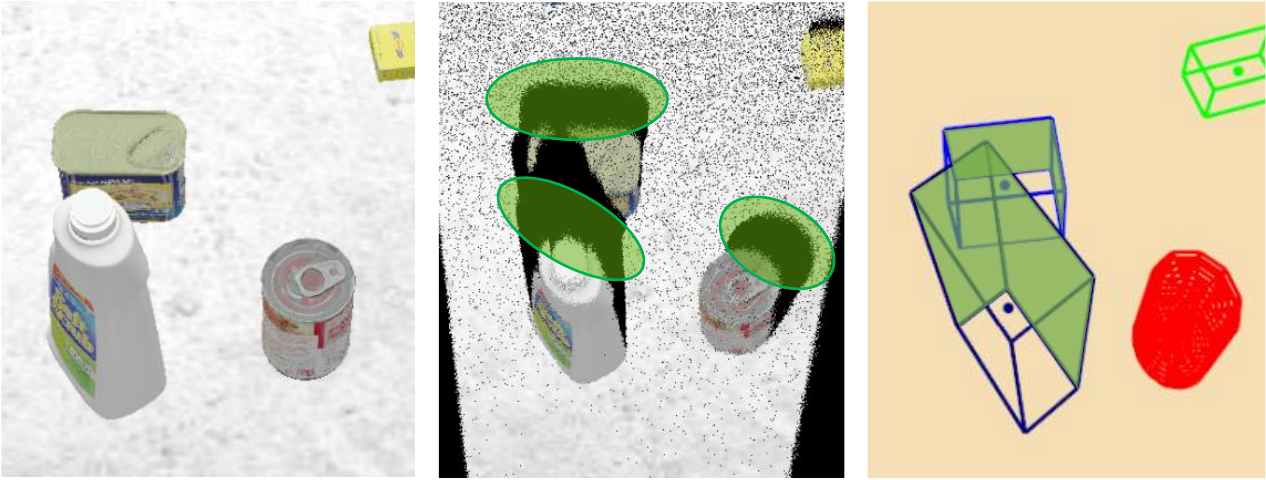}}
	\subfigure[Information gain under different camera views.]
    {\includegraphics[scale=0.45]{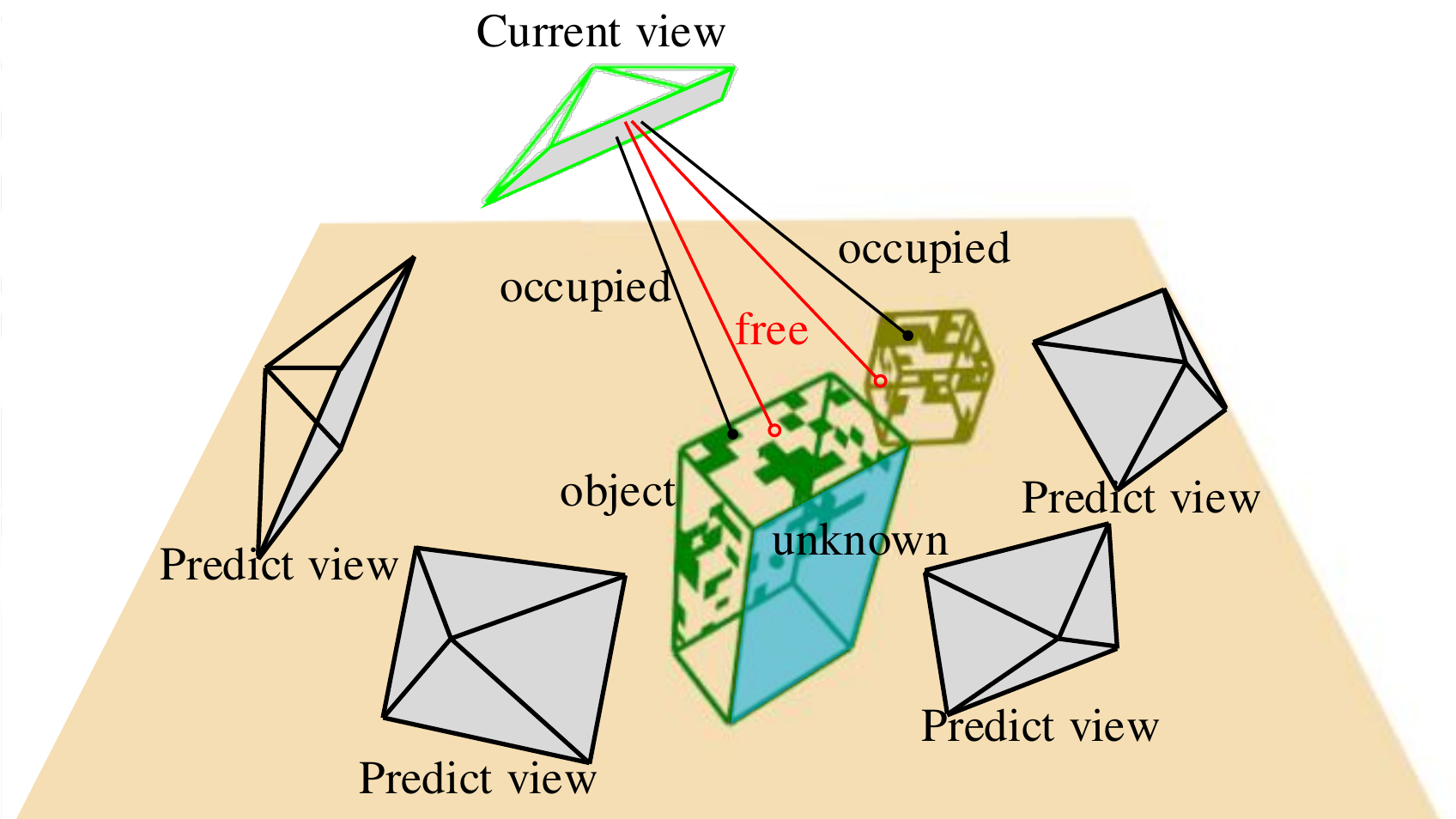}}
	\caption{Demonstration of the object-driven exploration.}
	\label{Perspective Prediction}
	\vspace{-4mm}
\end{figure}

\textbf{Information Gain Definition}: As illustrated in Fig. \ref{Perspective Prediction}(b), object-driven exploration aims to predict the information gain of different candidate camera views and then select the one to explore that maximizes the information gain, \textit{i.e.}, the NBV. The information in this work is defined as the uncertainty of the map, as mentioned in Section \ref{OCM}. The information gain is thus defined as the measurement of uncertainty reduction and accuracy improvement after the camera is placed at a specific pose. Conventionally, information gain is defined based on the area of unknown regions of the environment, \textit{e.g.}, the black holes in the medium subfigure of Fig. \ref{Perspective Prediction}(a), which may mislead the object map building. Compared with the conventional one, our proposed information gain is built on the observation completeness measurement of the object, shown in the right subfigure of Fig. \ref{Perspective Prediction}(a), and incorporates the influence on object pose estimation. 

\textbf{Information Gain Modeling}: As indicated by the definition, information gain is contingent on many factors; thus, we create a utility function to model the information gain by manually designing a feature vector to parameterize those factors. The following is the design of the feature vector used to characterize the object $\mathbf{x}$,
\begin{equation}
\mathbf{x}=\left(H_{obj}, \bar{H}_{obj},  R_{o}, R_{IoU}, \bar{V}_{obj}, s\right),
\end{equation}  
where $H_{obj}$, and $\bar{H}_{obj}$ are defined by Eq. (\ref{eq:grid}) - (\ref{eq:ba_grid}), $R_o$ is the ratio of occupied grids to the total grids of the object, which indicates the richness of its surface texture, $R_{IoU}$ is the 2D mean IoU with adjacent objects used for modeling occlusion under a specific camera view,  $\bar{V}_{obj}$ is the current volume of the object, and $s$ is a binary value used for indicating whether the object is fully explored.

The utility function for NBV selection then is defined as:
\begin{equation}
\label{uti}
f = \sum_{\mathbf{x}\in I}  \left( (1-R_{o})H_{obj}+\lambda(H_{IoU}+H_{V})  \right) s(\mathbf{x}),
\end{equation}
where $I$ is the predicted camera view, $\lambda$ is a weight coefficient ($\lambda=0.2$ in our implementation), and $H_{IoU}, H_{V}$ share the same formula,
\begin{equation}
\label{eq:plogp}
H=- p \log (p).
\end{equation}

The first item $\sum_{\mathbf{x}\in I}(1-R_{o})H_{obj}$ in Eq. \eqref{uti} is used to model the total weighted uncertainty of the object map under the predicted camera view. Here we give more weight to the unknown grids and the free ones by using $1-R_{o}$. The reason is to encourage more explorations in free regions to find more image features that are neglected by previous sensing. 

The second item $\sum_{\mathbf{x}\in I}H_{IoU}$ in Eq. \eqref{uti} defines the uncertainty of object detection, which is one of the critical factors affecting object pose estimation. The uncertainty is essentially caused by occlusions between objects. We use this item to encourage a complete observation of the object. The variable in Eq. \eqref{eq:plogp} is the rescaled 2D IoU, \textit{i.e.}, $p=R_{IoU}/2$.

The third item $\sum_{\mathbf{x}\in I}H_{V}$ in Eq. \eqref{uti} models the uncertainty of object pose estimation. Under different camera views, the estimated object poses are usually different and induce the changes in object volume. Here, we first fit a standard normal distribution using the normalized history volumes $\{\bar{V}_{obj}^{(0)}, \bar{V}_{obj}^{(1)}, \cdots, \bar{V}_{obj}^{(t)}\}$ of each object, and then take the probability density of $\bar{V}_{obj}^{(t)}$ as the value $p$ in Eq. \eqref{eq:plogp}. This item essentially encourages the camera view that can converge the pose estimation process.

The $s(\mathbf{x})$ in Eq. \eqref{uti} indicates whether the object should be considered during the calculation of the utility function. Set $s(\mathbf{x})\small{=}0$, if the following condition is satisfied: $(\bar{H}_{grid} < 0.5 \lor R_o > 0.5)  \land p(\bar{V}_{obj}^{(t)}) > 0.8$. If this condition holds for all the objects, or the maximum tries are achieved (10 in this work), the exploration will be finished.

Based on the utility function, the NBV that maximizes $f$ is continuously selected and leveraged to guide the exploration process, during which the global object map is also incrementally constructed, as depicted in Fig. \ref{Active_Mapping_Framework}.

\section{Experiment}
\label{S_EX}

The experiment will demonstrate the performance of essential techniques such as data association, object parameterization, and active exploration. In addition, the proposed object SLAM framework will be evaluated by various applications, such as object mapping, augmented reality, scene matching, relocalization, and robotic grasping, .

\begin{figure}[t]
	\centering
        \setlength{\abovecaptionskip}{0pt}
	\subfigure[Position distribution of point clouds in three directions.]{\includegraphics[scale=0.25]{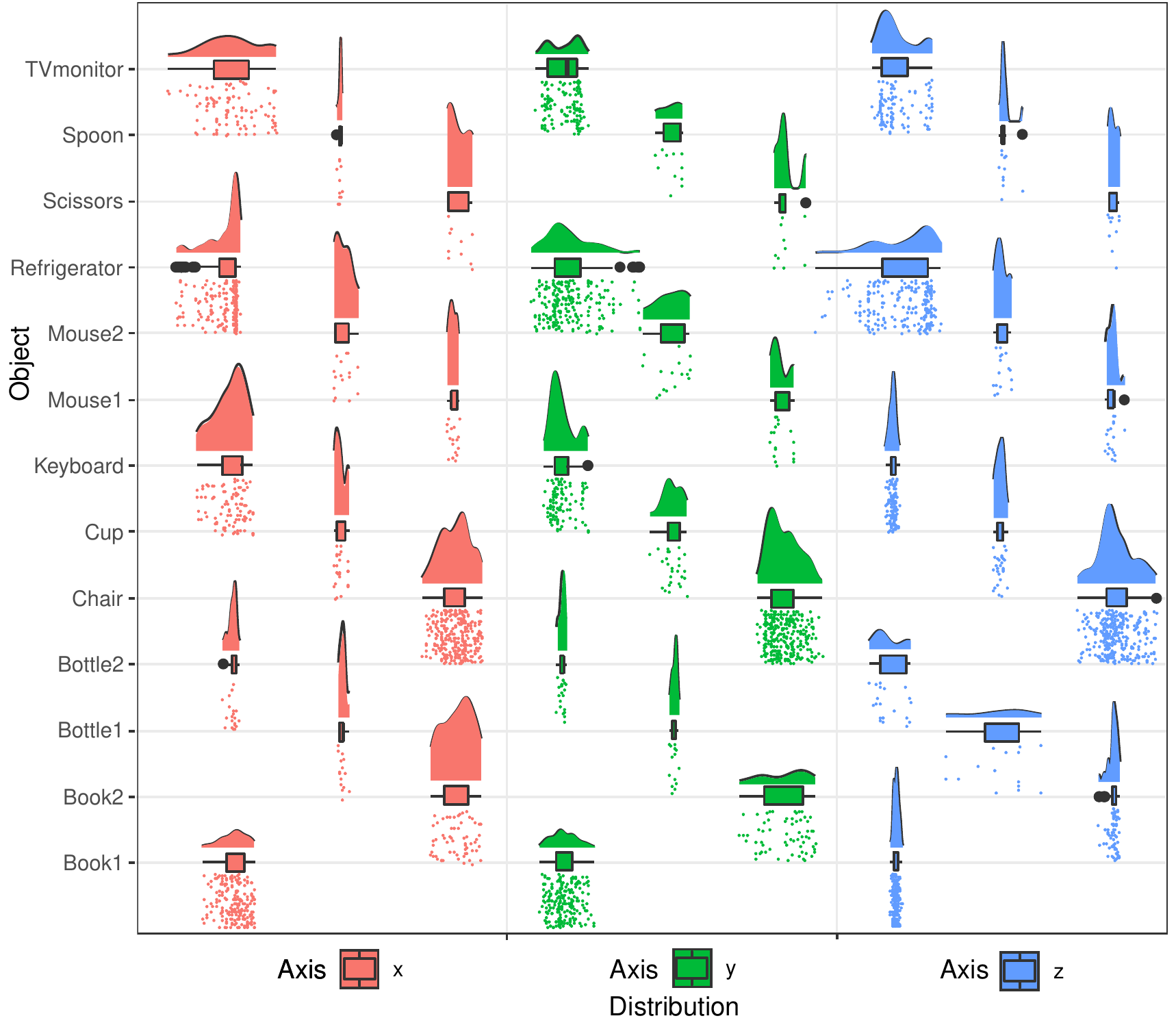}}
	\subfigure[Distance error distribution of centroids.]{\includegraphics[scale=0.29]{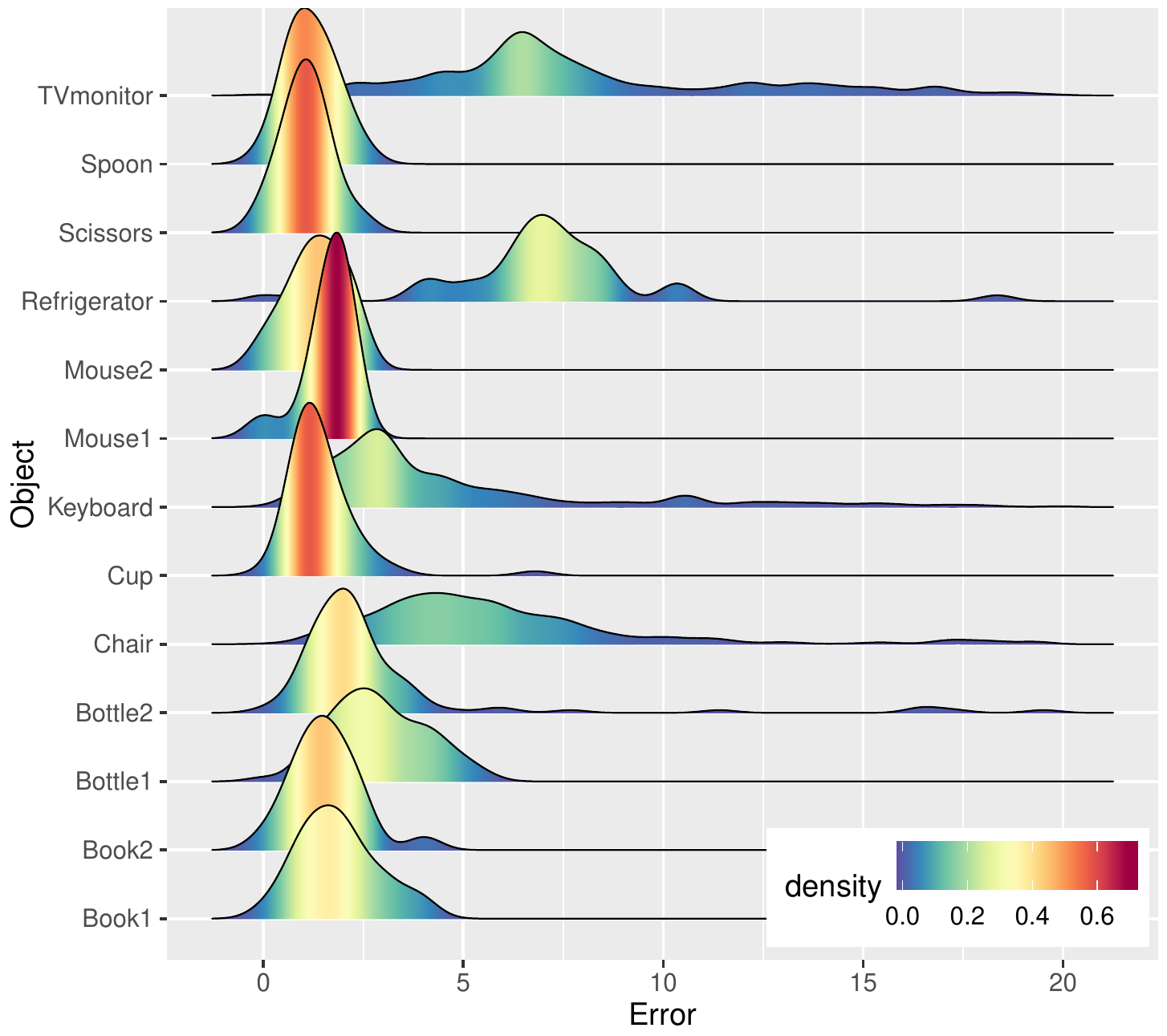}}
	\caption{Distributions of different statistics in data association.}
	\label{Distribution}
 \vspace{-10pt}
\end{figure}

\begin{figure}[t]
\centering
\includegraphics[scale=0.40]{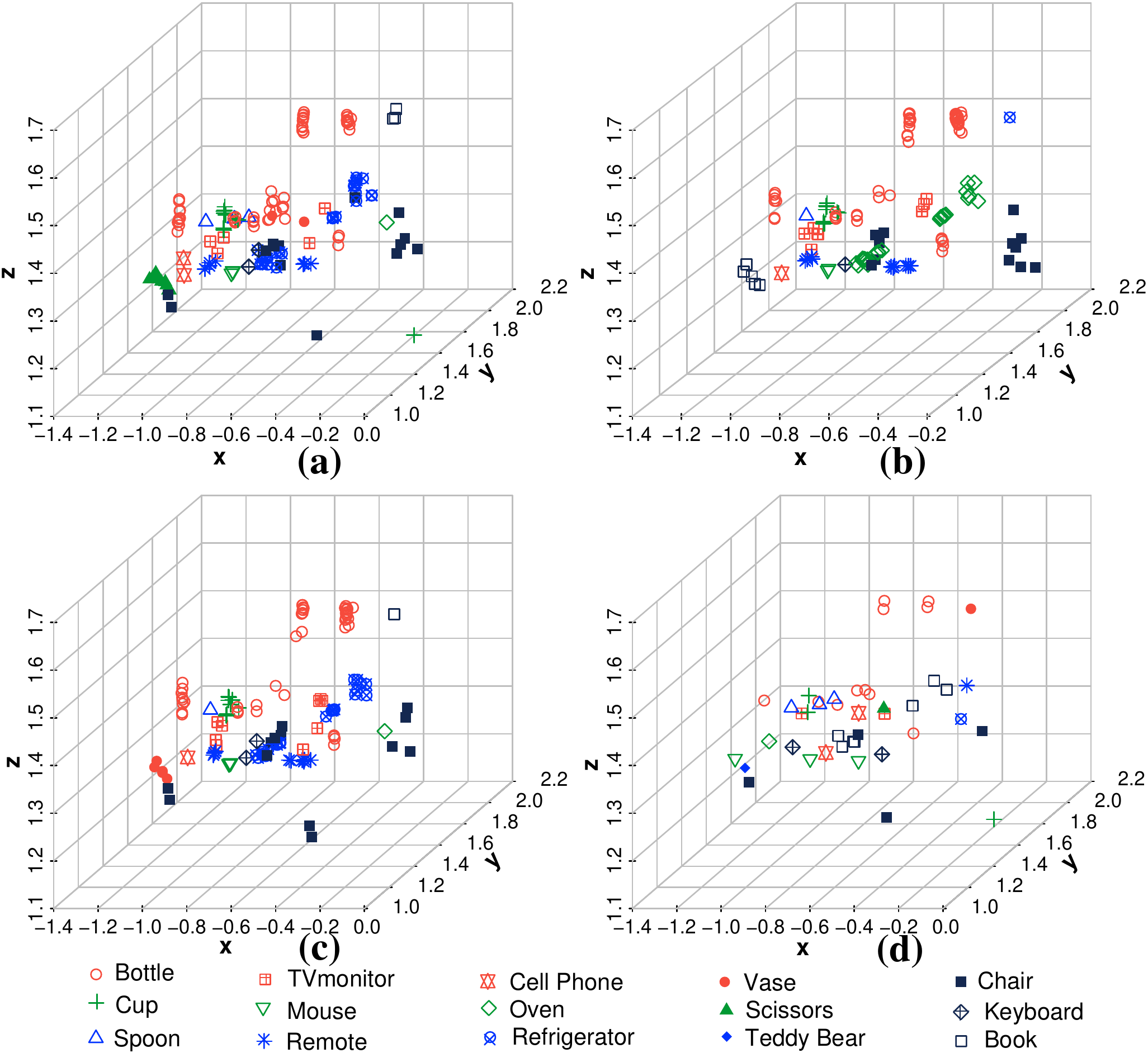}
\caption{Qualitative comparison of data association results. (a) IoU method. (b) IoU and nonparametric test. (c) IoU and t-test. (d) our ensemble method. }
\label{DataAssociationResults}
\vspace{-15pt}
\end{figure}

\vspace{-12pt}
\subsection{Distributions of Different Statistics}
\label{SEXDDS}

For data association, the adopted 3D statistics for statistical testing include the point clouds and their centroids of an object. To verify our hypothesis about the distributions of different statistics, we analyze a large amount of data and visualize their distributions in Fig. \ref{Distribution}.

Fig. \ref{Distribution} (a) shows the distributions of the point clouds from 13 objects during the data association in the TUM RGB-D fr3\_long\_office sequence \cite{sturm2012benchmark}. Obviously, these statistics do not follow a Gaussian distribution. The distributions are related to specific characteristics of the objects, and do not show consistent behaviors. 
Fig.~\ref{Distribution} (b) shows the error distribution of object centroids, which typically follow the Gaussian distribution. This error is computed between the centroids of objects detected in each frame and the object centroid in the final, well-constructed map.
This result verifies the reasonability of applying the nonparametric \textit{Wilcoxon Rank-Sum test} for point clouds and the t-test for object centroids.

\vspace{-10pt}
\subsection{Object-level Data Association Experiments}
We compare our method with the commonly-used Intersection over Union (IoU) method, nonparametric test (NP), and t-test. Fig. \ref{DataAssociationResults} shows the association results of these methods in the TUM RGB-D fr3\_long\_office sequence. It can be seen that some objects are not correctly associated in (a)-(c). Due to the lack of association information, existing objects are often misrecognized as new ones by these methods once the objects are occluded or disappear in some frames, resulting in many unassociated objects in the map. In contrast, our method is much more robust and can effectively address this problem (see Fig. \ref{DataAssociationResults}(d)). The results of other sequences are shown in Table \ref{tab1}, and we use the same evaluation metric as \cite{iqbal2018localization,iqbal2020data}, which measures the number of objects that are finally present in the map. The \textit{GT} represents the ground-truth object number. As we can see, our method achieves a high success rate of association, and the number of objects in the map goes closer to GT, which significantly demonstrates the effectiveness of the proposed method.

The results of our comparison with \cite{iqbal2018localization,iqbal2020data}, which is based on the nonparametric test, are reported in \ref{tab2}. As indicated, our method can significantly outperform \cite{iqbal2018localization,iqbal2020data}. Especially in the TUM dataset, the number of successfully associated objects by our method is almost twice that by \cite{iqbal2018localization,iqbal2020data}. The advantage in Microsoft RGBD \cite{shotton2013scene} and Scenes V2 \cite{lai2014unsupervised} is not apparent since the number of objects is limited.
Reasons for the inaccurate association of \cite{iqbal2018localization,iqbal2020data} lie in two folds: 1) The method does not exploit different statistics and only uses non-parametric statistics, thus resulting in many unassociated objects; 2) A clustering algorithm is leveraged to tackle the abovementioned problem, but it removes most of the candidate objects.

\vspace{-3pt}
\begin{table}[!t]
\renewcommand{\arraystretch}{1.1}
\setlength{\tabcolsep}{8pt}
\setlength{\abovecaptionskip}{-5pt}
\caption{DATA ASSOCIATION RESULTS}
\begin{center}
\label{tab1}
\begin{tabular}{c ccccc}
\toprule
            & IoU & IoU+NP & IoU+t-test & \textbf{Ours} & GT \\ \midrule
Fr1\_desk   & 62  & 47     & 41         & \textbf{14}   & 16 \\
Fr2\_desk   & 83  & 64     & 52         & \textbf{22}   & 25 \\
Fr3\_office & 150 & 128    & 130        & \textbf{42}   & 45 \\
Fr3\_teddy  & 32  & 17     & 21         & \textbf{6}    & 7  \\ \bottomrule
\end{tabular}
\end{center}
 \vspace{-20pt}
\end{table}

\begin{table*}[!h]
\setlength{\abovecaptionskip}{-5pt}
\caption{QUANTITATIVELY ANALYZED DATA ASSOCIATIONS}
\begin{center}
\label{tab2}
\renewcommand{\arraystretch}{1.1}
\begin{tabular}{c cccc ccccc lllll}
\toprule
\multirow{2}{*}{Seq} & \multicolumn{4}{c}{TUM}                                   & \multicolumn{5}{c}{Microsoft RGBD}                               & \multicolumn{5}{c}{Scenes V2}                                  \\ \cmidrule(r){2-5} \cmidrule(r){6-10} \cmidrule(r){11-15}
                     & fr1\_desk   & fr2\_desk   & fr3\_long\_office & fr3\_teddy & Chess       & Fire       & Office      & Pumpkin    & Heads       & 01         & 07         & 10         & 13         & 14         \\ \midrule
\cite{iqbal2018localization,iqbal2020data}    & -           & 11          & 15                & 2          & 5           & 4          & 10          & 4          & -           & 5          & -          & 6          & \textbf{3} & 4          \\
\textbf{Ours}        & \textbf{14} & \textbf{22} & \textbf{42}       & \textbf{6} & \textbf{13} & \textbf{6} & \textbf{21} & \textbf{6} & \textbf{15} & \textbf{7} & \textbf{7} & \textbf{7} & \textbf{3} & \textbf{5} \\
GT                   & 16          & 23          & 45                & 7          & 16          & 6          & 27          & 6          & 18          & 8          & 7          & 7          & 3          & 6          \\ \bottomrule
\end{tabular}
\end{center}
\vspace{-20pt}
\end{table*}

\vspace{-10pt}
\subsection{Qualitative Assessment of Object Parameterization}

\begin{figure}[htbp]
\vspace{-10pt}
	\centering
	\includegraphics[scale=0.47]{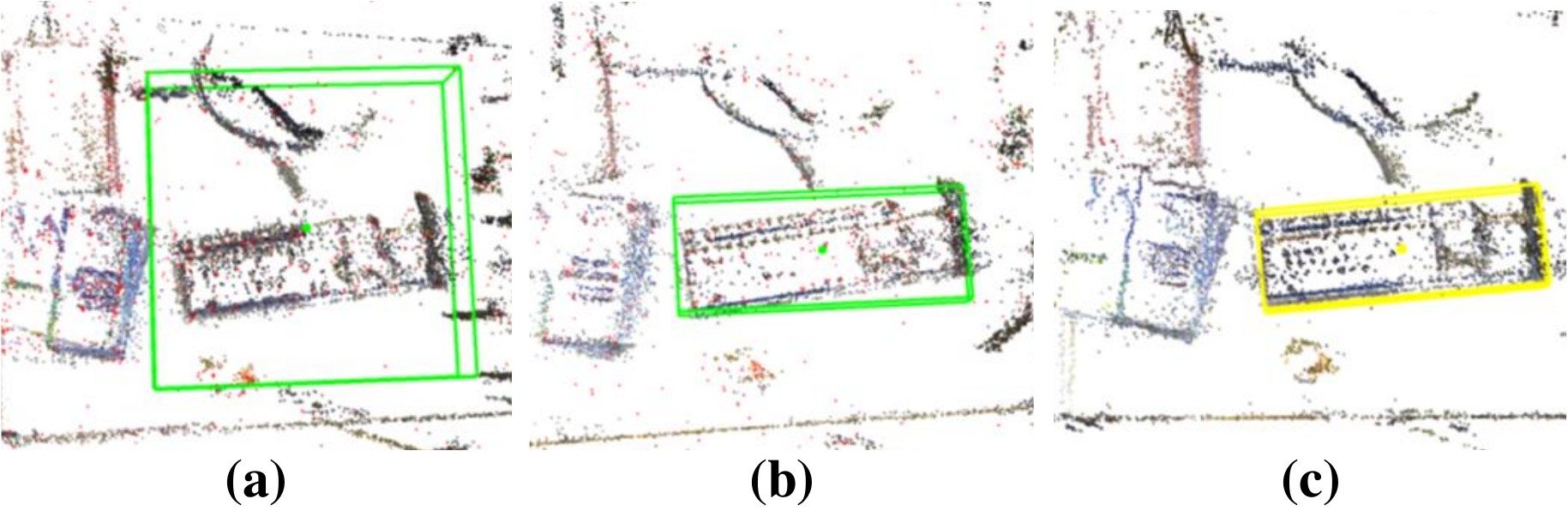}
	\caption{Visualization of the pose estimation. (a): Initial object pose and size. (b): Object pose and size after iForest. (c): object pose and size after iForest and line alignment.}
	\label{KeyboardResult}
 \vspace{-20pt}
\end{figure}

\begin{figure}[htbp]
	\centering
	\captionsetup{belowskip=-10pt}
	\includegraphics[scale=0.11]{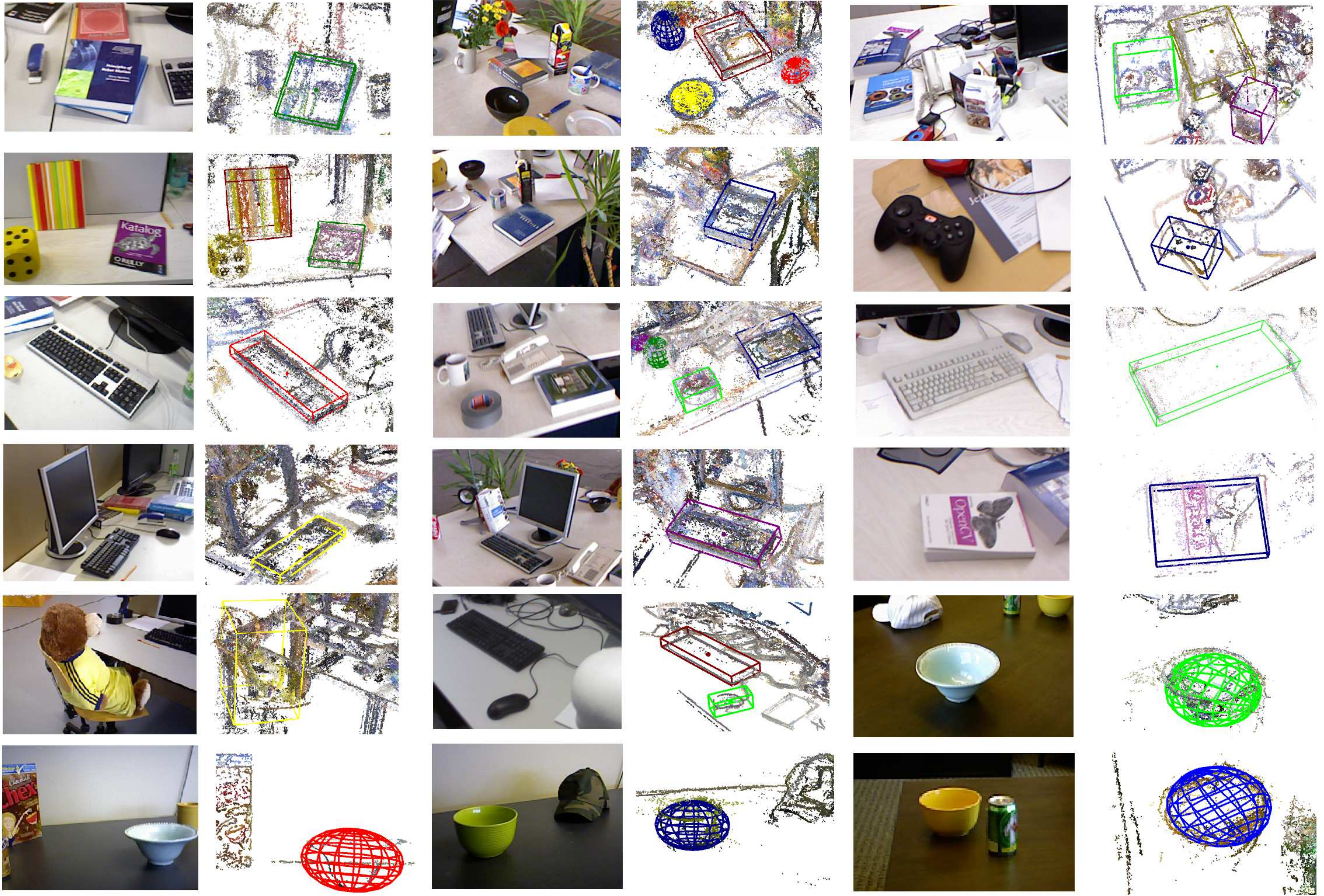}
	\caption{Results of object pose estimation. Odd columns: original RGB images. Even column: estimated object poses.}
	\label{ObjectPose}
 \vspace{-10pt}
\end{figure}

To demonstrate the accuracy of object parameterization, We superimpose the cubes and quadrics of objects on semi-dense maps for qualitative evaluation. Fig. \ref{KeyboardResult} is the 3D top view of a keyboard (Fig. \ref{line_alignment}(a)) where the cube characterizes its pose. Fig. \ref{KeyboardResult}(a) is the initial pose with large-scale error; Fig. \ref{KeyboardResult}(b) is the result after using iForest; Fig. \ref{KeyboardResult}(c) is the final pose after our joint pose estimation. Fig. \ref{ObjectPose} presents the pose estimation results of the objects in 14 sequences of the three datasets, in which the objects are placed randomly and in different directions. As is shown, the proposed method achieves promising results with a monocular camera, which demonstrates the effectiveness of our pose estimation algorithm.

\subsection{Object-Oriented Map Building}

\begin{figure*}[!htbp]
\centering
\setlength{\abovecaptionskip}{3pt}
\includegraphics[scale=0.085]{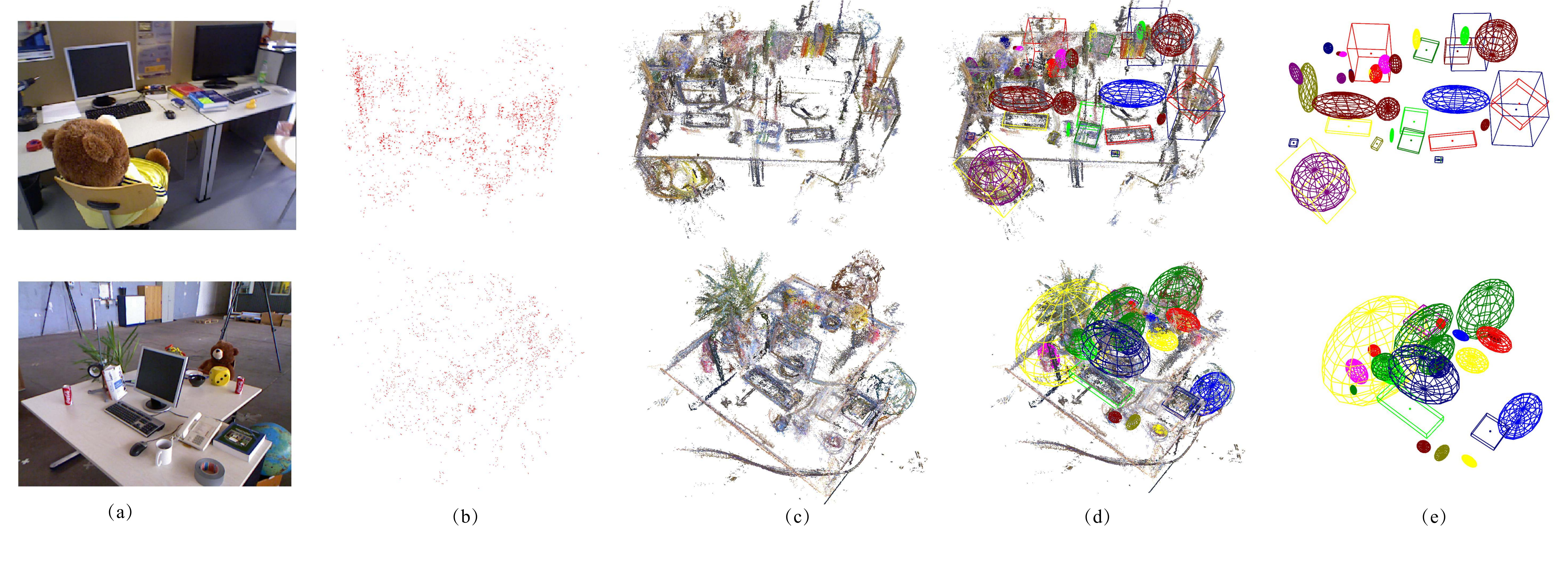}
\caption{Different map representations. (a) the RGB images. (b) the sparse map. (c) semi-dense map. (d) our semi-dense semantic map. (e) our lightweight and object-oriented map. (d) and (e) are build by the proposed method.}
\label{TUM_Mapping}
\end{figure*}

Then, we build the object-oriented semantic maps based on the robust data association algorithm, the accurate object pose estimation algorithm and a semi-dense mapping system \cite{he2018incremental}. Fig. \ref{TUM_Mapping} shows three examples of TUM fr3\_long\_office and fr2\_desk, where (d) and (e) show semi-dense semantic and object-oriented maps built by our object SLAM. Compared with the sparse map of ORB-SLAM2, our maps can express the environment much better. Moreover, the object-oriented map shows superior performance in environment understanding than the semi-dense map.

\begin{figure*}[!htbp]
\centering
\setlength{\abovecaptionskip}{0pt}
\captionsetup{belowskip=-10pt}
\includegraphics[scale=0.085]{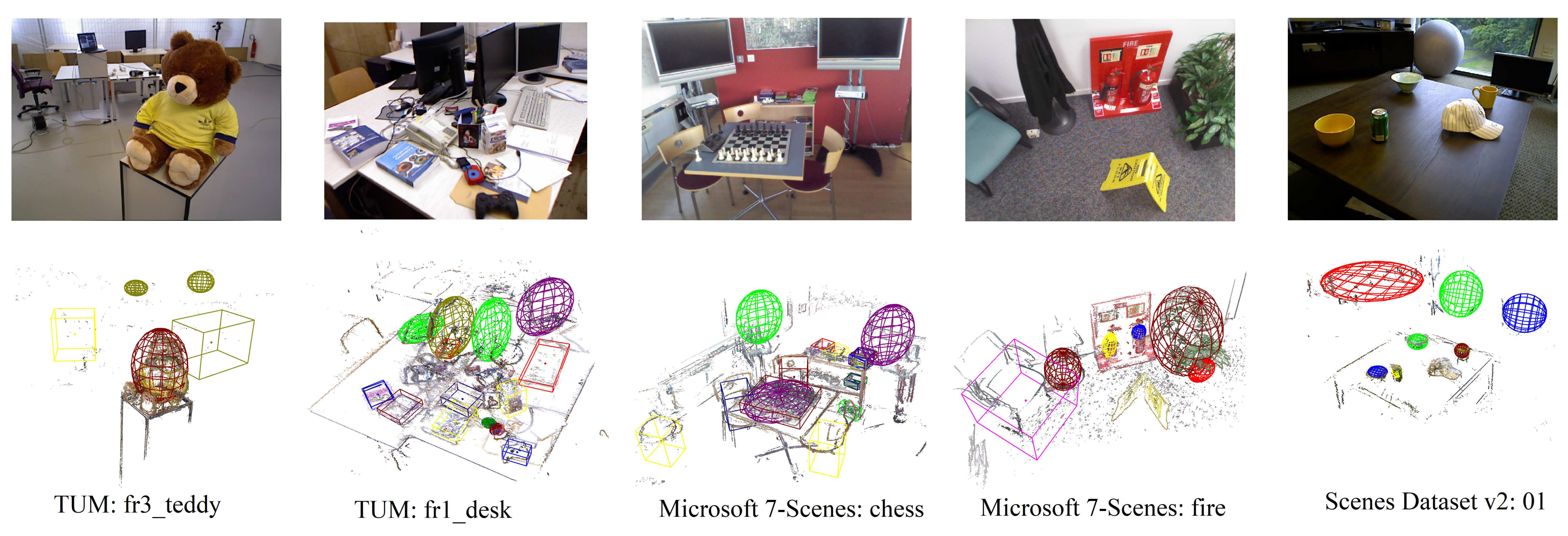}
\caption{Mapping results on the three datasets. Top: raw images. Bottom: simi-dense object-oriented map.}
\label{ThreeDatasets}
\end{figure*}

\begin{figure}[!htbp]
\centering
\setlength{\abovecaptionskip}{3pt}
\captionsetup{belowskip=-10pt}
\includegraphics[scale=0.38]{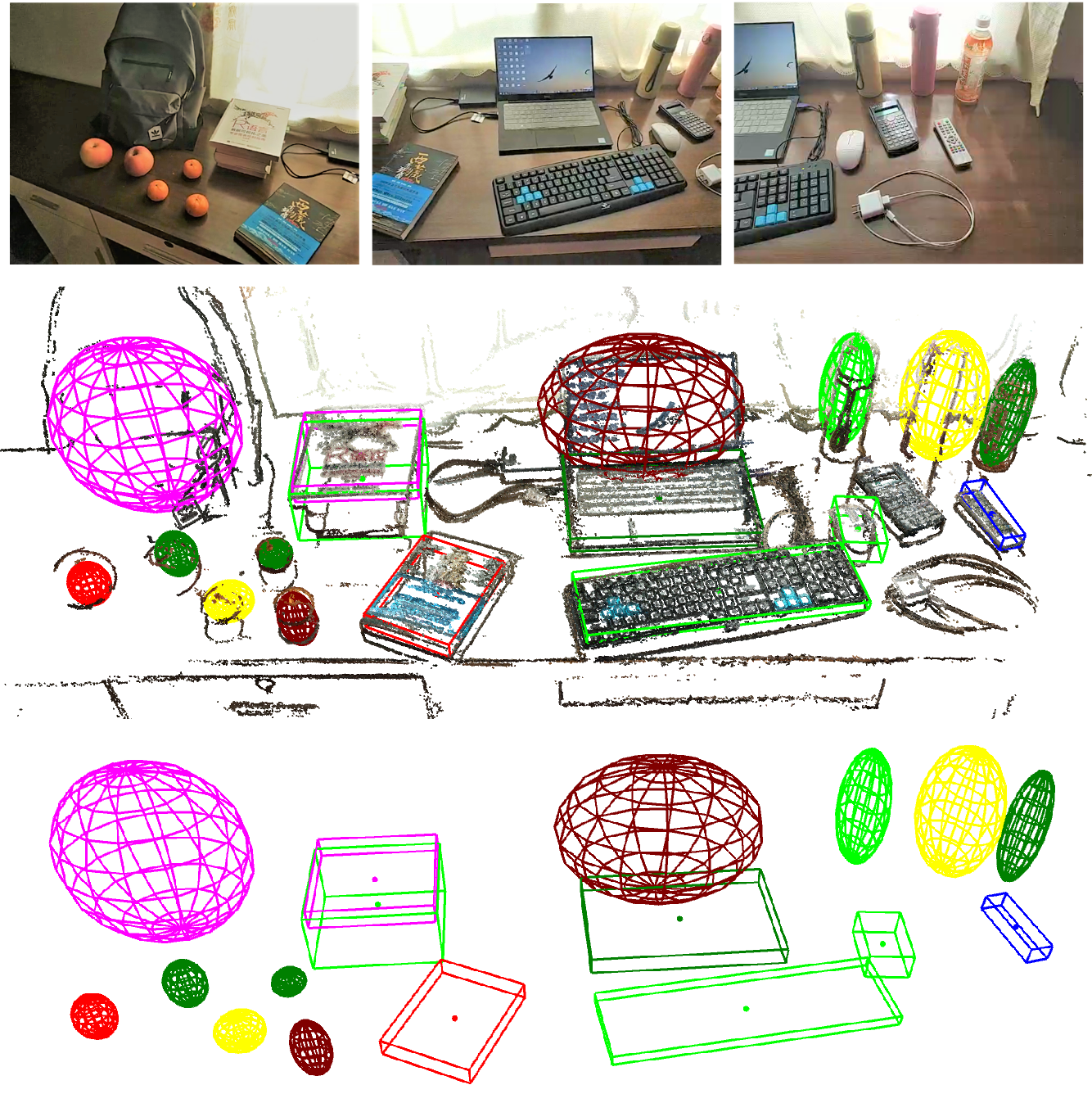}
\caption{Mapping results in a real scenario. Top: raw images. Middle: semi-dense object-oriented map. Bottom: lightweight and object-oriented map.}
\label{RealScene}
\end{figure}

The mapping results of other sequences in TUM, Microsoft RGB-D, and Scenes V2 datasets are shown in Fig. \ref{ThreeDatasets}. It can be seen that the system can process multiple classes of objects with different scales and orientations in complex environments. Inevitably, there are some inaccurate estimations. For instance, in the \textit{fire} sequence, the chair is too large to be well observed by the fast-moving camera, thus yielding an inaccurate estimation. We also conduct the experiment in a real scenario (Fig. \ref{RealScene}). It can be seen that even if the objects are occluded, they can be accurately estimated, which further verifies the robustness and accuracy of our system.

\vspace{-10pt}

\vspace{-5pt}

\begin{figure}[!hbt]
	\centering
    \vspace{6pt}
    \setlength{\abovecaptionskip}{3pt}
	\captionsetup{belowskip=-10pt}
	\includegraphics[scale=0.36]{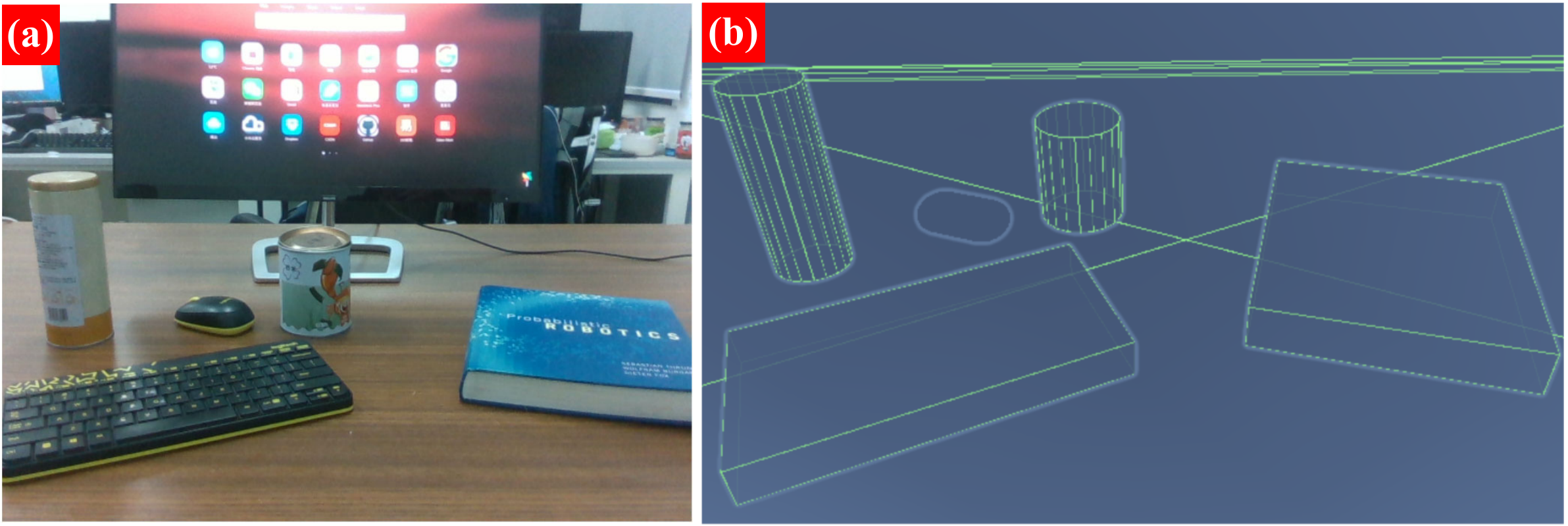}
	\caption{The raw image and the corresponding object map.}
	\label{AR_Map}
\end{figure}

\begin{figure}[!hbt]
	\centering
    \vspace{18pt}
    \setlength{\abovecaptionskip}{3pt}
	\captionsetup{belowskip=-10pt}
	\includegraphics[scale=0.26]{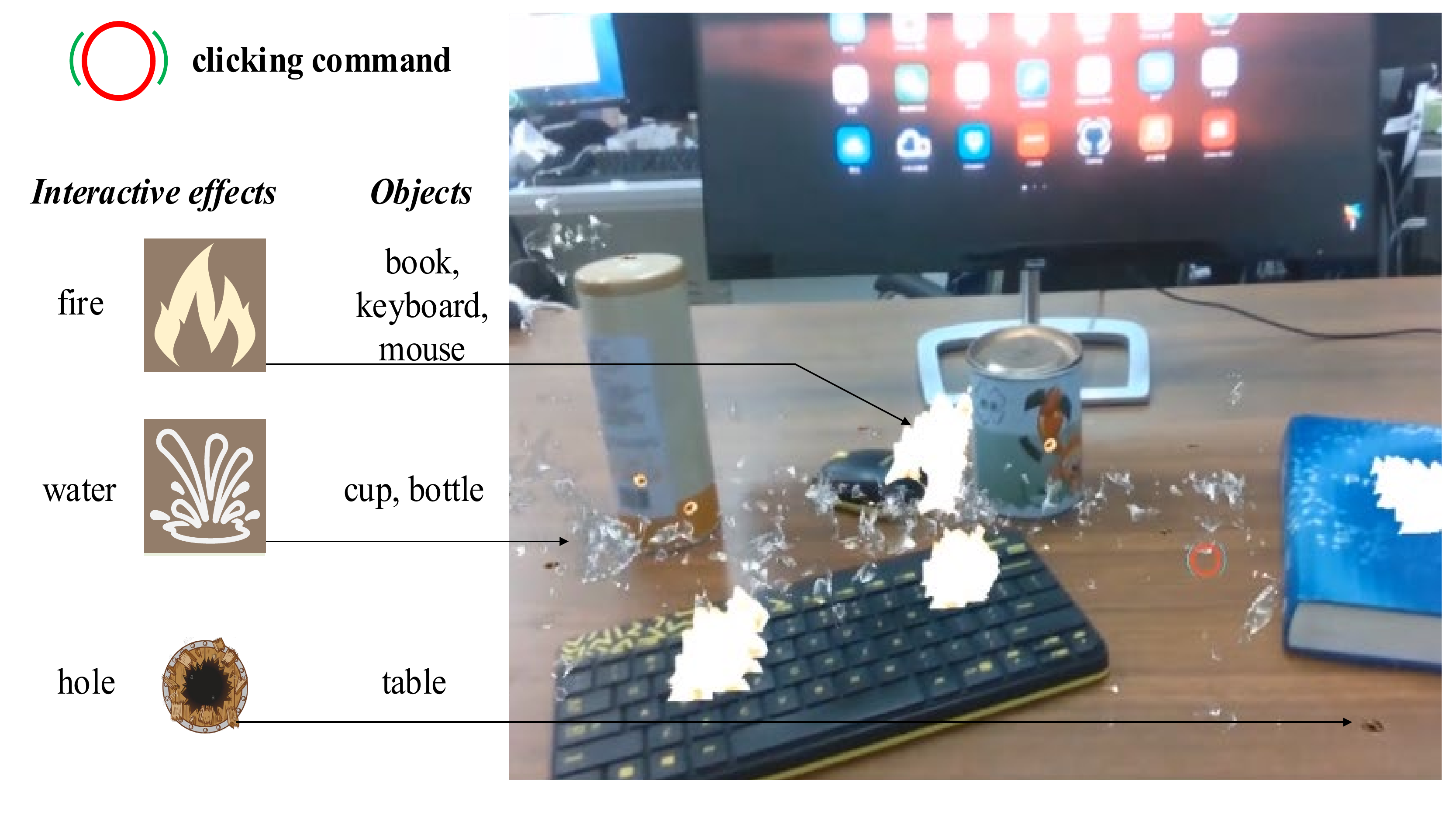}
	\caption{The demonstration of interaction. Reactions are visualized as a series of augmented reality events.}
	\label{AR_interaction}
\end{figure}

\subsection{Augmented Reality Experiment}

Early augmented reality used QR codes, 2D manual features, or image temples to register virtual 3D models, resulting in a restricted range of motion and poor tracking. The sparse point cloud map created by SLAM enables large-scale tracking and high-robust registration for AR. Geometric SLAM-based AR, however, is only concerned with accuracy and robustness, not authenticity. Conversely, our object SLAM-based AR provides complete environment information, thus a more realistic immersive experience can be achieved.

In the case of the desk scene in Fig. \ref{AR_Map}(a), we use the method described above to construct an object map, as shown in Fig. \ref{AR_Map}(b), in which we model objects such as the book, keyboard, and bottles.

\begin{figure*}[!t]
    \setlength{\abovecaptionskip}{0pt}
    \setlength{\belowcaptionskip}{-15pt}
	\centering
	\subfigure[Object-triggered \textbf{3D registration}. Top: raw images of the scene. Bottom: augmented reality scene with registered virtual models in place of the original objects.]{\includegraphics[scale=0.19]{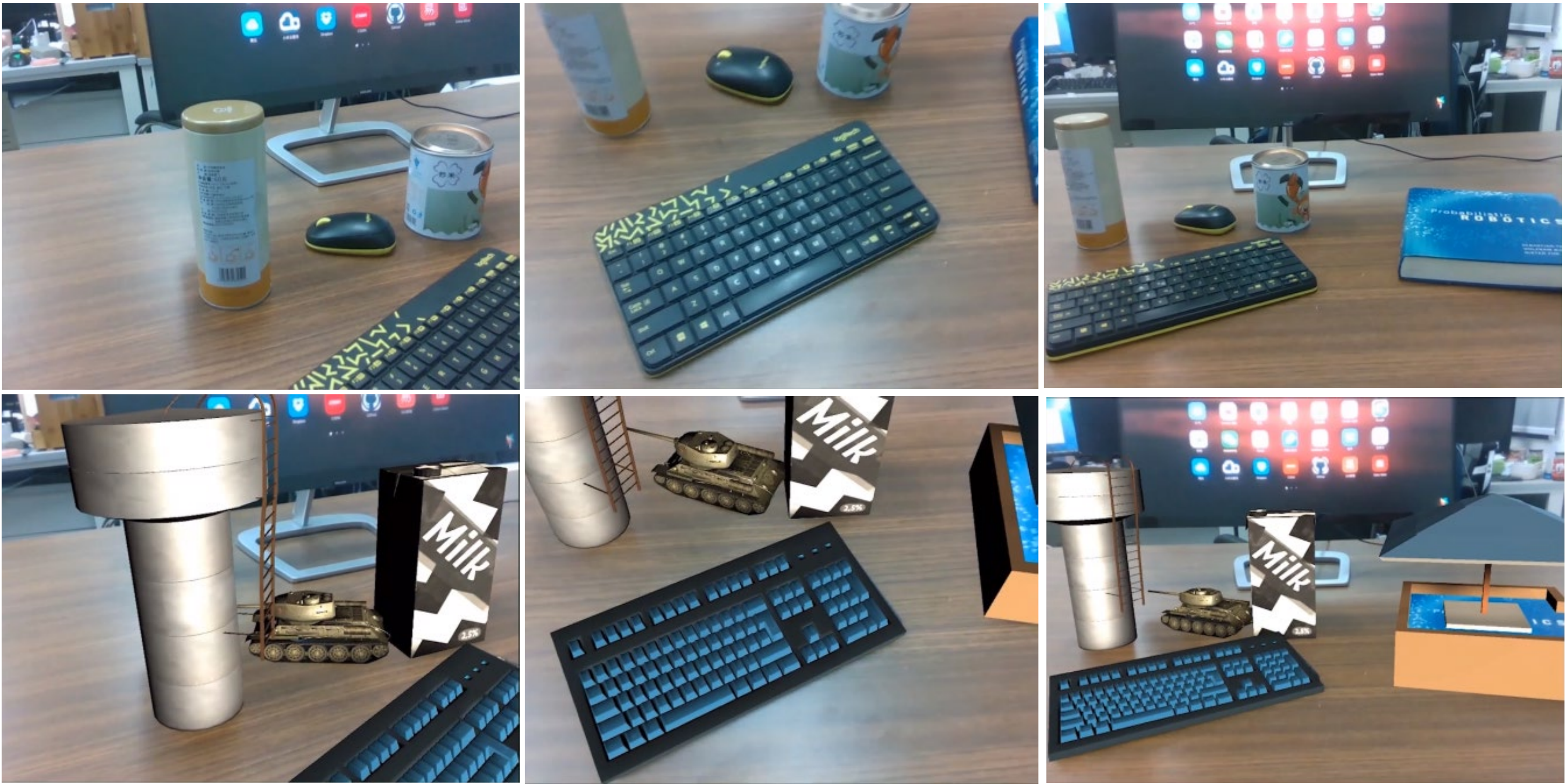}}
	\subfigure[The demonstration of \textbf{occlusion}. Top: the standard AR without occlusion. Bottom: our object SLAM-based AR.]{\includegraphics[scale=0.205]{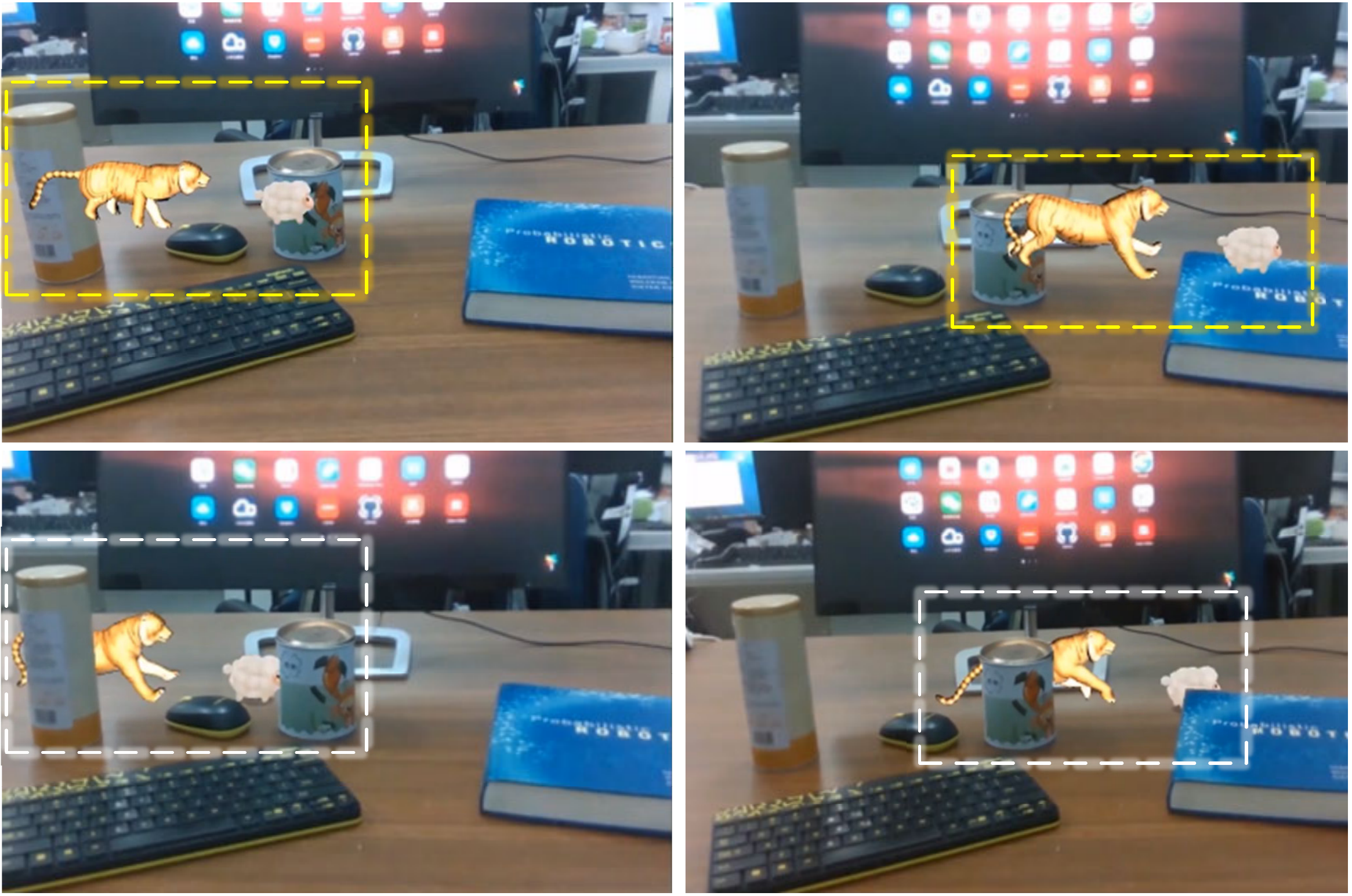}}
	\subfigure[The demonstration of \textbf{collision}. Top: the standard augmented reality. Bottom: our object-SLAM-based augmented reality with the awareness of collision.]{\includegraphics[scale=0.195]{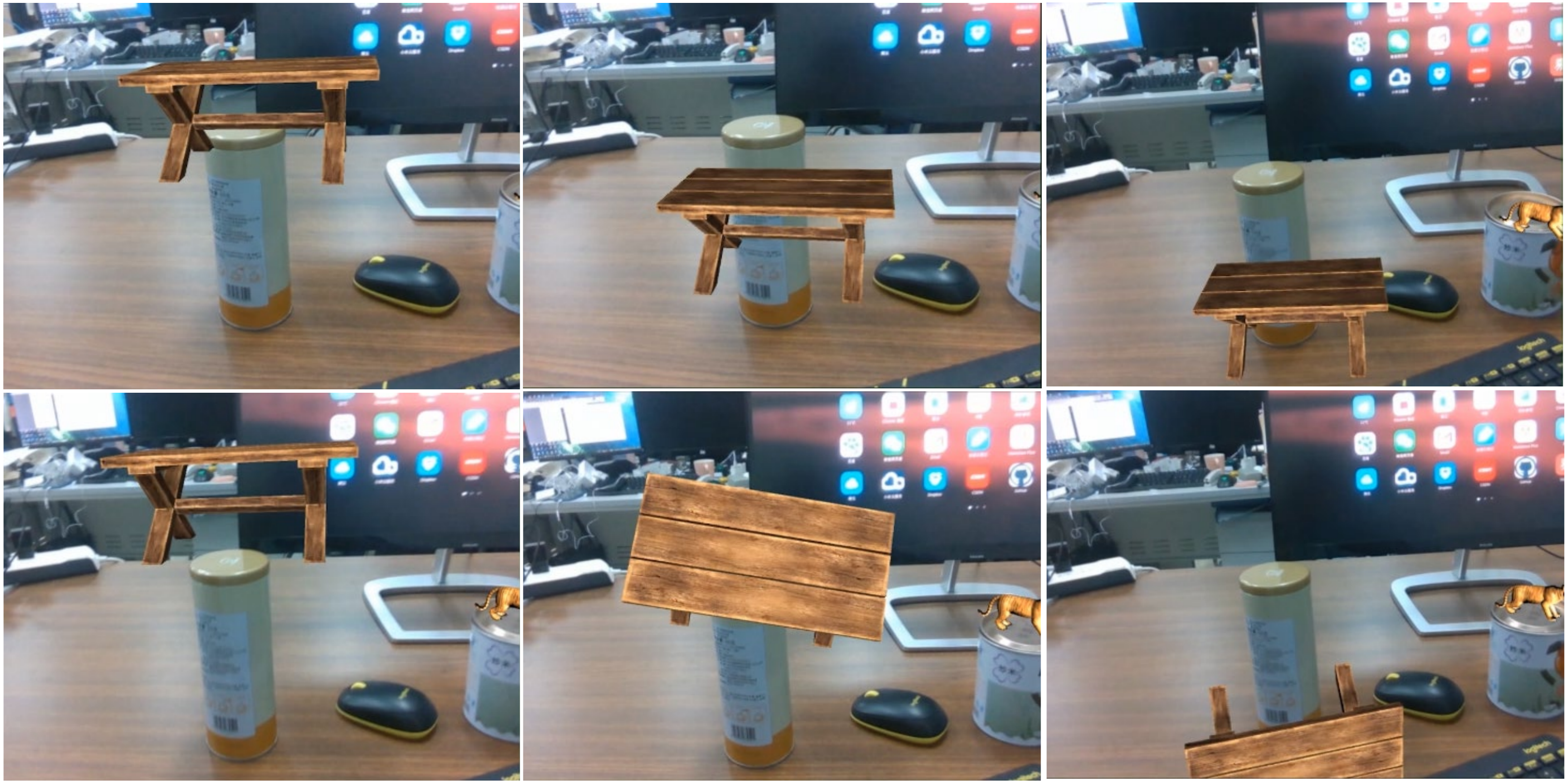}}
	\caption{3D registration, occlusion, and collision in object SLAM-based augmented reality.}
	\label{fig:AR}
\end{figure*}

\textbf{3D Registration}: We present an object-triggered virtual model registration method, instead of 3D registration triggered by a plane or position humanly specified. As shown in Fig. \ref{fig:AR}(a), the top row represents three raw frames from the video stream, while the bottom row represents the corresponding real-virtual integration scene. Virtual models can be seen registered on the desk to replace real objects based on the object semantics, pose, and size encoded in the object map.

\textbf{Occlusion and collision}: Physical occlusion and collision between the actual scene and virtual models is the crucial reflection of augmented reality. The top row, as seen in Fig. \ref{fig:AR}(b), is the result of common augmented reality, where virtual models are registered on the top layer of the image, resulting in an unrealistic separation of real and virtual scenes. The bottom row of Fig. \ref{fig:AR}(b) shows the outcome of our object SLAM-based augmented reality, in which the foreground and background are distinguished, and the real object obscures a portion of the virtual model, where the virtual and physical worlds are fused together.
Similarly, Fig. \ref{fig:AR}(c) depicts the collision effect. The virtual model in the top row falls on the desk without colliding with the bottle. Contrarily, the bottom row shows the outcome of our object SLAM-based augmented reality, in which the virtual model falls and collides with the real bottle, with the dropping propensity changing.

\textbf{Semantic interaction}: Interaction, cascading user command with the real scene and virtual models, plays a crucial role in augmented reality applications. As shown in Fig. \ref{AR_interaction}, clicking different real-world objects produces different virtual interactive effects.

The above functions, object-triggered 3D registration, occlusion, collision, and interaction, rely on accurate object perception of the proposed object SLAM framework. The experimental results demonstrate that object SLAM-based augmented reality has a fascinating benefit in areas such as gaming, military training, and virtual decorating.

\subsection{Object-based Scene Matching and Relocalization}
\label{sec:exp_scene_matching}

\begin{figure}[t]
	\centering
    \setlength{\abovecaptionskip}{0pt}
	\captionsetup{belowskip=-10pt}
	\subfigure[The matching result in TUM RGB-D fr2\_desk sequence.]{\includegraphics[scale=0.3]{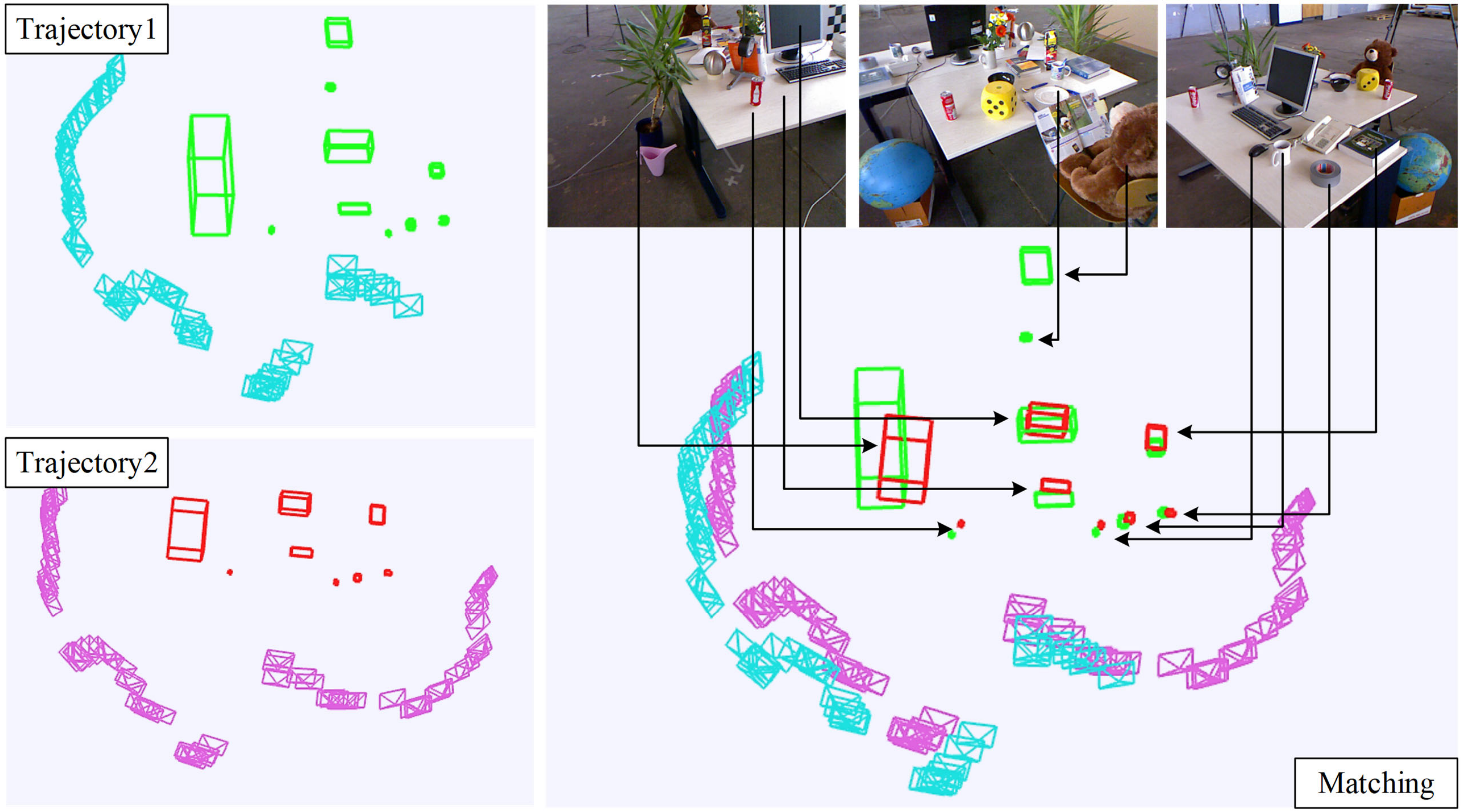}}
	\vspace{-2mm}
    \subfigure[The matching result in RGB-D Scenes v2 scene\_01 sequence.]{\includegraphics[scale=0.3]{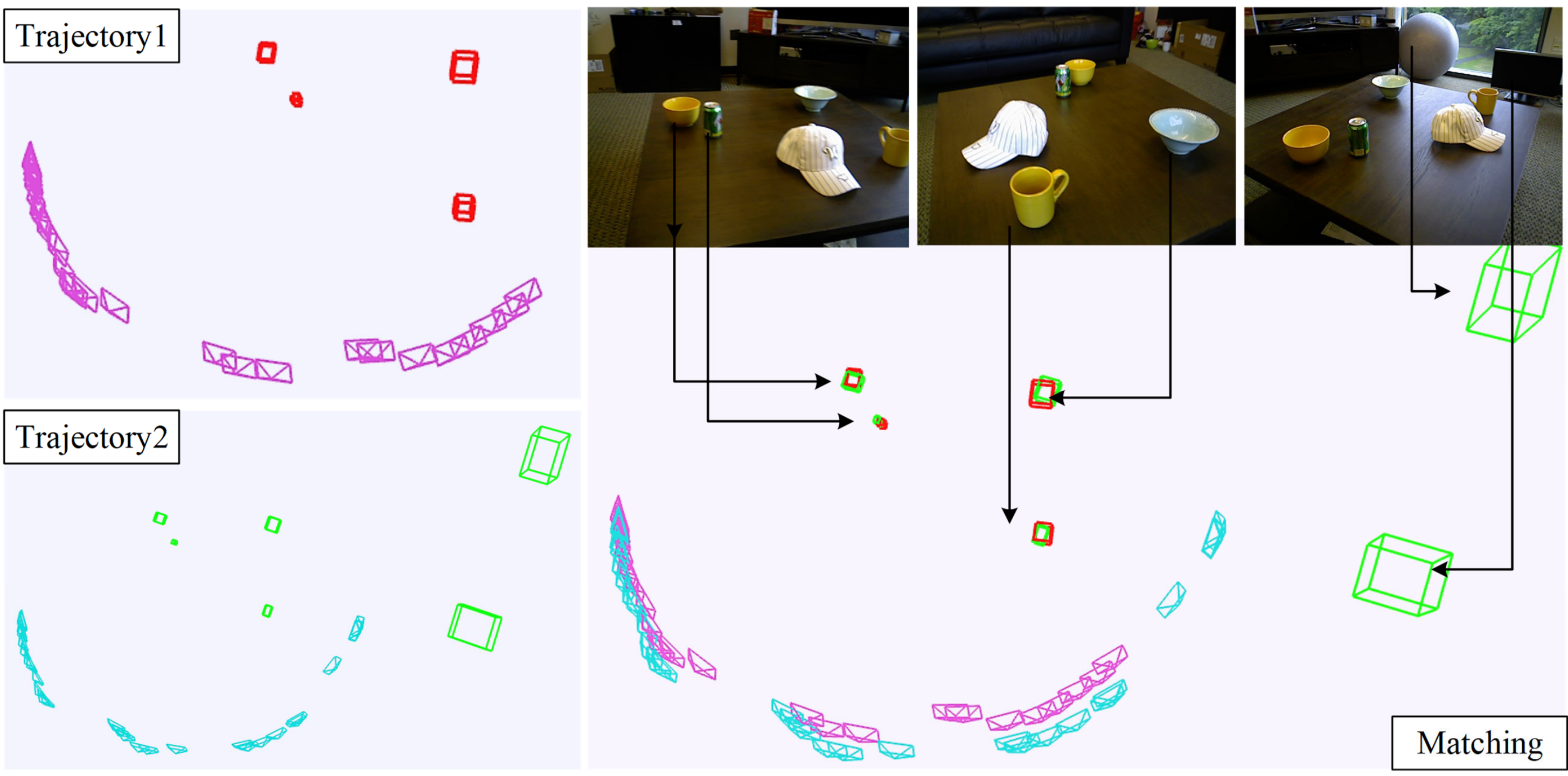}}
    \subfigure[The matching result under \textbf{different lighting conditions} in the real world.]{\includegraphics[scale=0.3]{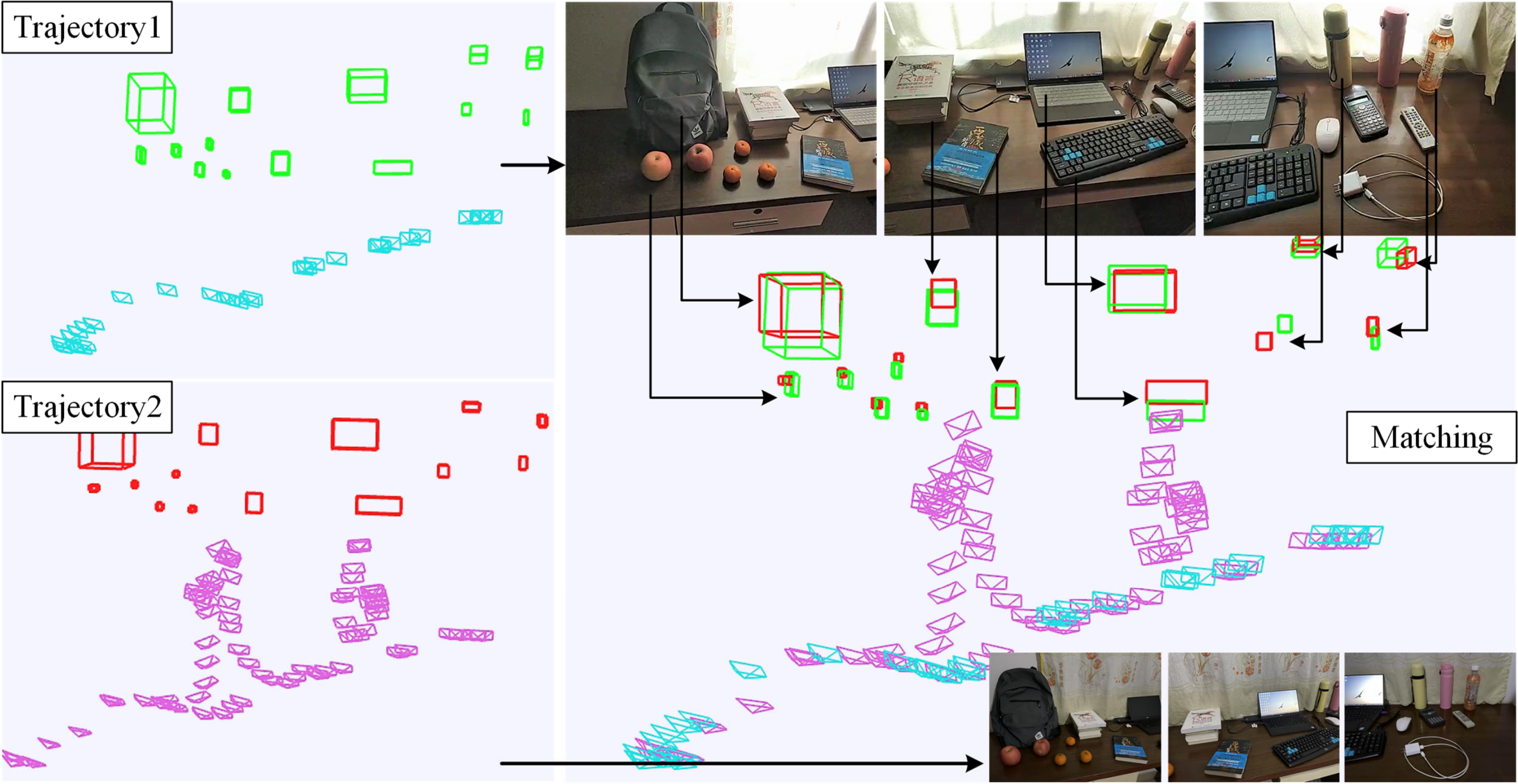}}
	\caption{The quantitative analysis of scene matching.}
	\label{scene_matching}
	\vspace{-4mm}
\end{figure}

\textbf{Scene Matching.} In this experiment, we evaluate the performance of the proposed object descriptor-based scene matching, which is crucial for multi-agent collaboration, scene reidentification, and multi-maps merging at different periods. We acquire two separate trajectories and their associated object maps in the same scene, then utilize the suggested method to figure out their relationship. Fig. \ref{scene_matching} illustrates the map-matching results in three settings.

The results of the TUM and Microsoft sequences are shown in Fig. \ref{scene_matching}(a) and Fig. \ref{scene_matching}(b). The two maps with different scales and numbers of objects match accurately, and the translation between them is also resolved. The match is not based on point clouds or BoW (Bag of Words) of keyframes, but the semantic object descriptor constructed by the object topological map. Additionally, the scale inconsistency of the two maps is also eliminated. Fig. \ref{scene_matching}(c) shows a real-world example of the matched result. Apart from the previous features, what is worth noting is that the two maps were \textbf{recorded under different illumination}. With this scenario, the traditional appearance-based method is trends to fail, demonstrating the robustness of the proposed object descriptor with the semantic level invariance property.

Table \ref{tab4} analyzes the performance time of the algorithm. The matching duration is found to be the primary cost, and the time is positively related to the number of objects. The average total duration is approximately 1.23ms, which is both practical and economical in various robot applications.

\vspace{-3pt}
\begin{table}[h]
\setlength{\abovecaptionskip}{-5pt}
\caption{TIME ANALYSIS OF SCENE MATCHING (ms)}
\begin{center}
\label{tab4}
\begin{tabular}{ccccccc}
\toprule
\multirow{2}{*}{Scene} & \multirow{2}{*}{\begin{tabular}[c]{@{}c@{}}Object \\ Num.\end{tabular}} & \multicolumn{2}{c}{Dscriptor Generate} & \multirow{2}{*}{Match} & \multirow{2}{*}{\begin{tabular}[c]{@{}c@{}}Pose \\ Resolve\end{tabular}} & \multirow{2}{*}{Total} \\ \cline{3-4}
                       &                                                                         & Map1               & Map2              &                        &                                                                          &                        \\ \hline
1                      & 6+4                                                                     & 0.203              & 0.167             & 0.458                  & 0.286                                                                    & 0.744                  \\
2                      & 10+8                                                                    & 0.213              & 0.184             & 0.673                  & 0.383                                                                    & 1.056                  \\
3                      & 14+14                                                                   & 0.483              & 0.437             & 1.001                  & 0.891                                                                    & 1.892                  \\
Ave                    & 10+8.7                                                                  & 0.300              & 0.263             & 0.711                  & 0.520                                                                     & 1.231                  \\ \bottomrule
\end{tabular}
\end{center}
\vspace*{-1.0\baselineskip}
\end{table}

\textbf{Relocalization.} We perform relocalization experiments with parallax to demonstrate the robustness of the proposed matching method to viewpoint changes. As illustrated in the figure in Tab. \ref{tab:reloc_rate_parallax}, we first construct a prior map with a set of prior keyframes and then utilize query keyframes for relocalization, which do not overlap the trajectories of prior keyframes and have parallax. We conduct several repeated experiments under different parallax conditions and compare the success rate of relocalization with ORB-SLAM3~\cite{campos2021orb}.
When the parallax is less than ${20}^{\circ}$, as shown in Tab. \ref{tab:reloc_rate_parallax}, ORB-SLAM3 achieves a relocalization success rate of 32.5\%; however, the rate drops sharply to 0 when the parallax is greater than ${20}^{\circ}$, which demonstrates that the appearance-based descriptor represented by ORB-SLAM3 is extremely sensitive to parallax. Conversely, our method is robust to parallax and achieves a success rate of over 12\% even under challenging large parallax.

\begin{table}[h]
\setlength{\abovecaptionskip}{0pt}
\renewcommand{\arraystretch}{1.0}
\caption{RELOCALIZATION SUCCESS RATE UNDER DIFFERENT PARALLAX}
\label{tab:reloc_rate_parallax}
\begin{tabular}{p{2.8cm}<{\centering} |p{0.5cm}<{\centering} p{0.5cm}<{\centering} ccc}
\toprule
\multirow{5}{*}{ 
\begin{minipage}[b]{0.32\columnwidth}
		\centering
		\raisebox{-.5\height}{\includegraphics[width=\linewidth]{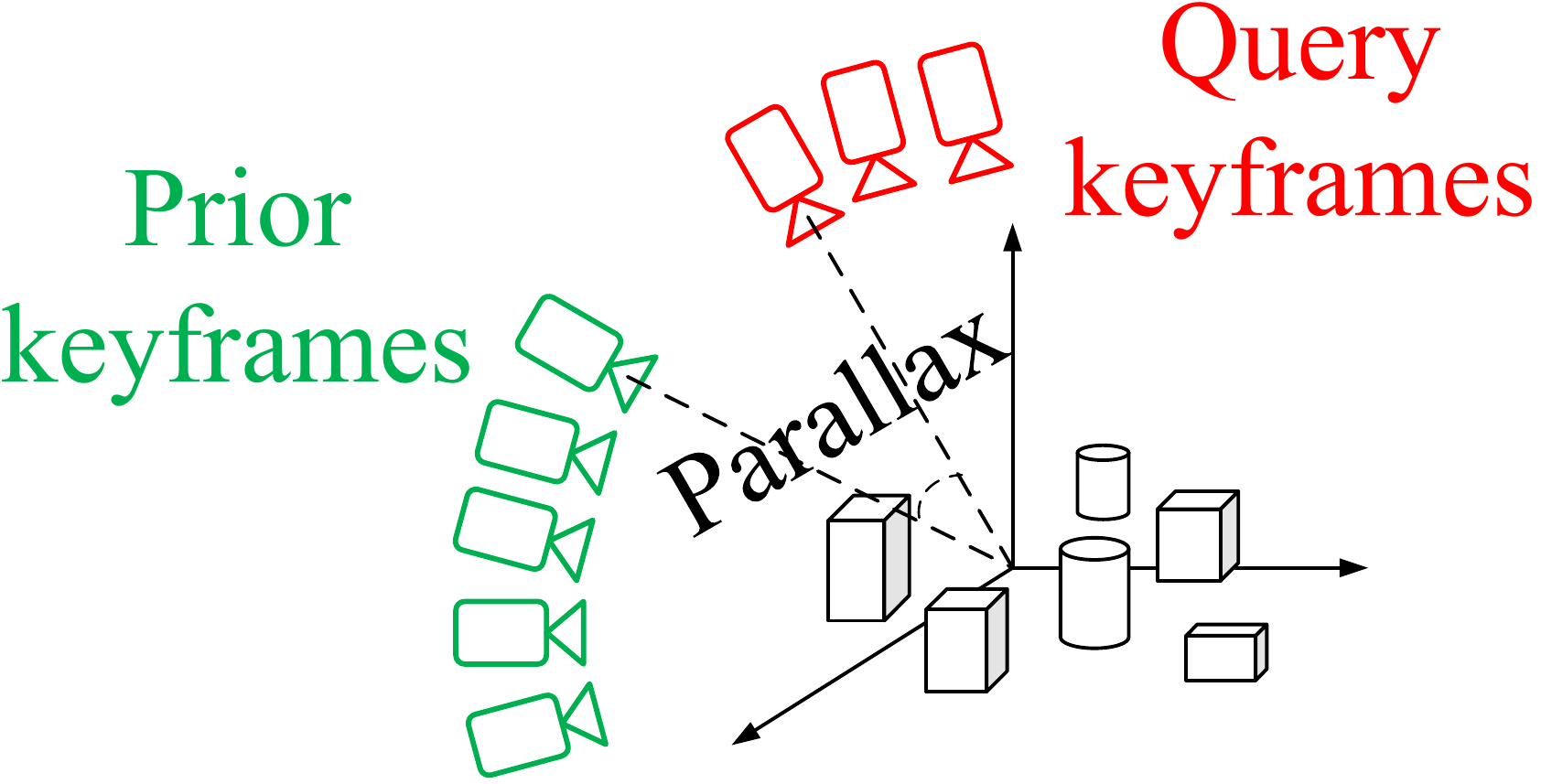}}
	\end{minipage}
} & \multirow{2}{*}{\makecell[c]{Total \\ times}} & \multirow{2}{*}{\makecell[c]{Succ. \\ times}} & \multirow{2}{*}{Parallax} & \multicolumn{2}{c}{Success rate(\%)} \\ \cline{5-6} 
                                                 &                              &                                   &                           & Ours              & \cite{campos2021orb}        \\ \cline{2-6} 
                                                 & 329                          & 49                                & \textless{}20             & 14.9              & \textbf{32.5}    \\
                                                 & 648                          & 96                                & 20-50                     & \textbf{14.8}     & 0                \\
                                                 & 900                          & 109                               & 50-100                    & \textbf{12.1}     & 0                \\ \bottomrule
\end{tabular}
\vspace{-10pt}
\end{table}

However, the accuracy of 14.9\% is still unsatisfactory. We found that the primary reason is that the observations of the two sets of keyframes are incomplete, thus resulting in inaccurate object modeling. To prove our hypothesis, we manually generate a scene with objects and divide it into a prior map and query map, assuming that prior and query keyframes generate them, respectively, and that the poses of the objects are obtained from the ground truth.
As depicted in the figure in Tab. \ref{tab:reloc_rate_gt}, we adjust the proportion of shared objects across the two maps and measure the success rate of relocalization. As demonstrated in Tab. \ref{tab:reloc_rate_gt}, we obtain a 100\% success rate with a public object ratio of over 50\% and retain over 80\% accuracy with a ratio of 33\%. The result demonstrates the effectiveness of our proposed object descriptor and matching algorithm. It also illustrates its sensitivity to object pose and suggests that more accurate object modeling methods can improve its performance.

\begin{table}[h]
\setlength{\abovecaptionskip}{0pt}
\caption{RELOCALIZATION SUCCESS RATE UNDER DIFFERENT PUBLIC OBJECT RATIO}
\label{tab:reloc_rate_gt}
\begin{tabular}{l|p{0.5cm}<{\centering} p{0.5cm}<{\centering} cc}
\toprule
\multirow{7}{*}{
\begin{minipage}[b]{0.37 \columnwidth}
		\centering
		\raisebox{-.5\height}{\includegraphics[width=\linewidth]{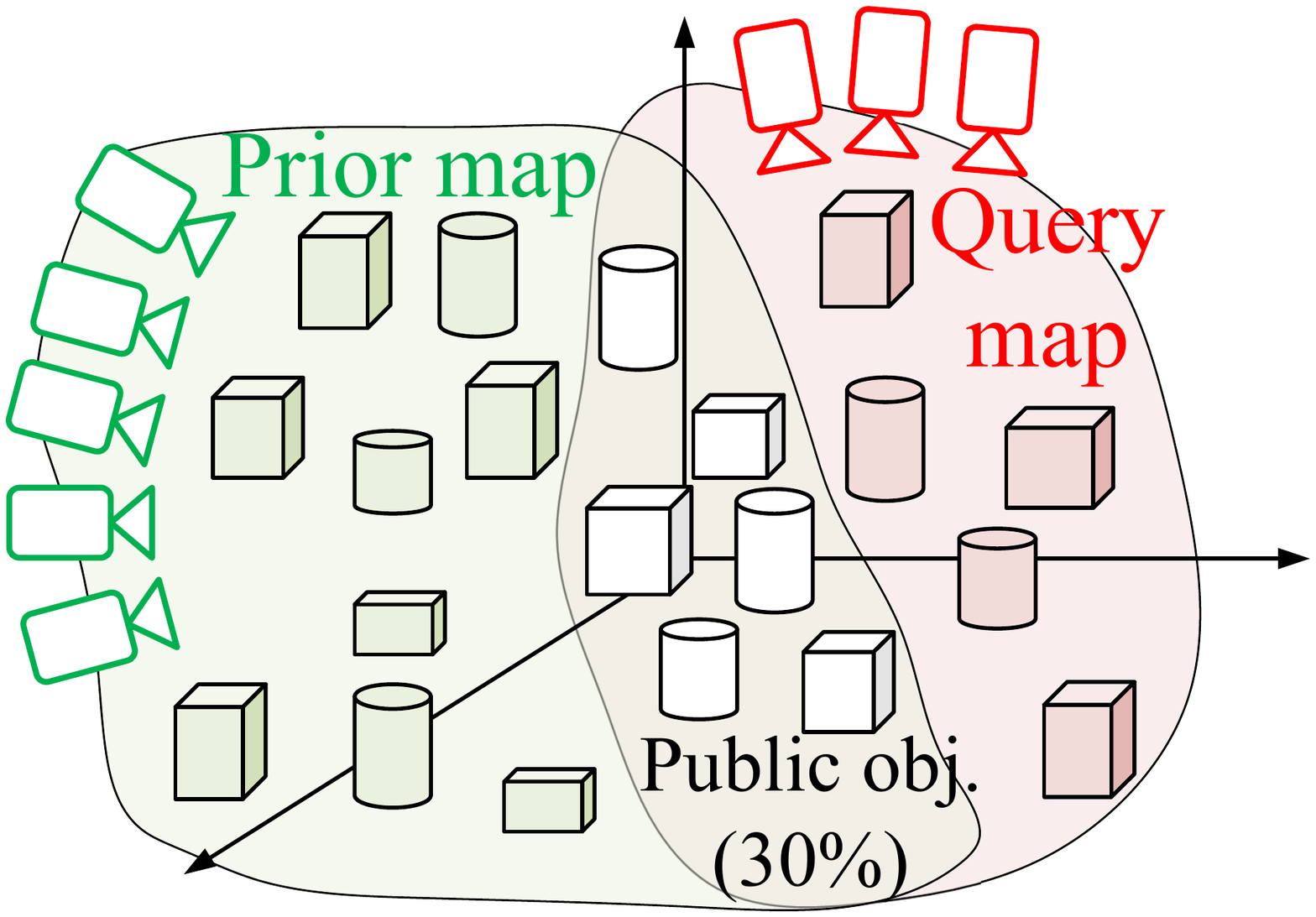}}
	\end{minipage}
} & \multicolumn{1}{l}{  {\makecell[c]{Total \\ times}}} & \multicolumn{1}{l}{  {\makecell[c]{Succ. \\ times}}} & \multicolumn{1}{l}{   {\makecell[c]{Public obj. \\ ratio (\%)}}} & \multicolumn{1}{l}{  {\makecell[c]{Success \\ rate (\%)}}} \\ \cline{2-5} 
                                      &   {500}                                      &   {500}                                           &   {60}                                    &   {100}                                            \\
                                      &   {500}                                      &   {500}                                           &   {55}                                    &   {100}                                            \\
                                      &   {500}                                      &   {500}                                           &   {50}                                    &   {100}                                            \\
                                      &   {500}                                      &   {499}                                           &   {44}                                    &   {99.8}                                           \\
                                      &   {500}                                      &   {467}                                           &   {38}                                    &   {93.4}                                           \\
                                      &   {500}                                      &   {406}                                           &   {33}                                    &   {81.2}                                           \\ \bottomrule
\end{tabular}
\vspace{-5pt}
\end{table}
\color{black}

\vspace{-5mm}
\subsection{Evaluation of Active Mapping}

\begin{table}[tb]
	\setlength{\abovecaptionskip}{-5pt}
	\centering
	\renewcommand{\arraystretch}{1.3}
 \setlength{\tabcolsep}{7pt}
	\caption{ACCURACY OF OBJECT POSE ESTIMATION}
	\begin{center}
		\label{tab3}
		\begin{tabular}{c c cccc}
			\toprule
			\multicolumn{1}{l}{Scene} & Metrics   & Ours          & Random.       & Cover.           & Init.        \\ \hline
			
			& 3D IoU       & \textbf{0.427}  & 0.3056       & 0.3329          & 0.3586       \\
			
			& 2D IoU  & \textbf{0.6225} & 0.4571       & 0.5221          & 0.5212       \\
			
			& CDE & \textbf{1.5272} & 2.2699       & 1.7876          & 2.2022       \\
			
			\multirow{-4}{*}{1}        & YAE    & 3.5             & 4.8          & 3.8             & \textbf{2.8} \\ \hline
			
			& 3D IoU       & \textbf{0.4307} & 0.3017       & 0.4224          & 0.3400       \\
			
			& 2D   IoU    & \textbf{0.8679} & 0.6422       & 0.7730          & 0.6480       \\
			
			& CDE & \textbf{1.4646} & 2.1096       & 1.5931          & 1.9822       \\
			
			\multirow{-4}{*}{2}        & YAE & 2.4             & \textbf{1.8} & 2.7             & 2.4          \\ \hline
			
			& 3D IoU       & \textbf{0.4132} & 0.3125       & 0.3685          & 0.2617       \\
			
			& 2D   IoU    & \textbf{0.6225} & 0.4915       & 0.5517          & 0.3909       \\
			
			& CDE & 1.5503          & 2.0672       & \textbf{1.4841} & 2.7489       \\
			
			\multirow{-4}{*}{3}        & YAE  & 3.9             & \textbf{3.7} & 3.8             & 4.9          \\ \hline
			
			& 3D IoU       & \textbf{0.4790} & 0.3824       & 0.3664          & 0.3007       \\
			
			& 2D IoU  & \textbf{0.6536} & 0.5886       & 0.4788          & 0.4869       \\
			
			& CDE & \textbf{1.3335} & 1.3514       & 1.7508          & 1.927        \\
			
			\multirow{-4}{*}{4}        & YAE       & 2.9             & 2.8          & \textbf{2.1}    & \textbf{2.1} \\ \hline
			
			& 3D IoU       & \textbf{0.5177} & 0.2696       & 0.2884          & 0.3720       \\
			
			& 2D IoU      & \textbf{0.6263} & 0.4297       & 0.4326          & 0.6142       \\
			
			& CDE & \textbf{1.3704} & 2.5077       & 2.1753          & 2.0084       \\
			
			\multirow{-4}{*}{5}        & YAE  & 3.9             & \textbf{2.1} & 3.9             & \textbf{2.1}          \\ \hline
			
			& 3D IoU       & \textbf{0.4411} & 0.3000       & 0.3597          & 0.2916       \\
			
			& 2D IoU    & \textbf{0.5437} & 0.4850       & 0.4783          & 0.5042       \\
			
			& CDE & \textbf{2.5998} & 3.4278       & 2.8411          & 3.4965       \\
			
			\multirow{-4}{*}{6}        & YAE    & \textbf{2.3}    & 2.7          & 3               & 2.7          \\ \hline
			
			& 3D IoU       & \textbf{0.4626} & 0.2118       & 0.4133          & 0.3153       \\
			
			& 2D IoU    & \textbf{0.6017} & 0.3839       & 0.5569          & 0.4541       \\
			
			& CDE & 1.49928         & 2.3822       & \textbf{1.4832} & 2.0467       \\
			
			\multirow{-4}{*}{7}        & YAE    & \textbf{2.1}    & 4.5          & 2.5             & 2.5          \\ \hline
			
			& 3D IoU       & \textbf{0.453}  & 0.2977       & 0.3645          & 0.3200       \\
			
			& 2D IoU      & \textbf{0.6483} & 0.4969       & 0.5419          & 0.5171       \\ 
			
			& CDE & \textbf{1.6207} & 2.3022       & 1.8736          & 2.3446       \\
			
			\multirow{-4}{*}{Mean}     & YAE   & 3               & 3.2          & 3.1             & \textbf{2.8} \\ \bottomrule
		\end{tabular}
	\end{center}
	\vspace{-8mm}
\end{table}

\begin{figure*}[t]
	\centering
	\captionsetup{belowskip=-10pt}
	\includegraphics[scale=0.31]{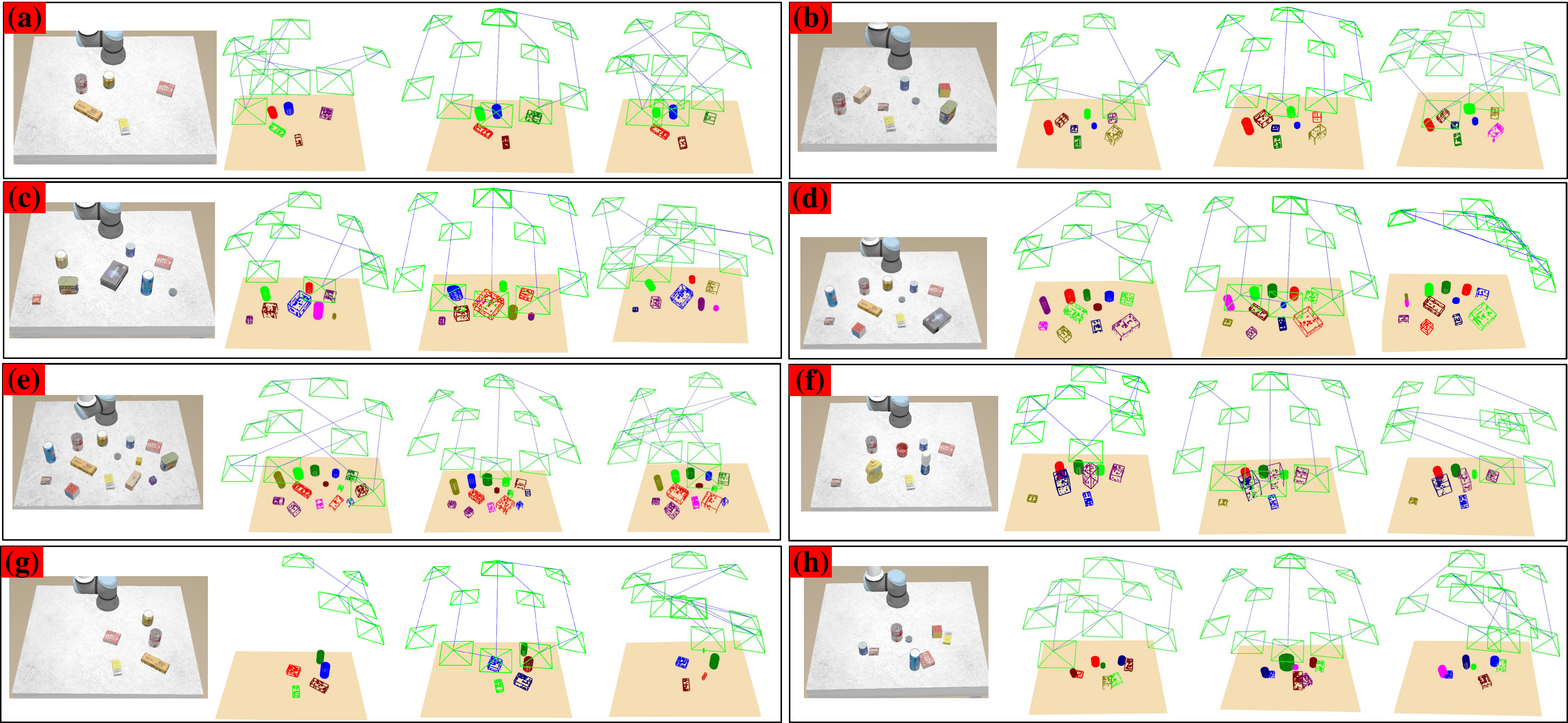} 
	\caption{Comparison of mapping results. The first column in the sub-picture: the scene image; the second column: the result of our object-driven exploration; the third column: the result of the coverage exploration; the fourth column: the result of the randomized exploration.}
	\label{Mapping_Result}
    \vspace{-2pt}
\end{figure*}

To validate the effectiveness of the active map building and the viability of robot manipulation led by the map, we conduct extensive evaluations in both simulation and real-world environments. The simulated robotic manipulation scene is set in Sapien \cite{xiang2020sapien}, shown in Fig. \ref{Mapping_Result}, where the number of objects and the scene complexities vary in different scenes.

The accurate position estimate is critical for successful robotic manipulation operations such as grasping, placing, arranging, and planning. However, precision is difficult to ensure when the robot estimates autonomously. To quantify the effect of active exploration on object pose estimation, like previous studies \cite{arruda2016active,morrison2019multi}, we compare our object-driven method with two typically used baseline strategies, \textit{i.e.}, randomized exploration (Random.) and coverage exploration (Cover.). As indicated in Fig. \ref{Mapping_Result}, for randomized exploration, the camera pose is randomly sampled from the reachable set relative to the manipulator, while for coverage exploration, a coverage trajectory based on Boustrophedon decomposition \cite{kaljaca2020coverage} is leveraged to scan the scene. At the beginning of all the explorations, an initialization step (Init.), in which the camera is sequentially placed over the four desk corners from a top view, is applied to start the object mapping process. The simulator provides the ground truth of object position, orientation, and size. Correspondingly, the accuracy of pose estimation is evaluated by the Center Distance Error (CDE, c$\rm m$), the Yaw Angle Error (YAE, $\rm degree$), and the IoU (including 2D IoU from the top view and 3D IoU) between the ground truth and our estimated results. 


Table \ref{tab3} shows the evaluation results in seven scenes (Fig. \ref{Mapping_Result}). We can see our proposed object-driven exploration strategy achieves a 3D IoU of 45.3\%, which is 15.53\%, 8.85\%, and 13.3\% higher than that of the randomized exploration, the coverage exploration, and the initialization, respectively. For 2D IoU, our method achieves an accuracy of 64.83\%, which is 15.14\%, 10.64\%, and 13.12\% higher than baseline methods. In terms of CDE, our method reaches 1.62cm, significantly less than other methods. For YAE, all exploration strategies achieve an error of approximately ${3}^{\circ}$, which verifies the robustness of our line-alignment-based yaw angle optimization method. The level of above precision attained is sufficient for robotic manipulation\cite{wang2019densefusion}. Moreover, we also find that randomized exploration sometimes performs worse than the initialization result (rows 2, 5, and 7), which indicates that increasing observations do not necessarily result in more accurate pose estimation, and purposeful exploration is necessary.

The mapping results are shown in Fig. \ref{Mapping_Result}. The cubes and cylinders are used to model the objects, including poses and scales (analyzed above), based on their semantic categories. The following characteristics are present: \textbf{1)} The system can accurately model various objects as the number of objects increases, as shown in Fig. \ref{Mapping_Result}(a)-(e), demonstrating its robustness.
\textbf{2)} Among objects of various sizes, our method focuses more on large objects with lower observation completeness (see Fig. \ref{Mapping_Result}(f)). 
\textbf{3)} When objects are distributed unevenly, our proposed strategy can swiftly concentrate the camera on object regions, thus avoiding unnecessary and time-consuming exploration (see Fig. \ref{Mapping_Result}(g).
\textbf{4)} For scenes with objects close to each other, our method can focus more on regions with fewer occlusions (see Fig. \ref{Mapping_Result}(h)). These behaviors verify the effectiveness of our exploration strategy. Additionally, our method has a shorter exploration path yet produces a more precise object posture.

\begin{figure*}[!htbp]
	\centering
    \setlength{\abovecaptionskip}{-3pt}
    \subfigcapskip=-5pt
	\captionsetup{belowskip=-10pt}
	\subfigure[Grasping process in the simulated environment.]
    {\includegraphics[scale=0.22]{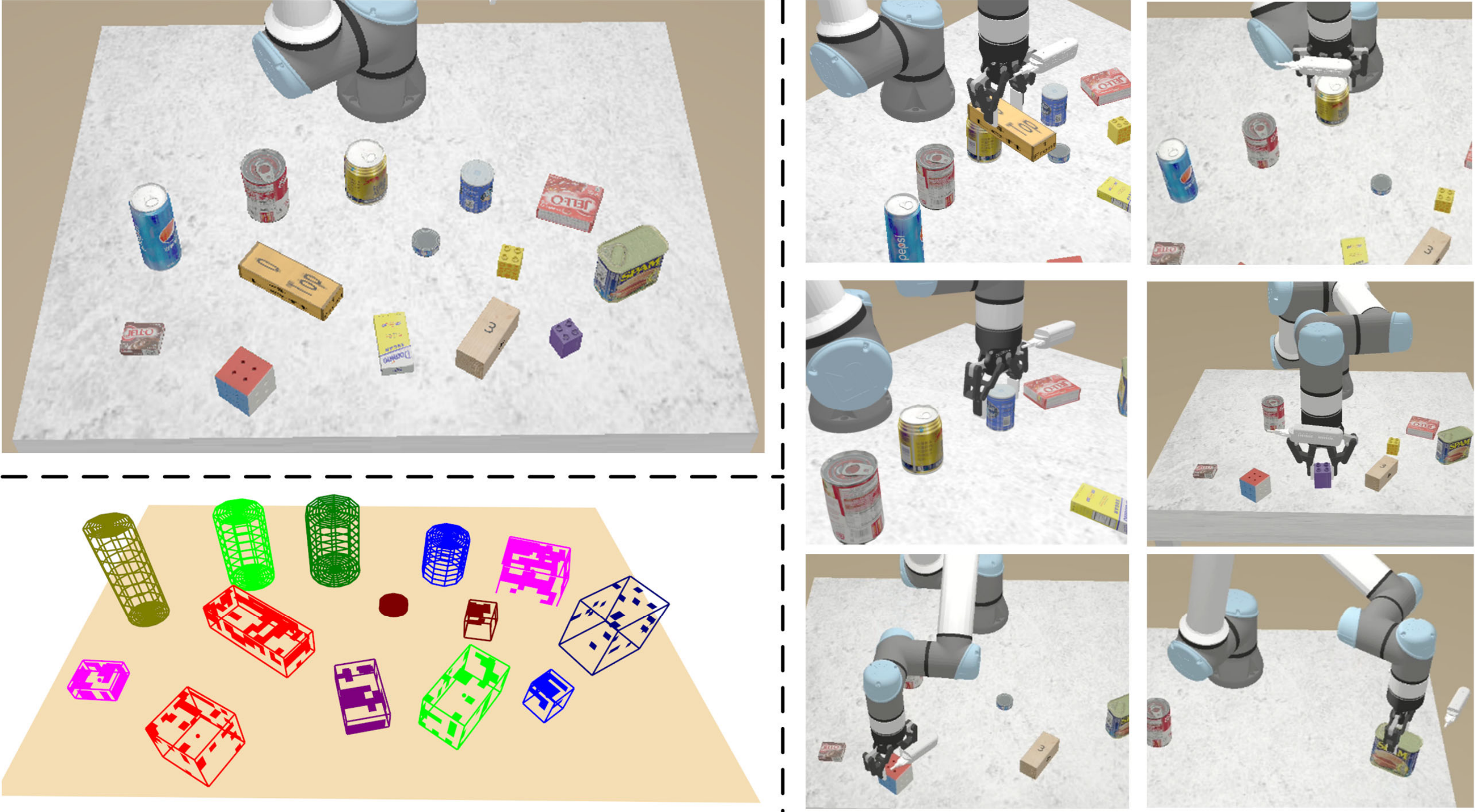}}
	\subfigure[Grasping process in the real world.]{\includegraphics[scale=0.38]{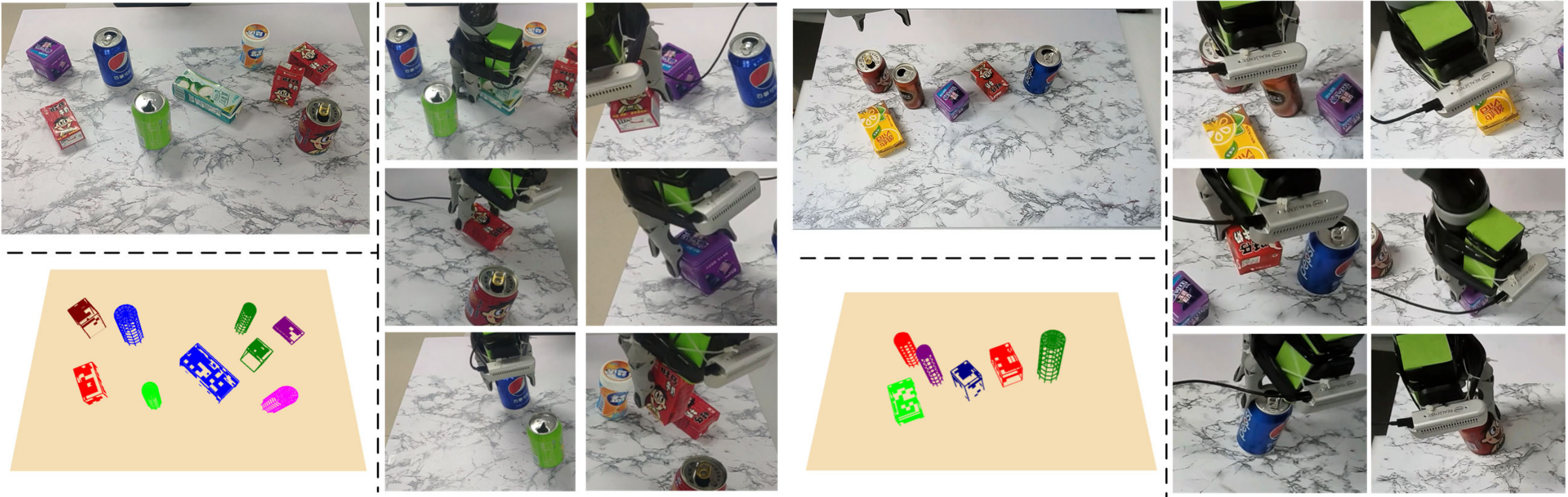}}
	\caption{The demonstration of grasping process.}
	\label{Grasp_Experiments}
	\vspace{-5pt}
\end{figure*}

\begin{figure}[!htbp]
	\centering
    \setlength{\abovecaptionskip}{3pt}
	\captionsetup{belowskip=-5pt}
	\includegraphics[scale=0.33]{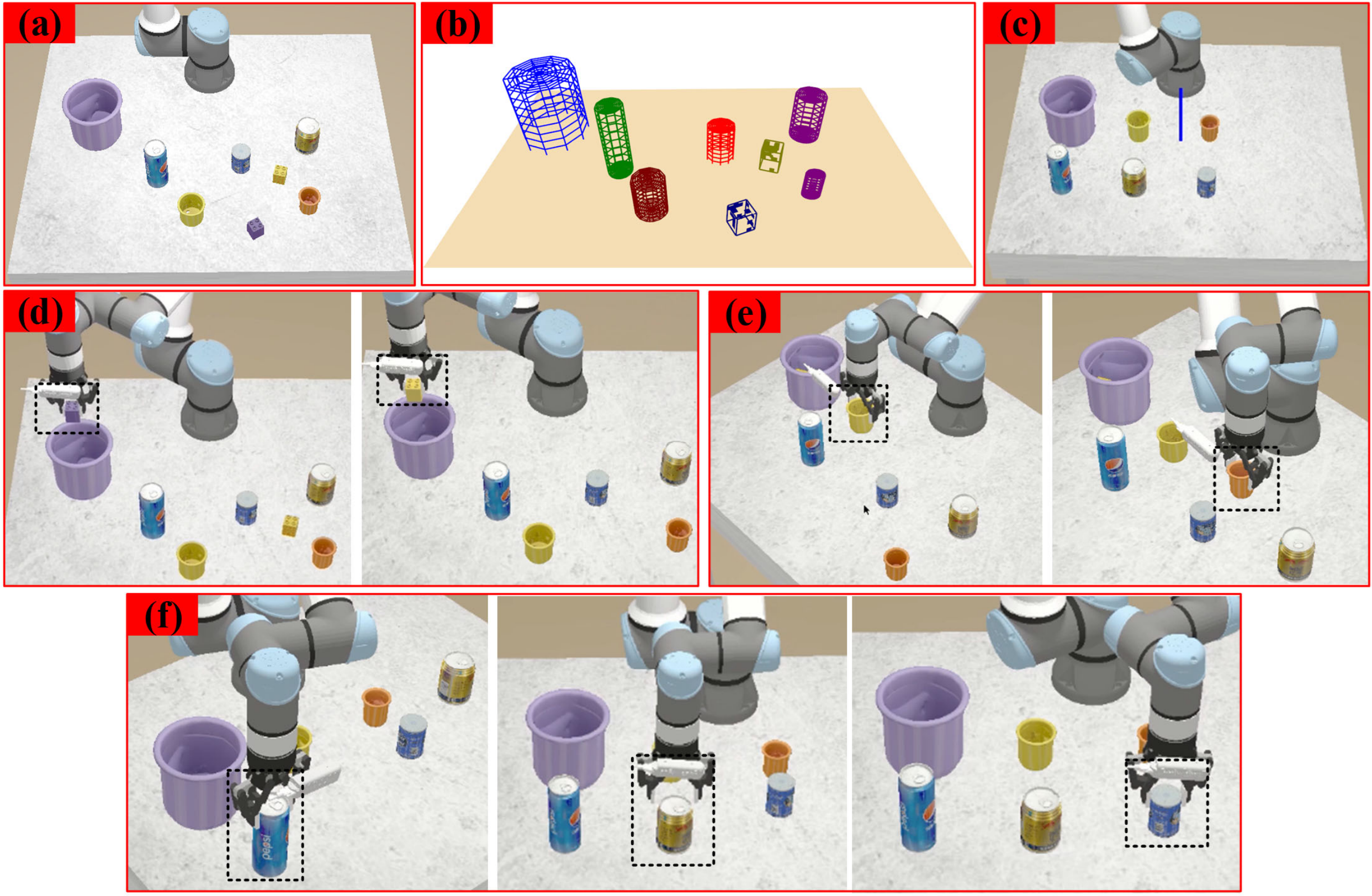}
	\caption{Object placement according to the global object map.}
	\label{Placement}
	\vspace{-8pt}
\end{figure}

\vspace{-10pt}
\subsection{Object Grasping and Placement}

This experiment uses the incrementally generated object map to perform object grasping. Fig. \ref{Grasp_Experiments}(a) and Fig. \ref{Grasp_Experiments}(b) illustrate the grasping process in simulated and real-world environments, with the object map included. After extensive testing, we obtained a grasping success rate of approximately 86\% in the simulator and 81\% in the real world, which may be affected by environmental or manipulator noises. It is found that the center and direction of the objects have a significant influence on grasping performance. The proposed method performs well regarding these two metrics, thus ensuring high-quality grasping. 
Overall, our object SLAM-based pose estimation results can satisfy the requirements of grasping.

We argued that the proposed object map level perception outperforms object pose-only perception and provides information for more intelligent robotics decision-making tasks in addition to grasping. Such include avoiding collisions with other objects, updating the map after grasping, object arrangement and placement based on object properties, and object delivery requested by the user. We design the object placement experiments to verify the global perception capabilities introduced by object mapping.
As shown in Fig. \ref{Placement}, the robot is required to manipulate the original scene (see Fig. \ref{Placement}(a)) to the target scene (see Fig. \ref{Placement}(c)) according to object sizes and classes encoded in the object map.

The global object map is shown in Fig. \ref{Placement}(b), which contains the semantic labels, size, and pose of the objects. The two little blocks are picked up and placed in the large cup (Fig. \ref{Placement}(d)), while the cups are ordered by volume (Fig. \ref{Placement}(e)) and the bottles by height (Fig. \ref{Placement}(f)). This task is challenging for the conventional grasping approach since lacking global perception such as object's height on the map, its surroundings, and could interact with which objects.

\vspace{-5pt}
\section{Discussion and Analyze}
\label{S_DIS}
In this section, we analyze the limitations and implementation details of our method and provide potential solutions for the object SLAM community.

\textbf{1) Data association.}
Experiments revealed that the two primary reasons for the failure of data association could be summed up as follows: \textbf{\romannumeral1)} Long-tailed distribution. In some cases, object centroids are located in the tail of the distribution, which violates the Gaussian distribution assumption and causes the association to fail. Although our multiple association strategy alleviates this to some extent. \textbf{\romannumeral2)} Detected semantic label mistakes. Even if the IoU-based or distribution-based method determines an association between two objects, the association will fail if the labels are inconsistent. Error in label recognition is one of the most common issues with detectors. More generalized, accurate detectors or fine-turning on specific datasets are potential alternatives.

The running time of data association is shown in the first two columns of Tab. \ref{tab:time_data_para}. Distribution-based methods include non-parametric and t-tests, while IoU-based methods include motion IoU and project IoU, where projecting 3D points to 2D images takes most of the time. Note that the total duration of data association is less than the sum in the table, approximately 8 ms per frame, because sometimes some strategies can be skipped. 
We perform data association on every frame, which would be more time-efficient if simply performed on keyframes, as CubeSLAM does.

\begin{table}[h]
\setlength{\abovecaptionskip}{0pt}
\setlength{\belowcaptionskip}{-2pt}
\caption{TIME ANALYSIS OF DATA ASSOCIATION AND OBJECT PARAMETERIZATION (ms/frame)}
\begin{center}
\begin{tabular}{cc cc}
\toprule
\multicolumn{2}{c}{  Data association} & \multicolumn{2}{c}{  Object Parameterization} \\ 
  Distribution-based     &   IoU-based    &   i-Forest          &   Line alignment          \\ \midrule
  4.92                   &   7.83         &   3.86              &   2.71                    \\ \bottomrule
\end{tabular}
\end{center}
\label{tab:time_data_para}
\vspace{-3mm}
\end{table}

\textbf{2) Object Parameterization.}
Two factors typically cause failure situations of object pose estimation: \textbf{\romannumeral1)} Object surfaces lack texture, or objects are only partially seen due to occlusion or camera viewpoint. In this case, few object point clouds are collected, significantly reducing the pose estimation performance. \textbf{\romannumeral2)} Too many outliers lead the object to be estimated too large, or the modeled object is extremely small since the i-Forest algorithm falls into a local optimum.
In the alternatives, the 3D detector~\cite{liu2021group} based on the complete point cloud may not be optimal due to the incremental characteristic of SLAM; image-based 6-DOF pose estimation~\cite{wang2019normalized,labbe2020cosypose} is limited by the scale of the training data, resulting in poor generalization~\cite{ming2021object}. Conversely, incremental detection/segmentation~\cite{wu2021scenegraphfusion} and joint point cloud-image multimodal RGB-D 3D object detection~\cite{wang2022multimodal,yang2022boosting} are potentially feasible.

The runtime of object parameterization is shown in the last two columns in Tab. \ref{tab:time_data_para}, which takes around 6.5 ms per frame on average. The full SLAM system (for camera tracking and semantic mapping) runs at about 10 fps.

\textbf{3) Augmented Reality.}
Augmented reality performance depends on object modeling, camera localization accuracy, and the rendering effect. Here we provide detailed engineering implementations for SLAM developers to migrate their algorithms to augmented reality applications.
The AR system comprises three modules: \textbf{\romannumeral1)} The Localization and Semantic Mapping (LSM) modules provide camera position, point cloud, and object parameters. Sections \ref{S_DA} and \ref{S_OP} introduce the techniques. \textbf{\romannumeral2)} ROS \cite{quigley2009ros} data transfer module: send images captured by the camera to the LSM module and then publish the estimated camera pose and map elements. \textbf{\romannumeral3)} Virtual-real rendering module: Use the Unity3D engine to subscribe to topics published by ROS, construct a virtual 3D scene, and render it to a 2D image plane.
Tab.~\ref{tab:AR_time} details the duration of each module, which is executed in parallel.

\begin{table}[h]
\setlength{\abovecaptionskip}{-5pt}
\setlength{\belowcaptionskip}{-3pt}
\caption{TIME ANALYSIS OF AUGMENTED REALITY}
\begin{center}
\begin{tabular}{c c}
\toprule
  {Module}                            &   {Time (ms/frame)} \\ \midrule
  {Localization and semantic mapping} &   {81.84}           \\
  {ROS data transfer}                 &   {10.96}           \\
  {Virtural-real rendering}           &   {25.00}              \\ \bottomrule
\end{tabular}
\end{center}
\label{tab:AR_time}
\vspace{-5mm}
\end{table}

\textbf{4) Scene Matching.} The principal causes for the failure of scene matching and relocalization are: \textbf{\romannumeral1)} There are few common objects between the two maps, resulting in a significant difference in the descriptors of the same object in the two maps, leading to matching fails. \textbf{\romannumeral2)} The parallax of the trajectories of the two maps is excessively large, and the observation is insufficient, which affects the accuracy of object modeling and the construction of descriptors.
Regarding the first issue, other non-object-level landmarks, such as planes and structural components, can be considered for descriptor construction. For the second challenge, more accurate object modeling techniques can improve the performance of matching and relocalization, as demonstrated by our experiments.

\textbf{5) Object Grasping.}
There are two limitations to the object grasping task: \textbf{\romannumeral1)} Textured objects and tabletops are required for point-based SLAM tracking to succeed. \textbf{\romannumeral2)} Objects are all regular cube and cylinder shapes in our experiments. Complex irregular objects may necessitate more detailed shape reconstruction and grasp point detection. Nonetheless, we demonstrate the potential of object SLAM for grasping tasks without object priors. Model-free and unseen object grasping will be the future trend.
In terms of running time, the speed is even faster than 10fps because, in this setting, the data association is more straightforward, and more time is spent on the active mapping analysis process.

\color{black}
\section{Conclusion}
\label{S_CON}

We presented an object mapping framework that aims to create an object-oriented map using general models that parameterize the object's position, orientation, and size. 
First, we investigated related fundamental techniques for object mapping, including multi-view data association and object pose estimation. We then center on the object map and validate its potential in multiple high-level tasks such as augmented reality, scene matching, and object grasping. Finally, we analyzed the limitations and failure instances of our method and gave possible alternatives to inspire the development of related fields.
The following points will be given significant consideration in future work: 1) Dynamic objects data association, tracking, and trajectory prediction; 2) Irregular and unseen object modeling and tightly coupled optimization with SLAM; 3) Object-level relocalization and loop closure; 4) Omnidirectional perception with multi-sensor and multiple semantic networks to realize spatial AI.


%



\ifCLASSOPTIONcaptionsoff
  \newpage
\fi



%
\bibliographystyle{IEEEtran}
\bibliography{Object_Mapping}\


%

\vspace{-20mm}
\begin{IEEEbiography}[{\includegraphics[scale=0.15]{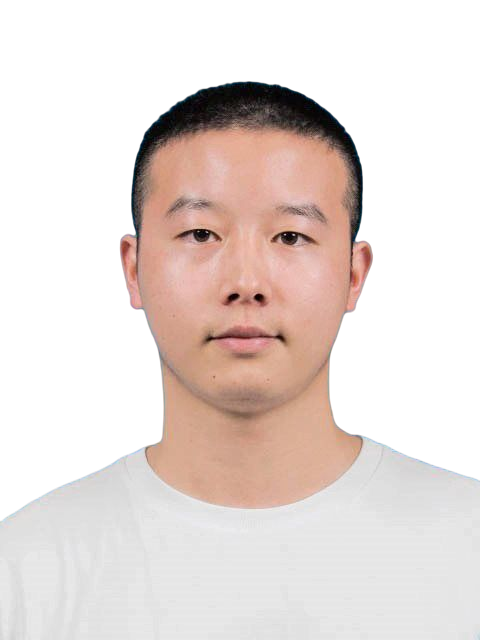}}]{Yanmin Wu}
received the B.S. degree in electronic information engineering from Shenyang Normal University, Shenyang, Liaoning, China, in 2018, and the M.S. degree in robot science and engineering from Northeastern University, Shenyang, Liaoning, China, in 2021.

He is currently pursuing a Ph.D. degree in computer applied technology at the School of Electronic and Computer Engineering, Peking University Shenzhen Graduate School,
Shenzhen, China. His research interests include visual SLAM, 3D reconstruction, and scene understanding.
\end{IEEEbiography}
\vspace{-20mm}

\begin{IEEEbiography}[{\includegraphics[scale=0.23]{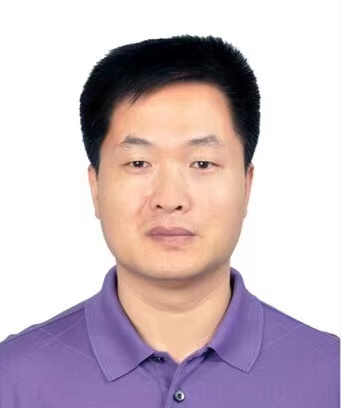}}]{Yunzhou Zhang}
received the B.S. and M.S. degrees in mechanical and electronic engineering from the National University of Defense Technology, Changsha, China, in 1997 and 2000, respectively, and the Ph.D. degree in pattern recognition and intelligent system from Northeastern University, Shenyang, China, in 2009.

He is currently a Professor with the Faculty of Robot Science and Engineering, Northeastern University. He leads the Cloud Robotics and Visual Perception Research Group. His research has been supported by funding from various sources, such as the National Natural Science Foundation of China, the Ministry of Education of China, and some famous high-tech companies. He has published many journal articles and conference papers on intelligent robots, computer vision, and wireless sensor networks. His research interests include intelligent robots, computer vision, and sensor networks.
\end{IEEEbiography}
\vspace{-14mm}

\begin{IEEEbiography}[{\includegraphics[scale=0.34]{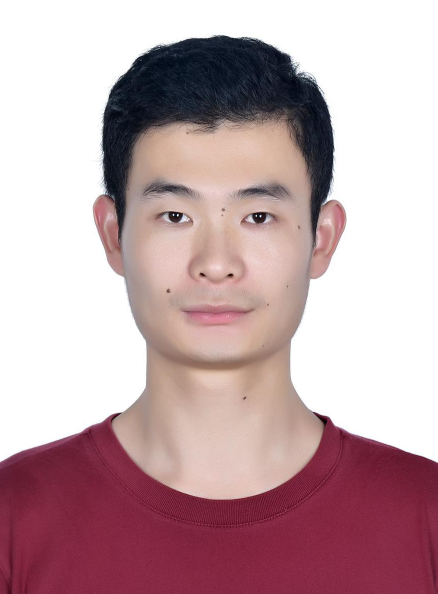}}]{Delong Zhu}
received the B.S. degree in computer science and technology from Northeastern University, Shenyang, Liaoning, China, in 2015, and the Ph.D. degree in electronic engineering from the Department of Electronic Engineering, The Chinese University of Hong Kong, Hong Kong, in 2020.

He spent nine months at the Robotics Institute, Carnegie Mellon University, Pittsburgh, PA, USA, as a Visiting Scholar. His research interests include motion planning in dynamic environments and deep reinforcement learning.
\end{IEEEbiography}

\begin{IEEEbiography}[{\includegraphics[scale=0.15]{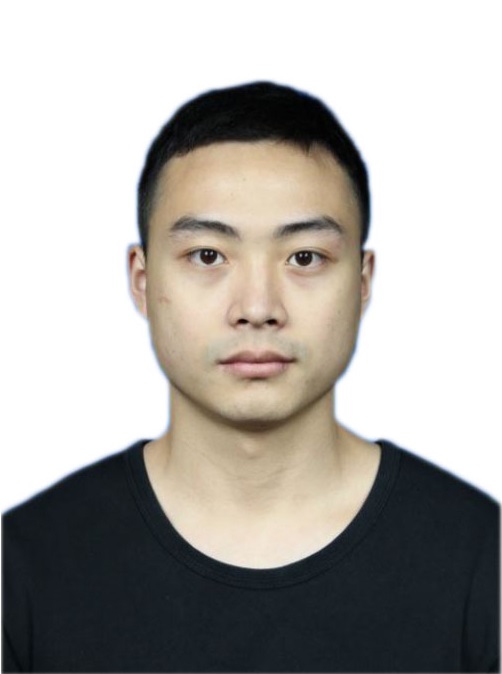}}]{Zhiqiang Deng}
received the B.S. degree in automation from Shenyang Jianzhu University, Shenyang, China, in 2020. He is currently pursuing the master’s degree in pattern recognition and intelligent systems from the College of Information Science and Engineering, Northeastern University, Shenyang, China. His research interests include visual simultaneous localization and mapping (SLAM) and augmented reality. 
\end{IEEEbiography}

\begin{IEEEbiography}[{\includegraphics[scale=0.75]{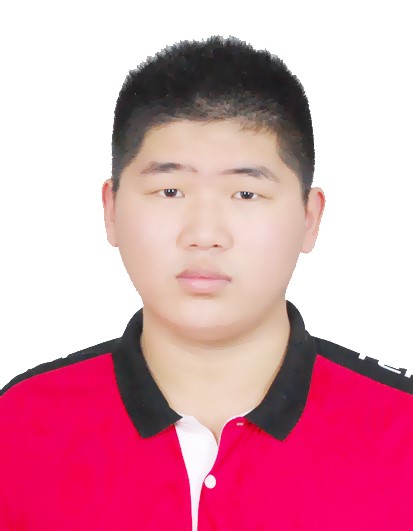}}]{Wenkai Sun}
received the B.S. degree in automation from Tiangong University, Tianjin, China, in 2020. He is currently pursuing the master’s degree in pattern recognition and intelligent systems from the College of Information Science and Engineering, Northeastern University, Shenyang, China. His research interests include semantic SLAM and augmented reality. 
\end{IEEEbiography}

\begin{IEEEbiography}[{\includegraphics[scale=0.21]{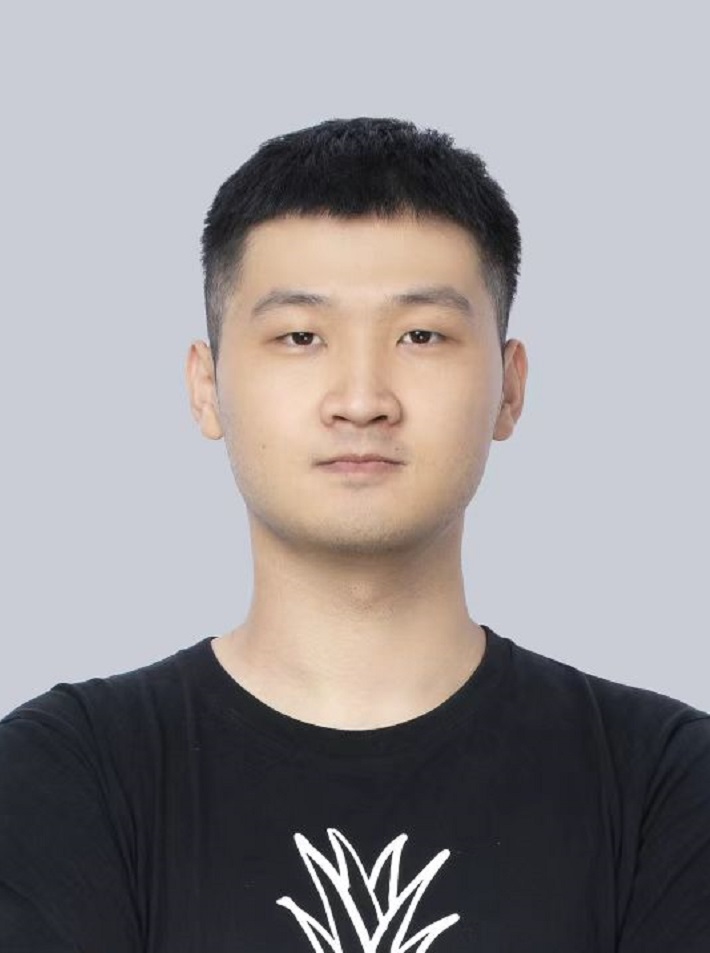}}]{Xin Chen}
received the B.S. degree in Mechanical and Electronic Engineering from Harbin University of Science and Technology, in 2019, and the M.S. degree in robot science and engineering
from Northeastern University, Shenyang, Liaoning,
China, in 2022. His research interests include object pose estimation and robot manipulation.
\end{IEEEbiography}

\begin{IEEEbiography}[{\includegraphics[width=1in,height=1.40in,clip,keepaspectratio]{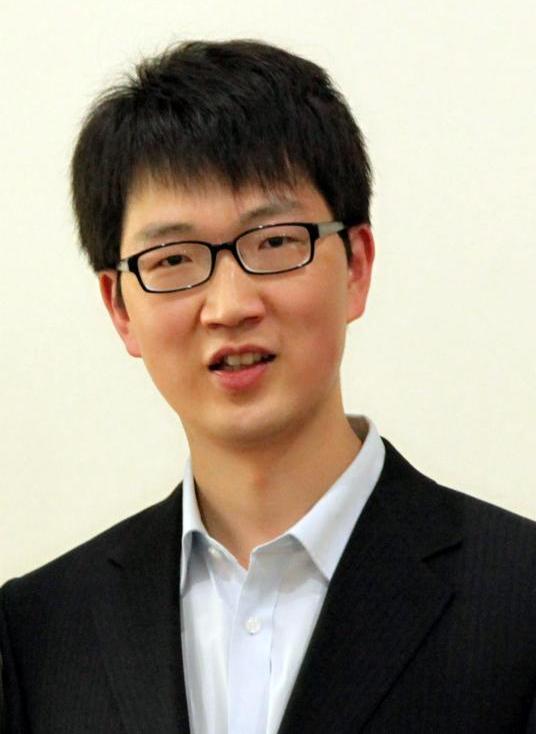}}]{Jian Zhang}
(M’14) received the B.S. degree from the Department of Mathematics, Harbin Institute
of Technology (HIT), Harbin, China, in 2007, and received his M.Eng. and Ph.D. degrees from the School of Computer Science and Technology, HIT, in 2009 and 2014, respectively. From 2014 to 2018, he worked as a postdoctoral researcher at Peking University (PKU), Hong Kong University of Science and Technology (HKUST), and King Abdullah University of Science and Technology (KAUST).

Currently, he is an Assistant Professor with the School of Electronic and Computer Engineering, Peking University Shenzhen Graduate School, Shenzhen, China. His research interests include intelligent multimedia processing, deep learning and optimization. He has published over 90 technical articles in refereed international journals and proceedings. He
received the Best Paper Award at the 2011 IEEE Visual Communications and Image Processing (VCIP) and was a co-recipient of the Best Paper Award of 2018 IEEE MultiMedia.
\end{IEEEbiography}






\end{document}